\documentclass[acmlarge]{acmart}

\usepackage{multirow}
\usepackage{pifont}

\usepackage{colortbl}

\newcommand{\ace}{\operatorname{ACE}}

\newcommand{\eat}[1]{}

\newcommand{\etal}{{et al.~}}       
\newcommand{\ie}{{i.e.~}}           
\newcommand{\wrt}{{w.r.t.~}}         
\newcommand{\aka}{{a.k.a.~}}

\newcommand\ci{\perp\!\!\!\perp}
\newcommand\nci{\not \! \perp\!\!\!\perp}

\def\circleleftarrow{\leftarrow\put(-2,2.5){\circle{2.5}}}
\def\cirlinecir{\put(0.9,2.1){\circle{2.5}}-\put(-0.9,2.1){\circle{2.5}}}

\def\astrightarrow{\put(0.4,-2.5){*}\rightarrow}
\def\astleftarrow{\leftarrow\put(-4,-2.5){*}}

\gdef\theorem@headerfont{\bfseries}
\newtheorem{definition}{\textbf{Definition}}

\AtBeginDocument{%
  \providecommand\BibTeX{{%
    \normalfont B\kern-0.5em{\scshape i\kern-0.25em b}\kern-0.8em\TeX}}}

\setcopyright{acmcopyright}
\copyrightyear{202*}
\acmYear{202*}
\acmDOI{XXXXXXX.XXXXXXX}

\acmJournal{JACM}
\acmVolume{xx}
\acmNumber{x}
\acmArticle{1}
\acmMonth{1}

\begin{document}

\title{Data-Driven Causal Effect Estimation Based on Graphical Causal Modelling: A Survey}

\author{Debo Cheng}
\email{chedy055@mymail.unisa.edu.au}
\authornote{Corresponding authors}
\orcid{0000-0002-0383-1462}
\author{Jiuyong Li}
\authornotemark[1]
\email{jiuyong.li@unisa.edu.au}
\orcid{0000-0002-9023-1878}
\author{Lin Liu}
\email{lin.liu@unisa.edu.au}
\orcid{0000-0003-2843-5738}
\author{Jixue Liu}
\email{jixue.liu@unisa.edu.au}
\orcid{0000-0002-0794-0404}
\author{Thuc Duy Le}
\email{thuc.le@unisa.edu.au}
\orcid{0000-0002-9732-4313}
\affiliation{
	\institution{UniSA STEM, University of South Australia}
	\state{Mawson Lakes}
	\city{Adelaide}
	\postcode{5095}
	\country{Australia}}

\renewcommand{\shortauthors}{Cheng and Li, et al.}

\begin{abstract}
In many fields of scientific research and real-world applications, unbiased estimation of causal effects from non-experimental data is crucial for understanding the mechanism underlying the data and for decision-making on effective responses or interventions. A great deal of research has been conducted  to address this challenging problem from different angles.  For estimating causal effect in observational data, assumptions such as Markov condition, faithfulness and causal sufficiency are always made. Under the assumptions, full knowledge such as, a set of covariates or an underlying causal graph, is typically required. A practical challenge is that in many applications, no such full knowledge or only some partial knowledge is available. In recent years, research has emerged to use search strategies based on graphical causal modelling to discover useful knowledge from data for causal effect estimation, with some mild assumptions, and has shown promise in tackling the practical challenge. In this survey, we review these data-driven methods on causal effect estimation for a single treatment with a single outcome of interest and focus on the challenges faced by data-driven causal effect estimation. We concisely summarise the basic concepts and theories that are essential for data-driven causal effect estimation using graphical causal modelling but are scattered around the literature. We identify and discuss the challenges faced by data-driven causal effect estimation and characterise the existing methods by their assumptions and the approaches to tackling the challenges. We analyse the strengths and limitations of the different types of methods and present an empirical evaluation to support the discussions. We hope this review will motivate more researchers to design better data-driven methods based on graphical causal modelling for the challenging problem of causal effect estimation.        
\end{abstract}



\maketitle

\section{Introduction}
\label{Sec:Intro}
Causal inference is a fundamental task in many fields, such as epidemiology~\cite{robins1992identifiability,greenland1999causal,hernan2020causal} and economics~\cite{imbens2015causal,abadie2016matching} for discovering and understanding the underlying
mechanisms of different phenomena~\cite{peters2017elements,scholkopf2022causality,pearl2009causality,pearl2018book,hernan2020causal}. One major task for causal inference is to estimate the causal effect (\aka treatment effect) of a treatment (or intervention) on an outcome of interest~\cite{spirtes2000causation,pearl2009causality,perkovic2018complete}. For example, medical researchers may want to query the causal effect of a new drug on a disease or the causal effect of temperature on atmospheric pollution~\cite{robins1986new,pearl2009causal,imbens2020potential,glymour2019review}. 

\textit{Randomised control trials} (RCTs) are the gold standard for estimating causal effects~\cite{deaton2018understanding,rubin1974estimating}. An RCT aims at removing the effects of other factors on the outcome by randomly assigning an individual to the treated group or the control group. Under randomised assignment, both groups of individuals have the same characteristics. Hence, the causal effects of the treatment on the outcome can be obtained directly by comparing the outcomes of the treated group and the control group. However, RCTs are often expensive or infeasible to conduct due to time constraints or ethical concerns~\cite{pearl2018book,maathuis2010predicting}. 

Estimating causal effect from observational data is an important alternative to RCTs~\cite{spirtes2000causation,pearl2009causality,imbens2015causal}. Generally speaking, four types of causal effects are often estimated, average causal effect (ACE)~\cite{rubin1974estimating,hill2011bayesian,henckel2022graphical,cheng2020causal}, average causal effect on the treated group (ACT)~\cite{Morgan2006Matching,abadie2016matching}, conditional average treatment effect (CATE)~\cite{athey2016recursive,athey2018approximate,athey2019generalized} and individual causal effect (ICE)~\cite{shalit2017estimating,yao2018representation}. ACE is to measure the average change of the outcome due to the application of a treatment at the population level. ACT aims to assess the average change of the outcome in the treated group only. CATE is a measurement of the average causal effect within  a sub-population. CATE is also known as heterogeneous causal effect (HCE). ICE is the causal effect at the individual level and is defined as the difference of the potential outcomes for an individual~\cite{rubin1974estimating,imbens2015causal}. 

This survey is focused on ACE estimation given its wide real-world applications. For example, doctors wish to know the overall efficacy of a new drug on blood pressure~\cite{rubin2007design,deaton2018understanding}; the government wants to assess the effectiveness of job training on the income in general~\cite{lalonde1986evaluating,cheng2020causal}; an airline is interested in evaluating the effects of ticket prices on customers' purchase tendency~\cite{hartford2017deep,hartford2021valid}. ACE estimation can involve either a single treatment or multiple treatments on a single outcome or multiple outcomes~\cite{wang2019blessings,ma2021multi,nabi2022semiparametric}. This survey focuses on the ATE estimation of a single treatment on a single outcome. The data-driven algorithms for causal effect estimation based on graphical causal modelling for multiple treatments and/or multiple outcomes are in  the theoretical
research stage~\cite{perkovic2018complete,jaber2019causal,jaber2019identification}, and there are currently no fully data-driven algorithms available for such problems. Hence, we will not discuss them in this survey. 

\begin{figure}[t]
	\centering
	\includegraphics[scale=0.37]{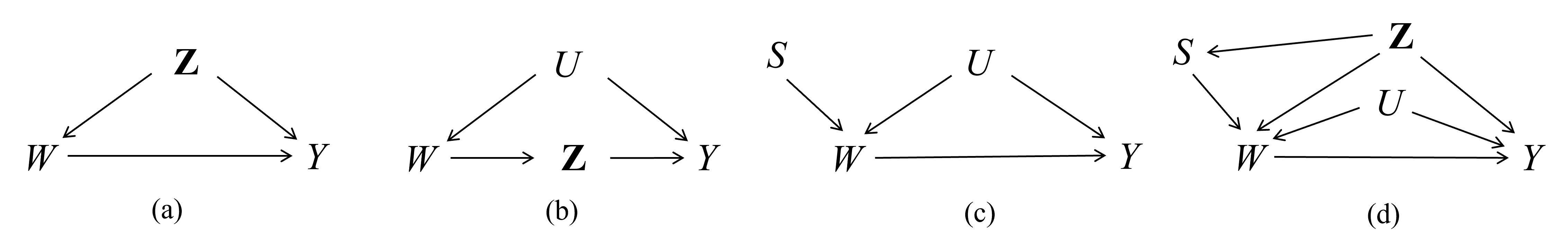}
	\caption{Exemplary DAGs  show (a) confounder, (b) mediator, (c) instrumental variable and (d) conditional instrumental variable. In DAG (a), $\mathbf{Z}$ is a set of confounders \wrt the pair $(W, Y)$; in DAG (b), $\mathbf{Z}$ is a set of mediator \wrt the pair $(W, Y)$;  in DAG (c), $S$ is an instrumental variable \wrt the pair $(W, Y)$; and in DAG (d),  $S$ is a conditional instrumental variable conditioning on  the set $\mathbf{Z}$ \wrt the pair $(W, Y)$.}
	\label{fig:intro}
\end{figure}

The main challenge for ACE estimation from observational data is confounding bias that is caused by confounders that are common causes affecting both the treatment $W$ and outcome $Y$~\cite{pearl2009causality,vanderweele2011new}. For example, the set of variables $\mathbf{Z}$ in Fig.~\ref{fig:intro} (a) is a set of confounders. To estimate the causal effect unbiasedly of  $W$ on $Y$ from observational data, the confounding bias caused by the set of confounders $\mathbf{Z}$ needs to be removed. The main techniques for obtaining unbiased ACE estimation from observational data are confounding adjustment~\cite{imbens2015causal,hernan2020causal} and   the instrumental variable (IV) approach~\cite{hernan2006instruments,imbens2014instrumental}.  Confounding adjustment aims at mitigating the influences of confounders on the outcome when estimating the causal effect of $W$ on $Y$, and it requires  all confounders are measured~\cite{pearl2009causality}. The IV approach leverages a special variable called the instrumental variable to eliminate confounding bias in the estimation of the causal effect of $W$ on $Y$  and the approach works even when there are unmeasured confounders~\cite{brito2002generalized}. 
 
One focus of this survey is to review the data-driven confounding adjustment methods which are based on the back-door criterion  or  variations~\cite{pearl2009causality,shpitser2012validity,maathuis2015generalized,perkovic2018complete} without requiring a given causal graph. In graphical causal modelling term, a set of variables used for confounding adjustment is a set of variables satisfying the back-door criterion (See Definition~\ref{Causalsufficiency} in Section 2) when the causal DAG is given and all variables in the set need observed and measured. For example, the set $\mathbf{Z}$ in Fig.~\ref{fig:intro} (a)   {meets} the back-door criterion and hence can be used as an adjustment set. There are variations of the back-door criterion for identifying adjustment sets given a causal graph~\cite{pearl2009causality,shpitser2012validity,maathuis2015generalized,perkovic2018complete}. When a causal graph is given, an adjustment set can be read from the graph. It is challenging but practical to identify an adjustment set using data without a causal graph.

 Another focus of this survey is to review data-driven methods that search for potential IVs or CIVs (conditional IVs and their corresponding conditioning sets) for ACE estimation. The  IV approach~\cite{angrist1995two,angrist1996,pearl1995testability,hernan2006instruments,martens2006instrumental,yuan2022auto} is a powerful way to estimate causal effect from observational data in the presence of latent confounders.  A standard IV is a special cause of $W$ (See Section Definition~\ref{def:standardIV} in Section 2), and an example of IV is shown in Fig.~\ref{fig:intro}. When a standard IV is known, the Two-Stage Least Squares (TSLS) estimator~\cite{angrist1995two} is commonly used to estimate the ACE. A standard IV can be relaxed to a conditional   IV (CIV)~\cite{brito2002generalized,imbens2014instrumental},  which needs a companion set, called a conditioning set, for ATE estimation. For example, in Fig.~\ref{fig:intro}(d), $S$ is a CIV conditioning on the set $\mathbf{Z}$. With a CIV $S$ and a conditioning set $\mathbf{Z}$ for the CIV, the Two-Stage Least Squares for CIV (TSLS.CIV  estimator) ~\cite{imbens2014instrumental} can be used to estimate the average causal effect unbiasedly. An IV or a CIV is generally not identifiable in data and needs to be given by domain knowledge. It is desirable to find an IV or a CIV (and its conditioning set) in data without the full domain knowledge or the causal graph, but with mild assumptions.

The data-driven methods surveyed in this paper do not require the full causal knowledge, \ie a causal graph,  adjustment set,  IV, or a CIV and its conditioning set, whereas most existing (non-data-driven) methods require the full knowledge~\cite{pearl2009causal,spirtes2010introduction,guo2020survey,witte2019covariate}. In many real-world applications, the full knowledge is unavailable. Hence, data-driven methods are needed  and practically useful.

 In recent years, data-driven methods have emerged to use search strategies (such as those used in causal discovery methods~\cite{spirtes2000causation,aliferis2010local}) to discover necessary information for causal effect estimation with mild assumptions ~\cite{maathuis2009estimating,le2013inferring,maathuis2015generalized,perkovic2017interpreting,perkovic2018complete,fang2020ida,cheng2022toward,cheng2022discovering}, and they have shown promise in tackling the practical challenge that users do not have the full knowledge for causal effect estimation. These data-driven methods mostly leverage graphical causal models, which provide a powerful language for discovering necessary knowledge from data for causal effect estimation~\cite{pearl2009causality,pearl2009causal,zander2016separators,van2019separators}. However, to date, there is not a comprehensive survey that discusses the challenges encountered by the data-driven methods  and the strategies to tackle the challenges.

This survey reviews the theories of the graphical causal models that support data-driven causal effect estimation, identifies three challenging problems for designing data-driven methods based on graphical causal modelling and discusses how the existing data-driven causal effect estimation methods handle the challenges. Our goal is to identify the challenges and  to guide users to choose appropriate methods and motivate researchers to design smarter data-driven methods for causal effect estimation based on graphical causal modelling. 

To summarise, the contributions of this survey are as follows.
 \begin{itemize}
 	\item We provide a concise review of the fundamental assumptions, theories and definitions of graphical causal modelling for the existing data-driven causal effect estimation methods. We aim at  providing an introduction to the set of concepts and theorems that are essential for data-driven causal effect estimation and major theoretical results in this area. The theorems and suppositions on which the data-driven methods are based are scattered in many different research papers. We summarise them for the convenience of researchers who are interested in venturing into this changeling and important area. 
 	\item We analyse and identify the major challenges confronting data-driven causal effect estimation, under the framework of graphical causal modelling, and categorise existing data-driven ACE estimation methods according to  assumptions and the approaches that they handle the challenges, and review the methods in detail. We provide practical guidance on using the methods   and an empirical evaluation of these methods.  
 	\item To the best of our knowledge, this is the first paper that provides a systematic survey on data-driven causal effect estimation methods through the lens of graphical causal modelling. This distinguishes our paper from the existing surveys. Specifically, the surveys in~\cite{stuart2010matching,guo2020survey,yao2021survey,witte2019covariate} focus on the causal effect estimation methods based on potential outcome methods and the surveys in~\cite{glymour2019review,vowels2021d,nogueira2022methods,yu2021unified} are on causal structure learning methods and their tool-kits. 
 \end{itemize}

\section{Background}
\label{sec:background}
In this section, we introduce the  fundamental concepts of causal graphs,   CPDAGs \& PAGs, and causal effect estimation  through covariate adjustment and by instrumental variables, respectively. Throughout the whole survey, we use an upper case letter, lower case letter, boldfaced upper case letter, and calligraphy letter, e.g., $X$, $x$, $\mathbf{X}$ and $\mathcal{G}$ to denote a variable, a variable value, a set of variables and a graph, respectively. In particular, we use $W$ and $Y$ to denote  the treatment variable and the outcome variable, respectively. 

\subsection{Causal Graphs}
\label{subsec:causalgraph}
Causal relationships among variables are normally represented by a directed acyclic graph (DAG), denoted as $\mathcal{G}=(\mathbf{X}, \mathbf{E})$, which contains a set of nodes (variables), directed edges between nodes, \ie $\mathbf{E}$, and has no directed cycles. In a DAG, when there is a directed edge $X_i\rightarrow X_j$, $X_i$ is known as the parent of $X_j$ and $X_j$ is a child of $X_i$.  A path between two nodes $(X_i, X_j)$  is a set of consecutive edges linking $(X_i, X_j)$  regardless of the edge directions,  and a directed path from $X_i$ to $X_j$ contains a set of consecutive edges pointing towards $X_j$, \ie $X_i\rightarrow \dots \rightarrow X_j$. $X_i$ is an ancestor of $X_j$ and $X_j$ is a descendant of $X_i$ if there is a directed path from $X_i$ to $X_j$. We use $Pa(X)$, $Ch(X)$, $An(X)$ and $De(X)$ to denote the sets of all parents, children, ancestors and descendants of $X$, respectively. 

In a DAG, if an edge $X_i\rightarrow X_j$ denotes that $X_i$ is a direct cause of $X_j$,   the DAG is termed a causal DAG. In a causal DAG $\mathcal{G}$, a directed path between two nodes is called a causal path and a non-directed path between two nodes is a non-causal path. Let $\ast$ be an arbitrary edge mark. $X_{i}$ is a collider on a path $\pi$ if $X_k\astrightarrow X_{i} \astleftarrow X_j$ is a sub-path of $\pi$. A collider path is a path with every non-endpoint node being a collider. A path of length one is a trivial collider path. 

Some assumptions are needed to use a causal DAG $\mathcal{G}$ to represent the data generation mechanism, including the Markov condition, the faithfulness assumption and causal sufficiency assumption.

\begin{definition}[Markov condition~\cite{pearl2009causality,aliferis2010local}] Given a joint distribution $P(\mathbf{X})$ and a DAG $\mathcal{G}=(\mathbf{X}, \mathbf{E})$,   $\mathcal{G}$ and $P(\mathbf{X})$ satisfy the \textit{Markov condition} if $\forall X_i \in \mathbf{X}$, $X_i$ is independent of all of its non-descendants in $\mathcal{G}$, given the set of parents of $X_i$, \ie $Pa(X_i)$. 
\end{definition}

The Markov condition indicates that the distribution of $X_i$ is determined by the distributions of its parents solely. With the Markov condition, $P(\mathbf{X})$ can be factorised as $P(\mathbf{X}) = \prod_i P(X_i | Pa (X_i))$ according to $\mathcal{G}$.
\begin{definition}[Faithfulness~\cite{spirtes2000causation,glymour1999computation}] A DAG $\mathcal{G}=(\mathbf{X}, \mathbf{E})$ is faithful to a joint distribution $P(\mathbf{X})$ if and only if every independence present in $P(\mathbf{X})$ is entailed by $\mathcal{G}$ and the Markov condition. A joint distribution $P(\mathbf{X})$ is faithful to a DAG $\mathcal{G}$ if and only if the DAG $\mathcal{G}$ is faithful to $P(\mathbf{X})$. 
\end{definition}

\begin{definition}[Causal sufficiency~\cite{spirtes2000causation}]
	\label{Causalsufficiency}
	A given dataset satisfies causal sufficiency if for every pair of observed variables, all their common causes are observed.
\end{definition}

When a distribution $P(\mathbf{X})$ and a DAG  $\mathcal{G}=(\mathbf{X}, \mathbf{E})$ are faithful to each other, if two variables $X_i$ and $X_j$ in  $\mathcal{G}$ are $d$-separated by another set of variables $\mathbf{Z}$ in $\mathcal{G}$ as defined below, $X_i$ and $X_j$ are conditional independent given $\mathbf{Z}$. That is, conditional independence and dependence relationships between variables in $P(\mathbf{X})$ can be read off  from the DAG based on $d$-separation or $d$-connection.

\begin{definition}[$d$-separation/$d$-connection~\cite{pearl2009causality}] 
A path $\pi$ in a DAG $\mathcal{G}=(\mathbf{X}, \mathbf{E})$ is said to be $d$-separated  by a set of variables $\mathbf{Z}$ if and only if (i) $\pi$ contains a chain $X_i \rightarrow X_k \rightarrow X_j$ or a fork $X_i \leftarrow X_k \rightarrow X_j$ such that the middle node $X_k \in \mathbf{Z}$, or (ii) $\pi$ contains a collider $X_k$ (\ie  $\pi$ contains the sub-path $X_i\rightarrow X_k \leftarrow X_j$) such that $X_k\notin \mathbf{Z}$ and none of the descendants of $X_k$ is in $\mathbf{Z}$. A set $\mathbf{Z}$ is said to d-separate $X_i$ from $X_j$ if and only if $\mathbf{Z}$ $d$-separates every path between $X_i$ and $X_j$ (denoted  as $X_i\ci_{d}X_j\mid\mathbf{Z}$), otherwise $X_i$ and $X_j$ are d-connected given $\mathbf{Z}$, \ie $X_i\nci_{d}X_j\mid\mathbf{Z}$.
\end{definition}

\begin{figure}[t]
	\centering
	\includegraphics[scale=0.33]{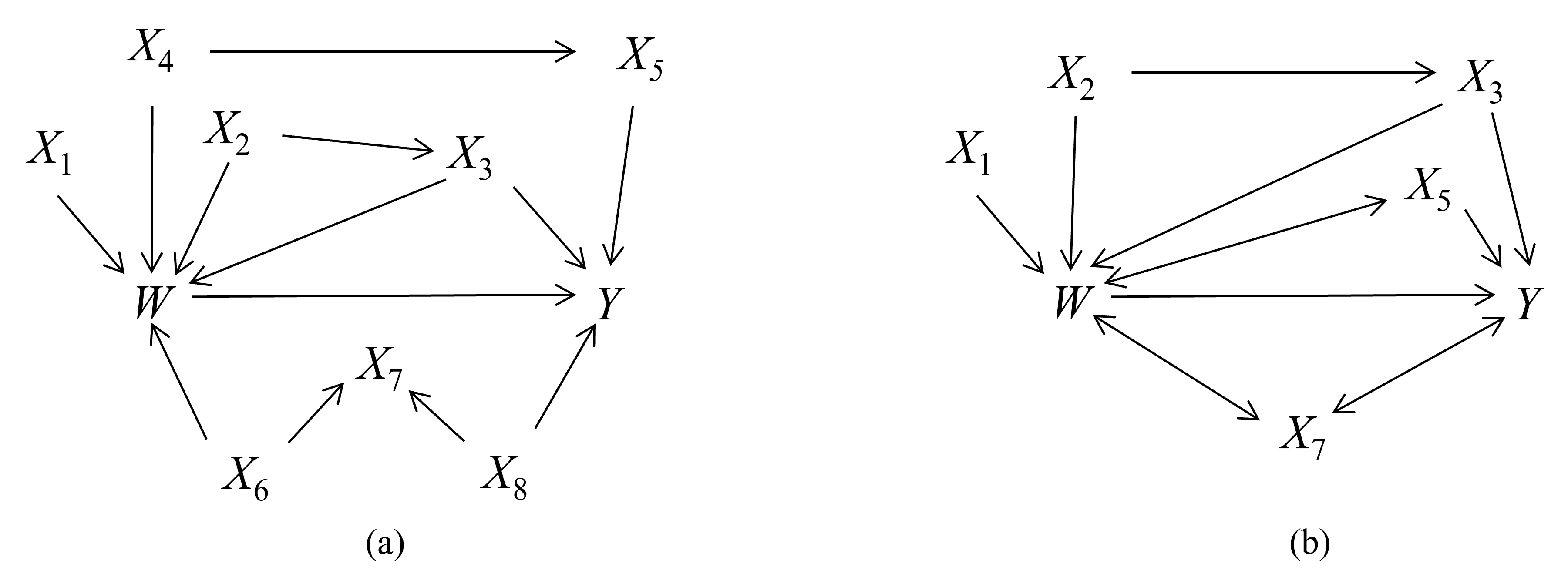}
	\caption{An  {exemplary} DAG (a) and MAG (b). DAG (a) consists of $\{X_1, X_2, \dots, X_8, W, Y\}$ and MAG (b) is obtained from DAG (a) with $\{X_4, X_6, X_8\}$ unmeasured.}
	\label{fig:exampleDAG_MAG}
\end{figure}

An  exemplary DAG representing a causal mechanism (causal relationships between all variables in a problem domain) is shown in Fig.~\ref{fig:exampleDAG_MAG} (a).  In the DAG, $X_1 \to W \to Y$ is a causal path, and this path is $d$-separated by $W$. When we consider a $d$-separate set $\mathbf{Z}$ for the pair $(X_1, Y)$, all five paths between the two variables need to be $d$-separated by $\mathbf{Z}$. $W$ should be in $\mathbf{Z}$ because it is the only variable $d$-separating the path $X_1 \to W \to Y$.  Now,   $\mathbf{Z}$ $d$-connects $(X_4, X_2)$, $(X_4, X_3)$, $(X_4, X_6)$, $(X_2, X_6)$, and $(X_3, X_6)$ on the four other paths respectively since $W$ is a collider in every path. Let's look into these four paths separately. Path $X_1 \to W \leftarrow X_6 \to X_7 \leftarrow X_8 \to Y$ is $d$-separated by $\emptyset$ since $X_7$ is a collider even when the collider $W$ is included in $\mathbf{Z}$. Path $X_1 \to W \leftarrow X_4 \to X_5 \to Y$ is $d$-separated by set $\{X_4\}$ or $\{X_5\}$. Paths $X_1 \to W \leftarrow X_2 \to X_3 \to Y$ and $X_1 \to W \leftarrow X_3 \to Y$ are $d$-separated by $\{X_3\}$.  Consequently, the pair $(X_1, Y)$ is $d$-separated by set $\mathbf{Z} = \{X_3, X_4, W\}$ or $\mathbf{Z} = \{X_3, X_5, W\}$. 
 
 In  numerous real-world applications, latent common causes (\aka latent confounders) are commonplace such that the causal sufficiency assumption is violated~\cite{spirtes2000causation,pearl2009causality}. In this situation, the ancestral graph is designed for representing causal ancestral relationships~\cite{richardson2002ancestral,spirtes2003causal}. An ancestral graph is a mixed graph without directed cycles or almost directed cycles. An ancestral graph contains both directed edges $\rightarrow$ and bidirected edges $\leftrightarrow$ between variables. An almost directed cycles occurs if $X_i \leftrightarrow X_j$ in $\mathcal{G}$ and $X_i\in An(X_j)$.
 Similar to $d$-separation in a DAG, $m$-separation  serves as a general criterion to read off the independencies and dependencies between measured variables entailed by an ancestral graph~\cite{richardson2002ancestral}.

\begin{definition}[$m$-separation/$m$-connection~\cite{richardson2002ancestral}]
\label{def:m-sep}
 In an ancestral graph $\mathcal{G}=(\mathbf{X}, \mathbf{E})$, a path $\pi$ between $X_{i}$ and $X_{j}$ is said to be $m$-separated by a set of nodes $\mathbf{Z}\subseteq \mathbf{X}\setminus\{X_i, X_j\}$ (possibly $\emptyset$) if
(i)  every non-collider on $\pi$ is a member of $\mathbf{Z}$;
(ii) every collider on $\pi$ is not a member of $\mathbf{Z}$ and none of the descendants of the colliders is in $\mathbf{Z}$.
Two nodes $X_{i}$ and $X_{j}$ are said to be $m$-separated  by $\mathbf{Z}$ in $\mathcal{G}$, denoted as $X_i \ci_m X_j|\mathbf{Z}$ if every path between $X_{i}$ and $X_{j}$ is $m$-separated by $\mathbf{Z}$; otherwise they are said to be m-connected by $\mathbf{Z}$, denoted as $X_i\nci_m X_j|\mathbf{Z}$.
\end{definition}

 We take the ancestral graph in Fig.~\ref{fig:exampleDAG_MAG} (b) as an example for understanding $m$-separation. $X_1 \to W \to Y$ is a causal path and the path is $m$-separated by $W$. When we consider an $m$-separated set $\mathbf{Z}$ for the pair $(X_1, Y)$, all five paths between the two nodes need to be m-separated by $\mathbf{Z}$. $W$ should be in $\mathbf{Z}$ because it is the only variable $m$-separating the path $X_1 \to W \to Y$. Path $X_1 \to W \leftrightarrow X_7 \leftrightarrow Y$ is $m$-separated by $\emptyset$ since $X_7$ is a collider. Path $X_1 \to W \leftarrow X_2 \to X_3 \to Y$ is $m$-separated by set $\{X_3\}$. Paths $X_1 \to W \leftarrow  X_3 \to Y$ and $X_1 \to W \leftrightarrow X_5 \to Y$ are $m$-separated by $\{X_3, X_5\}$. Hence, the pair $(X_1, Y)$ is $m$-separated by set $\mathbf{Z} = \{W, X_3, X_5\}$. 

 When there are latent confounders in a system, a Maximal Ancestral Graph (MAG) is commonly used to represent causal relationships among measured variables~\cite{richardson2002ancestral,spirtes2003causal,zhang2008causal}.

\begin{definition}[Maximal Ancestral Graph (MAG)] 
	A MAG $\mathcal{G}=(\mathbf{X}, \mathbf{E})$ contains directed and bidirected edges, and satisfies the conditions that (i) there is not a directed cycle or an almost directed cycle, and (ii) every pair of non-adjacent nodes $X_{i}$ and $X_{j}$ in $\mathbf{X}$ is $m$-separated by a set $\mathbf{Z}\subseteq \mathbf{X}\backslash \{X_{i}, X_{j}\}$.   
\end{definition}

A bidirected edge $X_i \leftrightarrow X_j$ in a MAG indicates the presence of a latent confounder between $X_i$ and $X_j$, and the absence of a direct causal relationship between $X_i$ and $X_j$. A causal path in a MAG is a directed path without a bidirected edge. In a MAG, $X_i$ and $X_j$ are \textit{spouses} if there exists a bidirected edge $X_i \leftrightarrow X_j$.  The assumptions of Markov condition and faithfulness are still held in a MAG. Note that the causal relationships between measured variables in a MAG are still the same as in the underlying DAG over the measured and unmeasured variables~\cite{richardson2002ancestral,zhang2008completeness,zhang2008causal}. Thus, in a MAG, we still use $Pa(X)$, $Ch(X)$, $An(X)$ and $De(X)$ to denote the sets of all parents, children, ancestors and descendants of $X$, respectively.

 Another type of causal graphs to represent causal relationships between variables with latent confounders is the acyclic directed mixed graph (ADMG)~\cite{tian2002general,pena2018reasoning,runge2021necessary}.  ADMGs are generalisations of DAGs~\cite{silva2011mixed,evans2014markovian}, meaning that the causal relationships between observed variables in ADMGs can be encoded by marginalising causal DAGs~\cite{richardson2002ancestral,richardson2003markov}. In an ADMAG, it is possible for two nodes to have more than one edge between them. For example, both a directed edge ($\rightarrow$) and a bidirected edge ($\leftrightarrow$) connect two nodes in an ADMG. Some works for using ADMG for causal discovery and inference can be found in~\cite{silva2011mixed,evans2014markovian,bhattacharya2020semiparametric,runge2021necessary}. However, it is challenging to learn an ADMG from data since it allows $W \rightarrow Y$ and $W \leftrightarrow Y$ simultaneously.  To the best of our knowledge, there are no established data-driven approaches for learning ADMGs from data, which leads to the limited use of ADMGs in data-driven causal effect estimation. Consequently, we have refrained from reviewing ADMG-based methods in the subsequent sections. 

\subsection{CPDAG \& PAG}
Learning a DAG (or a MAG) directly from observational data is challenging.   Instead, most existing structured learning algorithms~\cite{neapolitan2004learning,spirtes2010introduction,aliferis2010local} are capable of learning a Markov equivalence class of DAGs (or MAGs) from the data (with latent confounders). A Markov equivalence class refers to a set of DAGs (or MAGs) encode the same set of conditional independence relations between measured variables.

\begin{figure}[t]
	\centering
	\includegraphics[scale=0.38]{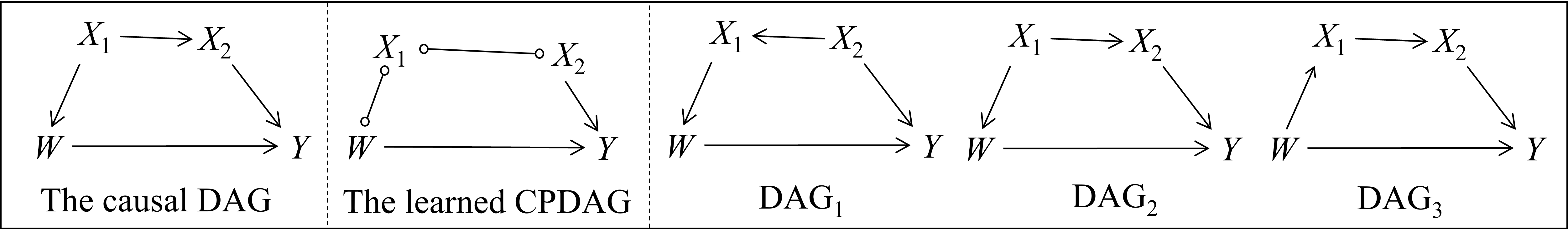}
	\caption{The underlying causal DAG is given in the first panel, the CPDAG learned from the data generated from the causal DAG is presented in the second panel, and  three DAGs ($DAG_1$, $DAG_2$, $DAG_3$) are encoded in the learned CPDAG.}
	\label{fig:examplecpdag_dags}
\end{figure}

A CPDAG (complete partial DAG) is used to represent a Markov equivalence class of DAGs~\cite{pearl2009causality,spirtes2010introduction,aliferis2010local}, \ie all the DAGs that encode the same d-separations as the underlying causal DAG. In a CPDAG, an undirected edge indicates that the orientation of this edge is uncertain. For example, the edge $W\;\cirlinecir\; X_1$ in the learned CPDAG in Fig.~\ref{fig:examplecpdag_dags} indicates a directed edge $W\rightarrow X_1$ or $W\leftarrow X_1$.

\begin{figure}[t]
	\centering
	\includegraphics[scale=0.34]{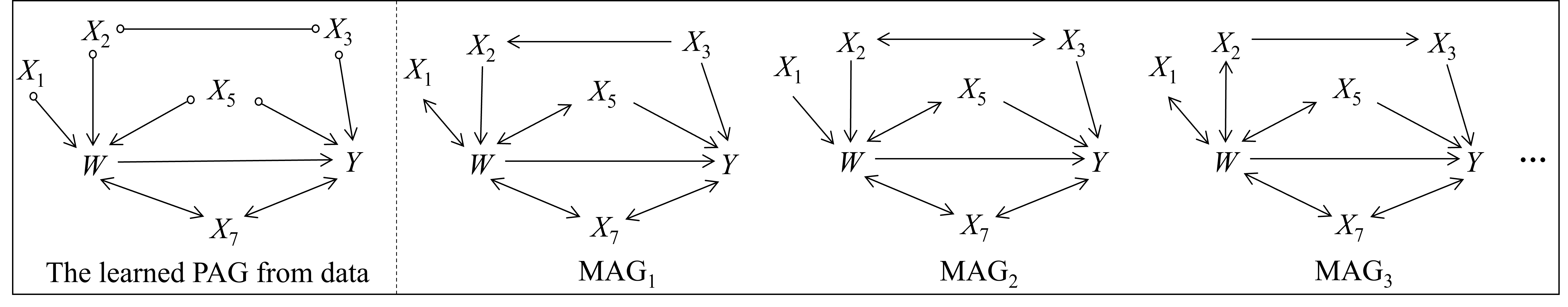}
	\caption{The left panel shows the PAG learned from the data generated from the MAG in Fig.~\ref{fig:exampleDAG_MAG} (b). There are around 50 MAGs encoded in the learned PAG and the right panel lists three exemplary MAGs encoded in the PAG.}
	\label{fig:example_pag_mags}
\end{figure}

A PAG (Partial Ancestral Graph) is often used to represent a Markov equivalence class of MAGs~\cite{spirtes2000causation,spirtes2003causal,richardson2002ancestral,ali2009markov}, \ie  all the MAGs that encode the same $m$-separations/$m$-connections as the underlying causal MAG. In a PAG, an undirected edge is allowed in addition to the edge types in a MAG.  Like an undirected edge in a CPDAG, an undirected edge in a PAG also indicates uncertainty. For example, the edge $W \circleleftarrow X_2$ in the PAG in Fig.~\ref{fig:example_pag_mags} indicates a directed edge  $W \leftarrow X_2$ or $W \leftrightarrow X_2$. 

Because a PAG contains some edges that are not oriented, we need to introduce the term ``\emph{definite status}'' for  nodes and paths to add some certainty  for $m$-separations and $m$-connections among nodes. In a PAG, a node $X_j$ is a \emph{definite non-collider} on $\pi$ if there exists at least an edge out of $X_j$ on $\pi$, or both edges have a circle mark ``$\circ$'' at $X_j$ and the sub-path $\langle X_i, X_j, X_k\rangle$ is an unshielded triple~\cite{zhang2008causal}. A node $X_i$ is known to have \emph{definite status} if it is a collider or a definite non-collider on the path. A collider on a path $\pi$ is referred to as a definite collider since it is always of \emph{definite status}. A path $\pi$ is of \emph{definite status} if every node (excluding non-endpoint node) on $\pi$ is of definite status. 

\subsection{Average Causal Effect, Confounding Adjustment and Graphical Criteria}
\label{subsec:ace}
Let $W$ be a binary variable representing treatment status ($W$=1 for treated and  {$W$}=0 for untreated) and $Y$ the outcome variable. The average causal effect (ACE) of $W$ on $Y$ is defined as $\ace(W, Y) =\mathbf{E}(Y\mid do(w=1))-\mathbf{E}(Y\mid do(w=0))$, where the $do$-operator, $do(\cdot)$~\cite{pearl2009causal} represents an intervention by setting a variable to a specific value in an experiment.

Confounding adjustment is the most common way for obtaining unbiased causal effect estimation from observational data. Given an adjustment set $\mathbf{Z}$, $\ace(W, Y)$ can be estimated unbiasedly as follows:
\[\ace(W, Y) =  \Sigma_{\mathbf{z}}[\mathbf{E}(Y\mid w=1,\mathbf{Z}=\mathbf{z})- \\  \mathbf{E}(Y\mid w=0,\mathbf{Z}=\mathbf{z})]P(\mathbf{Z}=\mathbf{z})\]

Therefore, the task of causal effect estimation is transformed into identifying an appropriate adjustment set. Note that, there may exist multiple adjustment sets, all leading to the unbiased estimation of $\ace(W, Y)$.

Based on causal graphs, two main criteria have been developed and widely used, the back-door criterion~\cite{pearl2009causal} and the generalised back-door criterion~\cite{maathuis2015generalized} for adjustment set identification in a causal DAG and MAG respectively. 

\begin{definition}[The back-door criterion~\cite{pearl2009causal}] In a causal DAG $\mathcal{G}=(\mathbf{X}, \mathbf{E})$, a set $\mathbf{Z}$ satisfies the back-door criterion w.r.t. the pair of variables $(W, Y)$ if (i) $\mathbf{Z}$ does not contain a descendant of $W$, and (ii) $\mathbf{Z}$ $d$-separates all the back-door paths between $W$ and $Y$ (i.e., the paths starting with an edge pointing to $W$).
\end{definition}
 
We use the DAG in Fig.~\ref{fig:exampleDAG_MAG} (a) to illustrate the back-door criterion. The set $\{X_3, X_5\}$ satisfies the back-door criterion and is a proper adjustment set w.r.t. $(W, Y)$,   as it $d$-separates all three back-door paths:  $W \leftarrow X_4 \to X_5 \to Y$, $W \leftarrow X_2 \to X_3 \to Y$, and $W \leftarrow X_3  \to Y$. The other sets satisfying the back-door criterion  are $\{X_3, X_4\}$, $\{X_2, X_3, X_4\}$, $\{X_2, X_3, X_5\}$, $\{X_3, X_4, X_5\}$ and $\{X_2, X_3, X_4, X_5\}$. 

It is worth mentioning that in a given causal graph, the three do-calculus rules proposed by Pearl~\cite{pearl1995causal,pearl2009causality} are sound and complete for determining whether a causal effect is identifiable~\cite{shpitser2006identification,shpitser2008complete}.   

The generalised back-door criterion in a MAG is introduced as follows.

\begin{definition}[The generalised back-door criterion~\cite{maathuis2015generalized}]
	 In a MAG $\mathcal{G}=(\mathbf{X}, \mathbf{E})$, a set $\mathbf{Z}$ satisfies the generalised back-door criterion \wrt  $(W, Y)$ if (i) $\mathbf{Z}$ does not include a descendant of $W$, and (ii) $\mathbf{Z}$ $m$-separates all non-causal paths between $W$ and $Y$.
\end{definition}

We use the MAG in Fig.~\ref{fig:exampleDAG_MAG} (b) to illustrate the generalised back-door criterion. The set $\{X_2, X_5\}$ satisfies the generalised back-door criterion \wrt $(W, Y)$  {as} it $m$-separates all non-causal paths between $W$ and $Y$: $W \leftarrow X_2\rightarrow X_3 \to Y$, $W \leftarrow X_3 \to Y$, $W \leftrightarrow X_5\rightarrow Y$,  and $W\leftrightarrow  X_7 \leftrightarrow Y$. The other sets satisfying the generalised back-door criterion are $\{X_3, X_5\}$ and $\{X_2, X_3, X_5\}$.

Two other graphical criteria, the adjustment criterion~\cite{shpitser2012validity} and the generalised adjustment criterion~\cite{perkovic2018complete} extend the back-door criterion and the generalised back-door criterion,  respectively, for the identification of adjustment sets. The adjustment criterion can be used to determine an adjustment set in a DAG, while the generalised adjustment criterion can be used to determine an adjustment set in four types of graphs, including DAG, CPDAG, MAG and PAG.  The generalised adjustment criterion used for data-driven causal effect estimation will be introduced in Subsection~\ref{subsec:GAC}.

 Once an adjustment set is obtained, a confounding adjustment method can be used to remove confounding bias. Many confounding adjustment methods have been developed, such as Nearest Neighbor Matching (NNM)~\cite{rosenbaum1985bias,sekhon2008multivariate,cheng2022sufficient}, Propensity Score Matching (PSM)~\cite{Morgan2006Matching,rubin1973matching,sekhon2008multivariate}, Covariate Balance Propensity Score (CBPS)~\cite{imai2014covariate}, Inverse Probability of Treatment effect Weighting (IPTW)~\cite{hirano2003efficient,robins2000marginal,cole2008constructing,hernan2020causal}, Doubly Robust Learning (DRL)~\cite{bang2005doubly,funk2011doubly,benkeser2017doubly}. For more details on confounding adjustment, please refer to the surveys~\cite{stuart2010matching,sauer2013review,imai2014covariate,witte2019covariate,yao2021survey,guo2020survey}, or the books~\cite{pearl2009causality,koller2009probabilistic,imbens2015causal,morgan2015counterfactuals,hernan2020causal}.

 \subsection{Instrumental Variable Approach \& Other Alternative Approaches}
\label{Subsec:IV}
A proper adjustment set must include all variables that are sufficient to remove the confounding bias between $W$ and $Y$.  However, when there is a latent common cause of $W$ and $Y$, confounding adjustment does not work.  In such cases, the instrumental variable (IV) approach is a powerful way to estimate average causal effect~\cite{hernan2006instruments,martens2006instrumental,abadie2003semiparametric,angrist1995two}.  

\begin{figure}[t]
	\centering
	\includegraphics[scale=0.42]{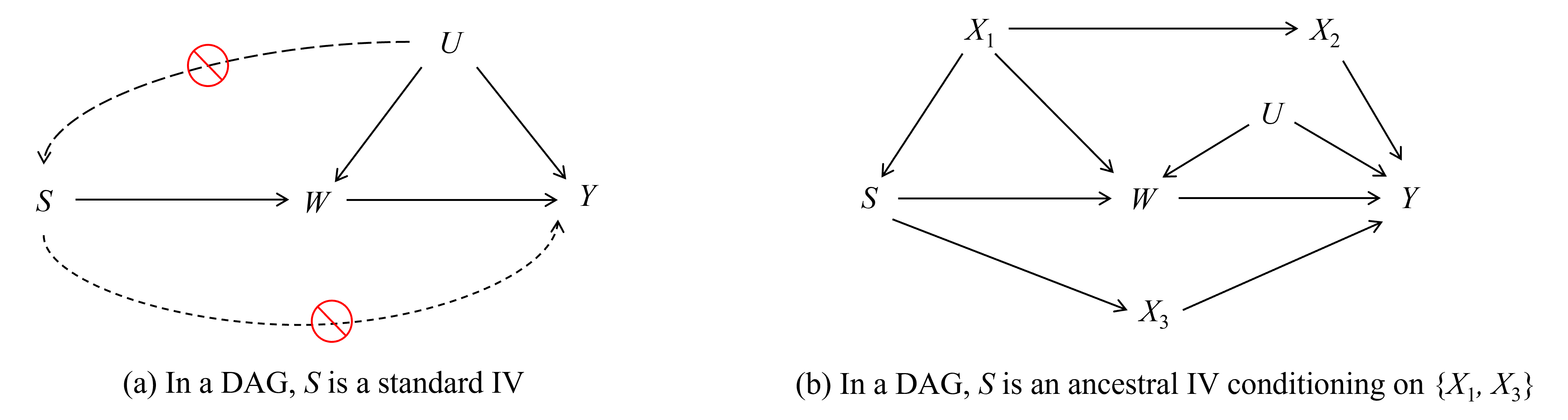}
	\caption{Comparing a standard IV (a) with an ancestral IV $S$ (b). In DAG (a), $S$ satisfies the three conditions of Definition~\ref{def:standardIV} and thus is a standard IV. In DAG (b), $S$ does not satisfy conditions (ii) and (iii) of a standard IV, but it is an ancestral IV conditioning on $\{X_1, X_3\}$.}
	\label{fig:exampleIV_AIV}
\end{figure}

\begin{definition}[Standard IV~\cite{bowden1990instrumental,hernan2006instruments}]
\label{def:standardIV}
A variable $S$ is said to be a standard IV w.r.t. the pair of variables $(W, Y)$, if (i) $S$ has a causal effect on $W$, (ii) $S$ affects $Y$ only through $W$, and (iii) $S$ does not share common causes with $Y$. 
\end{definition}

As an example, in Fig.~\ref{fig:exampleIV_AIV} (a), $S$ is a standard IV w.r.t. the pair of variables $(W, Y)$ since $S$ is a direct cause of $W$, all of its effect on $Y$ is through $W$, and $S$ and $Y$ do not share common causes. 

When a standard IV is given, under the assumption that the data is generated by a linear structural equation~\cite{angrist1996,robins1997causal,vanderweele2011new}, the $\ace(W, Y)$ can be calculated as $\sigma_{sy}/\sigma_{sw}$, where $\sigma_{sy}$ and $\sigma_{sw}$ are the regression coefficients of $S$ when regressing $Y$ on $S$ and when regressing $Y$  on $W$, respectively. The two stage least square (TSLS) regression is one of the most commonly used IV estimators~\cite{angrist1995two,angrist1996,imbens2014instrumental}. The non-linear or non-parametric implementations of the IV estimators for calculating $\sigma_{sy}$ and $\sigma_{sw}$ can be found in~\cite{hartford2017deep,singh2019kernel,sjolander2019instrumental,bennett2019deep}. 
Data-driven causal effect estimation methods do not assume knowing IV. Thus the survey does not review the standard IV methods requiring a known standard IV. The reviews on the standard IV methods can be found in~\cite{bowden1990instrumental,hernan2006instruments,martens2006instrumental,guo2020survey}. 

A standard IV is difficult to find because of its  stringent conditions. Ancestral IV (AIV)~\cite{van2015efficiently} defined below relaxes some of the conditions of a standard IV. AIV is a practical definition of Conditional IV (CIV) and is used in data-driven causal effect estimation. More explanation is provided late.

\begin{definition}[Ancestral IV (AIV)~\cite{van2015efficiently}]
	\label{def:AIV}
	 A variable $S\in \mathbf{X}$ is said to be an AIV \wrt the ordered pair of variables $(W, Y)$, if there exists a set of measured variables $\mathbf{Z}\subseteq\mathbf{X}\setminus \{S\}$ such that (i) $S\nci_{d} W\mid\mathbf{Z}$, (ii) $S\ci_{d} Y\mid\mathbf{Z}$ in $\mathcal{G}_{\underline{W}}$ where $\mathcal{G}_{\underline{W}}$ is the DAG obtained by removing $W\rightarrow Y$ from $\mathcal{G}$, and (iii) $\mathbf{Z} \subseteq (An(Y)\cup An(S))$ and $\forall Z\in\mathbf{Z}$, $Z\notin De(Y)$. In this case $\mathbf{Z}$ is said to instrumentalise $S$.
\end{definition}

An example  {of} AIV is provided in Fig.~\ref{fig:exampleIV_AIV} (b). We see that $S\rightarrow X_3 \rightarrow Y$ violates requirement (ii) of a standard IV, and $S\leftarrow X_1 \rightarrow X_2 \rightarrow Y$ violates requirement (iii) of a standard IV. The set of variables $\mathbf{Z} = \{X_1, X_3\}$ instrumentalises $S$ to be an AIV since $S\nci_{d} W\mid\mathbf{Z}$ in the DAG, $S\ci_{d} Y\mid\mathbf{Z}$ when $W \to Y$ is removed from the DAG, and all elements in $\mathbf{Z}$ are ancestors of $Y$.  

Given an ancestral IV $S$  {instrumentalised} by $\mathbf{Z}$, the causal effect of $W$ on $Y$, $\ace(W, Y)$, can be estimated by $\sigma_{sy\mathbf{z}}/\sigma_{sw\mathbf{z}}$, where $\sigma_{sy\mathbf{z}}$ and $\sigma_{sw\mathbf{z}}$ are the regression coefficient of $S$ when $Y$ is regressed on $S$ and $\mathbf{Z}$, and the regression coefficient of $S$ when $Y$ is regressed on $W$ and $\mathbf{Z}$, respectively.  

An AIV is a CIV (Conditional IV)~\cite{brito2002generalized}, except that an AIV has the extra requirement that $\mathbf{Z} \subseteq (An(Y)\cup An(S))$. When a causal graph is not given and is to be discovered from data, a CIV may produce misleading results since it may be a variable that is uncorrelated with $W$~\cite{van2015efficiently}. For example, in the DAG with $X_1 \leftarrow U_1  \rightarrow X_2 \leftarrow U_2 \rightarrow W \rightarrow Y$ and $W\leftarrow U \rightarrow Y$ where $\{U, U_1, U_2\}$ are latent variables, $X_1$ is a CIV by definition, but is actually independent of $W$. This  could result in a misleading conclusion. An AIV  circumvents such problematic cases. 

 The IV approach has been extensively studied and developed, particularly in the field of economics~\cite{arellano1995another,angrist1996,abadie2003semiparametric}. There are two types of IV estimators: parametric and nonparametric. For a parametric IV estimator, the linearity and homogeneous causal effects assumptions are required to identify causal effects from data with latent confounders. The TSLS estimator is one of the most well-known parametric estimators~\cite{angrist1995two}. Without these assumptions, even if a valid IV exists, causal effects are non-identifiable in real-world scenarios. A nonparametric IV estimator relaxes these assumptions and can identify causal effects without relying on distribution assumptions and functional form restrictions~\cite{hartford2017deep,wu2022instrumental}. Recently, some nonparametric IV estimators have been developed based on machine learning models, such as kernel-based IV regression~\cite{singh2019kernel}, deep learning-based IV estimator~\cite{hartford2017deep}, and moment conditions-based IV estimator~\cite{bennett2019deep}.

 The front-door criterion, introduced on page 83 of~\cite{pearl2009causality}, is  an alternative graphical criterion for causal effect estimation when there exists a latent confounder affecting  {both $W$ and $Y$}. The front-door criterion can be used to identify a front-door adjustment set, which lies on the causal paths between $W$ and $Y$. The front-door criterion is to address the problem where the causal effect of $W$ on $Y$ is not identifiable by the back-door criterion or its extensions. For example, Fig.~\ref{fig:intro} (b) shows a case of front-door adjustment, in which $\mathbf{Z}$ satisfies the front-door criterion.

 The front-door criterion and the IV approach are used to estimate the effects of $W$ and $Y$ in the presence of latent confounders~\cite{pearl2009causality}. Broadly speaking, the IV approach selects a pretreatment variable as a valid IV,  whereas the front-door criterion identifies a post-treatment variable as an adjustment set, their suitability depends on the specific research question and study design. For data-driven causal effect estimation, pretreatment variables are often assumed since otherwise, the discovery of causal structure around $W$ and $Y$ will be very challenging. So, there is not a data-driven method using the front-door criterion yet. Recently, Bhattacharya and Nabi~\cite{bhattacharya2022testability} used
an auxiliary anchor variable and Verma constraint to test the satisfaction of the front-door criterion and showed an intersection model in which the IV and front-door conditions are satisfied can be tested in data. The results open a door for designing a data-driven solution using the front-door criterion.

 Alternative to IV-based methods for dealing with latent confounders, sensitivity analysis can be employed to obtain bound estimates when latent confounders exist. Sensitivity analysis aims to explore the effect of residual latent confounders in causal effect estimation from observational data~\cite{vanderweele2017sensitivity,robins2000sensitivity}. An early work on sensitivity analysis by Cornfield et al.~\cite{cornfield1959smoking} focused on the relative prevalence of a binary latent factor (such as a gene switching on or off) among smokers and non-smokers and its ability to explain the observed association between smoking and lung cancer. Robin et al.~\cite{robins2000sensitivity} provided a set of sensitivity parameters to capture the difference between the conditional distributions of the outcome of treated and control units. Mattei et al.~\cite{mattei2014identification} proposed an approach to identifying causal effects of a binary treatment when the outcome is missing on a subset of units and nonresponse dependence on the outcome cannot be ruled out. Duarte et al.~\cite{duarte2021automated} presented a novel approach to deriving the bound of causal effects in discrete settings and automating the process of partial identification. Scharfstein et al.~\cite{scharfstein2021semiparametric} used semiparametric theory to obtain    {a nonparametric} efficient influence function of the ACE, for fixed sensitivity parameters. The sensitivity analysis methods are not strictly graphical causal modelling based, so in this survey, we do not cover the details of these methods. For those who are interested in sensitivity analysis methods, please refer to the literature in~\cite{christopher2002identification,tortorelli1994design,ding2016sensitivity}.

\section{Challenges for Data-Driven Causal Effect Estimation}
\label{sec:challenges}
In many applications, a DAG or MAG is unknown and we need to estimate causal effects directly from data. An intuitive approach is to recover the causal graph from data and subsequently using it to identify an adjustment set or an AIV  along with its conditioning set for causal effect estimation. However, this is not straightforward and data-driven causal effect estimation faces the following major challenges.   

\subsection{Latent Variables}
\label{subsec:latentvariables}
Latent variables exist in practical scenarios, because in a system some variables may be unmeasurable, or may not measured during data collection. In this survey, we consider that a latent variable is the unmeasured common cause of a pair of measured variables as stated in~\cite{pearl2009causality,spirtes2000causation,spirtes2003causal,richardson2002ancestral}. We also assume that there are no unmeasured selection variables due to selection bias cannot be addressed  solely by confounding adjustment~\cite{van2014constructing,bareinboim2014recovering,pearl2009causality,perkovic2018complete}. When there are latent variables in a system, the causal sufficiency assumption is violated since the assumption requires the common causes of every pair of measured variables in this system are also measured. When the causal sufficiency does not hold, the causal effect estimation from such system becomes challenging and difficult~\cite{spirtes2010introduction,pearl2009causal}. Latent variables mainly cause two difficulties in causal effect estimation  discussed below. 

\begin{figure}[t]
	\centering
	\includegraphics[scale=0.33]{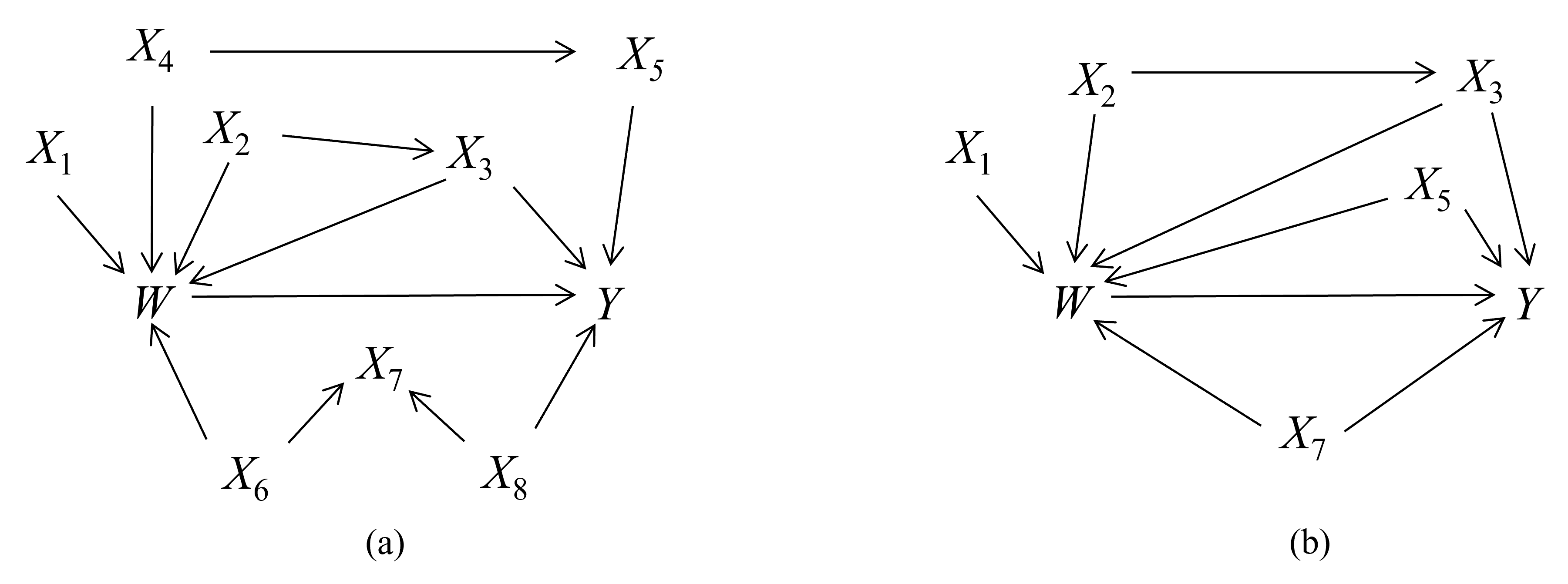}
	\caption{An exemplary DAG (a) and learned DAG (b) from data. Note that in the data,  $X_4, X_6, X_8$ are unmeasured.}
	\label{fig:DAGwitLatentvariables}
\end{figure}

The first difficulty is that latent variables introduce bias in the estimation of causal effects.
For example, assume that the system modelled by Fig.~\ref{fig:exampleDAG_MAG} (b) (\ie the underlying data generation mechanism is a causal MAG) contains three latent variables $X_4$, $X_6$ and $X_8$. If we do not consider the latent variables, and still try to learn a DAG from data, we  will obtain a DAG as shown in Fig.~\ref{fig:DAGwitLatentvariables} (b).  When using the learned DAG in Fig.~\ref{fig:DAGwitLatentvariables} (b),   the estimated  $\ace(W, Y)$ will be biased since based on the learned DAG $X_7$ is on the back-door path between $W$ and $Y$ and thus is included the adjustment set. However, the inclusion of $X_7$ in an adjustment set introduces a spurious association between $W$ and $Y$ because $X_7$ is a collider in Fig.~\ref{fig:DAGwitLatentvariables} (a). This is the well-known ``$M$-bias''~\cite{greene2003econometric,pearl2009myth}.

The second difficulty is that latent variables make causal effect impossible to estimate using the confounding adjustment approach when they are common (direct or indirect) causes of $W$ and $Y$. In this case, the unconfoundedness assumption used by confounding adjustment is violated~\cite{imbens2015causal,rubin1974estimating,rubin2007design,guo2020survey}. Causal effect estimation in this case can be done using the instrumental variable (IV) approach with a standard IV or an ancestral IV (AIV). However, normally an IV needs to be identified by domain knowledge~\cite{martens2006instrumental,van2015efficiently,silva2017learning,sjolander2019instrumental,cheng2022ancestral}, and it is difficult to identify IVs from data. There are only a few data-driven methods available for finding IVs from data~\cite{chu2001semi,kuroki2005instrumental,cheng2022discovering}, but they all require strong assumptions, such as, two variables or  half of the variables in a system are standard IVs or AIVs.
  
From the above discussion, we also know that whether there is a latent confounder between $W$ and $Y$ determines the choice between the two totally different approaches (confounding adjustment and IV approach) for causal effect estimation. In the following, we introduce a graphical criterion for determining  the visibility of $W \rightarrow Y$  {or whether there is not} a latent variable between $W$ and $Y$ in a MAG or a PAG.

\begin{definition}[Visibility~\cite{zhang2008causal}] 
	\label{def:visibility}
	A directed edge $W \rightarrow Y$ in a MAG (or a PAG) is visible if there is a node $X_{i}$ not adjacent to $Y$, such that either there is an edge between $X_{i}$ and $W$ that is into $W$, or there is a collider path between $X_{i}$ and $W$ that is into $W$ and every node on this path is a parent of $Y$. Otherwise, $W \rightarrow Y$ is said to be invisible.  
\end{definition}

\begin{figure}[t]
	\centering
	\includegraphics[scale=0.31]{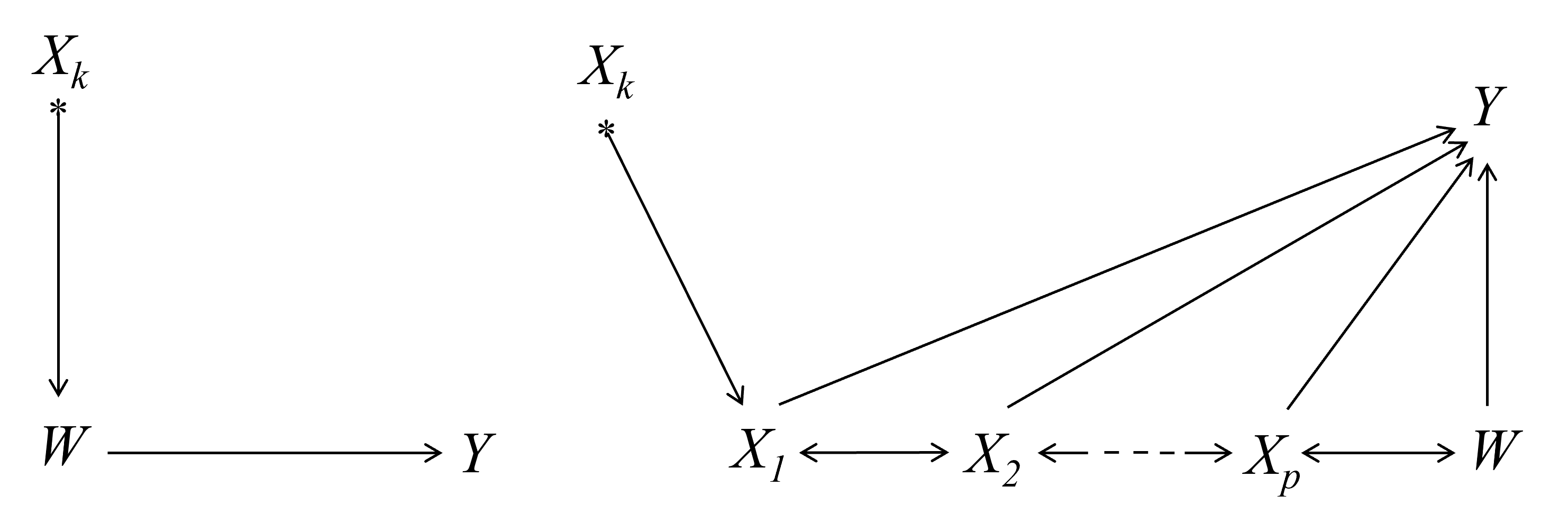}
	\caption{Two possible configurations of the visible edge $W\rightarrow Y$. Note that $X_k$ and $Y$ are non-adjacent.}
	\label{fig:visibleedge}
\end{figure}

There are two cases in a MAG (or a PAG) as illustrated in Fig.~\ref{fig:visibleedge}  in which $W \rightarrow Y$ is visible. When $W \rightarrow Y$ is invisible, there is a latent confounder affecting $W$ and $Y$, and the causal effect of $W$ on $Y$ is non-identifiable~\cite{pearl2009causal,zhang2008causal}, \ie confounding adjustment is not able to recover an unbiased causal effect. When this is a valid IV in the underlying causal DAG, the IV approach can be used to obtain an unbiased causal effect. The details of data-driven methods are introduced in Section~\ref{sec:DDCestimation}.

\subsection{Uncertainty in Causal Structure Learning}
\label{subsec:markoveqdags}
In real-world applications, the underlying causal graphs are rarely available and have to be recovered from data~\cite{spirtes2000causation,witte2019covariate}. Thus, causal structure learning algorithms play a critical role in causal effect estimation from observational data when the causal DAGs/MAGs are unknown~\cite{maathuis2009estimating,maathuis2010predicting,malinsky2017estimating}.  A causal structure learning algorithm, such as PC algorithm~\cite{spirtes2000causation} or RFCI algorithm~\cite{colombo2012learning}, learns from data a CPDAG or PAG, i.e. the output of such an algorithm is a CPDAG or PAG. More details on causal structure learning algorithms can be found in the surveys~\cite{yu2021unified,vowels2021d,glymour2019review,nogueira2022methods}.

The uncertainty of the causal structures learned from data leads to uncertainty in causal effect estimation from observational data. We demonstrate this with an example in Fig.~\ref{fig:examplecpdag_dags}, where the Markov equivalence class of the DAGs is represented by the CPDAG in the second panel. The causal effect of $W$ on $Y$ based on DAG$_3$ is different from the causal effect based on DAG$_1$ since DAG$_3$ has two causal paths from $W$ to $Y$, while there is only one causal path $W\rightarrow Y$ in DAG$_1$. A real-world problem is more complex than this toy example and has more variables. The number of causal graphs in a Markov equivalence class could increase exponentially with the number of variables~\cite{cheng2020causal,cheng2020local}. Thus there may be a large number of estimated values for a causal effect from one dataset, leading to uncertainty in causal effect estimation.  

Two approaches have been developed so far to deal with the non-uniqueness of causal structure learning~\cite{van2014constructing,perkovic2018complete}. The first approach provides a bound estimation of a causal effect, i.e. a  multiset of causal effects estimated based on the DAGs or MAGs enumerated from the CPDAG or PAG learned from data. The second approach produces a unique estimation of a causal effect. It has been shown that when a CPDAG or a PAG is adjustment amendable \wrt the pair of variables $(W, Y)$  as defined below, the causal effect of $W$ on $Y$ can be estimated uniquely if proper adjustment sets \wrt the pair of variables $(W, Y)$ are identified from the DAGs or MAGs encoded in the CPDAG or PAG. 

\begin{definition}[Amenability~\cite{van2014constructing,perkovic2018complete}] 
\label{amenability}
	A CPDAG (or a PAG) $\mathcal{G}$ is adjustment amendable \wrt the pair variables $(W, Y)$ if each possible directed path from $W$ to $Y$ in $\mathcal{G}$ starts with a visible edge out of $W$.
\end{definition}  

For example, in Fig.~\ref{fig:example_pag_mags}, the PAG is adjustment amenable \wrt the pair of variables $(W, Y)$ and the two sets ${X_2, X_5}$ and $\{X_3, X_5\}$ are the proper adjustment sets for all MAGs in the learned PAG. The causal effects from all MAGs in the Markov equivalence class are consistent, and there is one unique estimate.  

Some knowledge is required to determine  {whether} a CPDAG or PAG is adjustment amendable or not. If we know that all the other variables in a system are pretreatment variables~\cite{de2011covariate,entner2013data,haggstrom2018data} \wrt the pair of variables $(W, Y)$ and a variable is correlated with $W$ but not $Y$, then the CPDAG or PAG learned from the data generated by the system is adjustment amendable relative to the pair of variables $(W, Y)$~\cite{cheng2020local}.

In general, the amenability assumption can be satisfied in real-world applications. For example, it is possible that domain experts know there is a causal effect between $W$ and $Y$, and the directed edge between $W$ and $Y$ is the only causal path. In this case, if the edge $W\rightarrow Y$ is visible, and the amenability assumption is  met.

\subsection{Time Complexity}
\label{subsec:timeeff}
The main computational cost for causal effect estimation is to recover causal graphs from data (or search for an adjustment set globally using properties in the underlying causal graph). An optimal search for a causal structure from data is fundamentally NP-complete~\cite{chickering1996learning}, and so is an exhaustive search for an adjustment set directly from data. If using a heuristic strategy for global causal structure learning, the optimal results may not be returned.

Alternatively, local causal structure learning~\cite{aliferis2010local,yu2021unified} is faster and can handle hundreds of variables. Local causal structure learning algorithms aim to recover the local causal structure around a given variable instead of a complete causal DAG. In general, the local causal structure of a variable (referred to as $X$) contains the sets of parent nodes of the variable $Pa(X)$ and the child nodes of the variable $Ch(X)$. Some methods have been developed to search for an adjustment set in the local causal structure around $W$ and $Y$~\cite{perkovic2017interpreting,fang2020ida,cheng2020local,cheng2020causal}.

\section{Data-Driven Causal Effect Estimation Methods}
 \label{sec:DDCestimation}
 In this section, we will review the details of the existing data-driven causal effect estimation methods based on graphical causal modelling. Especially, we will discuss the assumptions, and approaches taken by the methods for addressing the three challenges presented in Section~\ref{sec:challenges}. 
 
\subsection{Methods Classification}
\label{subsec:methodsclass}
  Table~\ref{tab:methods} presents our classification of existing data-driven methods. We classify the existing data-driven causal effect estimation methods into three categories based on their approaches to handling uncertainty in structure learning, their time complexities, and whether they consider the presence of a latent variable between $W$ and $Y$.

 \begin{table}[ht]
 	\centering
 	\footnotesize
 	\caption{A summary of data-driven causal effect estimation methods. Assumptions made by the methods are in the parentheses following the names of the methods. The implementation information (if available) of the methods is provided in the bottom section of the table. The superscript of a method is the index of the implementation details of the method in the bottom section. } 
 	\begin{tabular}{|l|c|c|c|c|}
 		\hline                            
 		\textbf{Uncertainty in} & \textbf{Time}    &  \multicolumn{3}{c|}{\textbf{Latent confounders}}   \\ \cline{3-5} 
 		\textbf{structure learning} & \textbf{complexity}     &  \textbf{Not considered} & \multicolumn{2}{c|}{\textbf{Considered}}   \\ \cline{4-5}
 		 & &    &  \textbf{Not between $(W, Y)$} & \textbf{Between $(W, Y)$} \\ \hline
 		\textbf{Bound estimation} &       \textbf{Global} &IDA$^{1}$   & LV-IDA$^{2}$   &   IV.tetrad$^{9}$  (two valid CIVs)  \\
 		&                             & Semi-local IDA$^{1}$   (fixed edges) & CE-SAT &   AIV.GT$^{10}$ (two valid AIVs)  \\	
 		&                             & DIDA \& DIDN (some fixed edges)  &  &  \\\hline
 		\textbf{Bound estimation} &            \textbf{Local} &  &  DICE (pretreatment)  &    \\		\hline	             
 		\textbf{Unique estimation} &                            \textbf{Global} & & EHS$^{5}$  (pretreatment)  & sisVIVE$^{7}$  (a half of valid IVs)\\
 		&                             & & GAC$^{1}$  (amenability)  &  modeIV$^{8}$ (Modal validity) \\
 		&             & & DAVS$^{6}$  (amenability)  &    $AIViP$ (a given AIV) \\	 \hline 	
 		\textbf{Unique estimation} &                            \textbf{Local} & CovSel$^{3}$  \& CovSelHigh$^{4}$  (pretreatment) &  CEELS (pretreatment) & \\	\hline\hline
 		\textbf{Packages} & \textbf{Language}    &  \multicolumn{3}{c|}{\textbf{URL}}   \\ \hline 
 		1. pcalg & R &  \multicolumn{3}{l|}{https://cran.r-project.org/web/packages/pcalg/index.html} \\  
 		2. LV-IDA	&   R  & \multicolumn{3}{l|}{https://github.com/dmalinsk/lv-ida}   \\ 
 		3.	CovSel	& R & \multicolumn{3}{l|}{https://cran.r-project.org/web/packages/CovSel/index.html}  \\  
 		4. CovSelHigh & R & \multicolumn{3}{l|}{https://cran.r-project.org/web/packages/CovSelHigh/index.html} \\  	
 	   5. EHS  & Matlab & \multicolumn{3}{l|}{https://sites.google.com/site/dorisentner/publications/CovariateSelection}  \\
 	    6. DAVS   & R & \multicolumn{3}{l|}{https://github.com/chengdb2016/DAVS}  \\ 
 	    7. sisVIVE &  R &  \multicolumn{3}{l|}{https://cran.r-project.org/web/packages/sisVIVE/index.html}    \\ 
 	    8.  modeIV & Python & \multicolumn{3}{l|}{https://github.com/jhartford/ModeIV-public} \\
 	    9. IV.tetrad &  R &  \multicolumn{3}{l|}{http://www.homepages.ucl.ac.uk/~ucgtrbd/code/iv$\_$discovery}   \\
 	    10. AIV.GT & R & \multicolumn{3}{l|}{https://github.com/chengdb2016/AIV.GT} \\  \hline
 	\end{tabular}
 	\label{tab:methods}
 \end{table}
 
 A data-driven causal effect  method may produce multiple estimated values for a causal effect due to the uncertainty in structure learning. Although these estimated values may contain the unbiased estimation of the causal effect, it is not clear which of them is the unbiased estimation. In this survey, we classify the existing causal effect estimation methods into two categories based on how they address this uncertainty: \textbf{Bound estimation} methods, which produce multiple estimated values of a causal effect, and \textbf{Unique estimation} methods, which produce a unique (single) estimated value of causal effect.

 Since there is no known causal graph, data-driven methods need to recover the underlying causal graph from observational data using a causal discovery method, which can be a global structure learning method or a local structure learning method~\cite{aliferis2010local,glymour2019review}. Therefore, we classify these methods as either \textbf{Global} or \textbf{Local} based on the causal discovery method used. Note that the former uses the global search, resulting in higher time complexity, while the latter uses the local search, which has lower time complexity.

 Regarding the challenge of latent variables, we classify existing methods into two categories: those that do not consider latent confounders and those  that do. For the methods that consider latent confounders (\ie causal sufficiency does not hold), we further categorise them as either \textbf{Not between} $(W, Y)$, which consider latent confounders that affect either $W$ or $Y$ but not both (\ie $W\rightarrow Y$ is visible in the corresponding MAG or PAG), or \textbf{Between} $(W, Y)$, which consider latent confounders that affect both $W$ and $Y$, such as $W\leftarrow U\rightarrow Y$ in the causal DAG (\ie $W\rightarrow Y$ is invisible in the corresponding MAG or PAG).

\begin{figure}[t]
	\centering
	\includegraphics[scale=0.5]{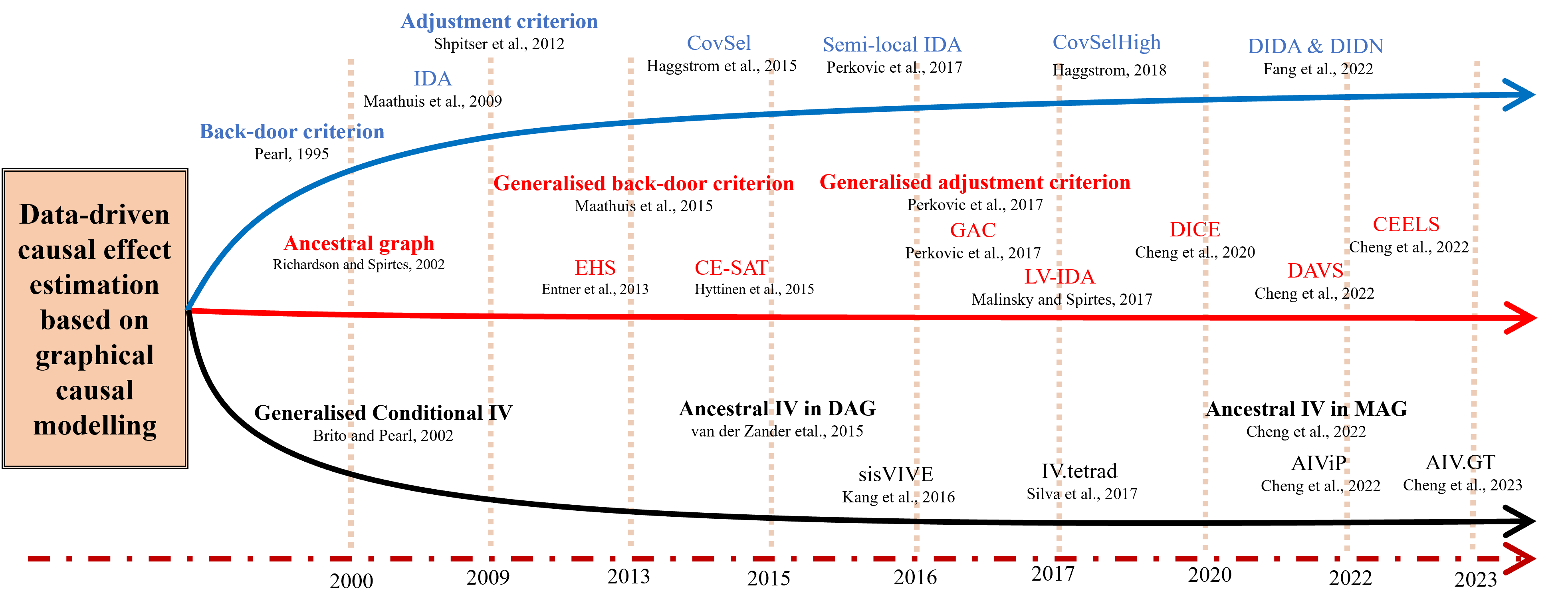}
	\caption{The key milestones of the data-driven algorithms/methods reviewed in this survey for estimating the causal effects from observational data and graphical causal modelling theories or criteria supporting the algorithms. The blue line indicates the timeline of the criteria/methods without considering latent confounders. The black line indicates the timeline of the concepts/criteria/methods considering the latent confounders that do not affect both $W$ and $Y$. The red line represents the timeline of concepts/methods considering the latent confounders that affect both $W$ and $Y$. Names in boldface font indicate theories or concepts and names in normal font indicate methods.}
	\label{fig:timelines}
\end{figure}

In the bottom section of Table~\ref{tab:methods}, we provide information on accessing the software packages for the methods that have been made publicly available online.  Furthermore, Fig.~\ref{fig:timelines} summarises the key milestones of data-driven causal effect estimation based on graphical causal modelling and graphical causal modelling theories supporting the algorithms. The theories were developed in the 1990s and early 2000s, while the methods largely emerged in the past ten years. The number of methods remains small relevant to the importance of causal effect estimation. There is a need for more methods to advance this important field.

 In the following subsections, we will review the details of the data-driven methods following the divisions in Table~\ref{tab:methods}, while using latent variables as the main storyline to structure the review.
  
 \subsection{Methods Considering No Latent Variables}
 \label{subsec:nolatentvariables}
 Many data-driven methods based on graphical causal models have been developed to estimate causal effects from data without considering latent variables. These  methods can be  {categorised} into two types: IDA and its extensions, and data-driven adjustment set discovery methods. In this section, we will review the details of these two types of methods.  
 
 \subsubsection{IDA and Its Extensions}	
 \label{subsec:ida}  
 IDA (\underline{I}ntervention \emph{do-calculus} when the \underline{D}AG is \underline{A}bsent) was proposed by Maathuis et al.~\cite{maathuis2009estimating}. IDA assumes a Gaussian (causal) Bayesian network model over the measured variables $\mathbf{X}$. There are two versions of IDA algorithms, the global version (referred to as the basic IDA method) and the local version (referred to as the IDA algorithm).  The basic IDA algorithm learns a CPDAG using the PC algorithm~\cite{spirtes2000causation}, and enumerates all Markov equivalent DAGs encoded in the learned CPDAG. Finally, for each enumerated DAG, IDA estimates the causal effect of $W$ on $Y$ by adjusting on $Pa(W)$ if there is a causal path between $W$ and $Y$ since $Pa(W)$ blocks all back-door paths between $W$ and $Y$~\cite{pearl2009causality}. IDA returns a multiset of causal effects that correspond to the enumerated DAGs. 
 
   {The CPDAG could potentially contain an excessive number of DAGs, making the enumeration process time-consuming. Moreover}, a number of DAGs encoded in the CPDAG may have the same adjustment set for estimating the causal effect of $W$ on $Y$, so it is not necessary to enumerate all possible DAGs for causal effect estimation. For instance, in Fig.~\ref{fig:examplecpdag_dags}, DAG$_1$ and DAG$_2$ have the same adjustment set $\mathbf{Z}=\{X_1\}$. To improve the computational efficiency, the IDA algorithm~\cite{maathuis2009estimating} has used a local criterion for finding a set of all possible adjustment sets locally, \ie the set of adjacent nodes of $W$ (referred to as $Adj(W)$) that contains all possible parents of $W$. For each subset of $Adj(W)$, IDA checks the \textit{local validity} of the orientation around $W$, \ie no new collider around $W$. Hence, the IDA can be used for fast estimation of causal effects without enumerating all DAGs. For example, the learned CPDAG in Fig.~\ref{fig:examplecpdag_dags}, $Adj(W)={X_1}$ and the IDA only need to check two sets $\emptyset$ and $\{X_1\}$ with respect to the locally valid the orientation. Both sets are locally valid and are considered possible adjustment sets. In this example, the possible causal effects can be obtained by adjusting for $\emptyset$ and $\{X_1\}$, respectively. The time efficiency of IDA relies largely on the PC algorithm~\cite{spirtes2000causation} for discovering a CPDAG from observational data, which fundamentally is not scalable to a large number of variables and only when the underlying DAG is very sparse, IDA can deal with high dimensional data. 
 
 Using the domain knowledge about some edge directions  {can substantially diminish the uncertainty in} causal effect estimations from observational data. The work in a biological application using IDA~\cite{zhang2016predicting} has shown that the uncertainty of the bound estimation of the causal effect can be reduced significantly by using known causal edges as constraints for CPDAG learning. 
 
 Perkovi{\'c} \etal~\cite{perkovic2017interpreting} proposed a semi-local IDA algorithm to produce a bound estimation of a causal effect when background knowledge (\ie some directed edge orientation information) is available. The algorithm uses maximally oriented partially DAGs (maximal PDAGs) to represent CPDAGs augmented with some known edges with or without direction information. Semi-local IDA conducts a local search to find all possible parent sets of a treatment variable in a maximal PDAG. In the process, the Meek orientation rules~\cite{meek1995causal} are used to check the validity of the parents of the treatment variable.  The check is global, and hence the algorithm is called semi-local. The multiset of causal effects is estimated by adjusting for each possible parent set. The advantage of semi-IDA over maximal PDAGs is the incorporation of more edge information (\ie orientation
 information) than IDA over CPDAGs. Because of the extra check on the validity of possible parents using the Meek orientation rules~\cite{meek1995causal}, the time complexity of semi-local IDA is slightly higher than IDA.  
 
 Fang and He proposed an improved semi-local IDA algorithm, \underline{IDA} with some \underline{D}irect causal information (DIDA)~\cite{fang2020ida}, to improve the performance and time efficiency of semi-local IDA. DIDA also uses a maximal PDAG to find possible parent sets of a treatment variable, but DIDA employs a newly designed criterion to check the validity of parents using the variables adjacent to the treatment. DIDA improves the performance of semi-local IDA and is faster than IDA. Fang and He have also proposed an \underline{IDA} algorithm using the knowledge of \underline{N}on-ancestral information, referred to as NIDA~\cite{fang2020ida}, for causal effect estimation. NIDA has a similar time efficiency as DIDA. Both provide bound estimates of causal effects.   
 
IDA, semi-local IDA, DIDA and NIDA all conduct a local search for finding a possible parent set of the treatment, but we classify them as global algorithms in Table~\ref{tab:methods}.   This is because they learn a complete CPDAG from data. These methods do not consider latent confounders.  
  
 \subsubsection{Data-Driven Adjustment Set Discovery}
 \label{subsubsec:BayesiannCI}
 In practice, there are two main criteria for finding an adjustment set to remove confounding bias in the estimation of the causal effect of $W$ on $Y$: the common causes of $W$ and $Y$~\cite{glymour2008methodological}, and either all causes of $W$ or all causes of $Y$~\cite{vanderweele2011new}. Both criteria need domain knowledge since the causes of $W$ or $Y$ are not generally discoverable from data. When there are no latent variables, the set of parents of a variable in a learned DAG are their causes, but recovering a DAG is impossible due to the uncertainty in causal structure learning~\cite{spirtes2000causation,maathuis2009estimating,spirtes2010introduction} as introduced in Section~\ref{sec:challenges}. 
 
 With the pretreatment assumption, \ie all measured variables excluding the pair of variables $(W, Y)$ are not descendant nodes of $W$ or $Y$, the parents of $W$ and $Y$ can be learned from data since the PC (parent and children) set contains parents only. Under the pretreatment assumption, De Luna et al.~\cite{de2011covariate} have proposed a set of criteria for identifying adjustment sets, including the parent set of $W$, the parent set of $Y$,  the parent set of $W$ excluding those having no paths linking $Y$, the parent set of $Y$ excluding those having no paths linking $W$ except via $W$, the common parents of $W$ and $Y$, and the union of $W$'s parent set and $Y$'s parent set. H{\"a}ggstr{\"o}m \etal~\cite{haggstrom2015covsel} implemented these criteria in the $R$ package \emph{CovSel} by using marginal coordinate hypothesis testing (for continuous covariates and outcome) and kernel smoothing (for continuous/discrete covariates and outcome). However, the implementation is computationally expensive and has a prohibitive time consumption for high-dimensional datasets. H{\"a}ggstr{\"o}m et al.~\cite{haggstrom2018data} thus presented an improved implementation, \textit{CovSelHigh}, which uses the local structure learning algorithm, Max-Min Parents and Children (MMPC) and Max-Min Hill-Climbing (MMHC)~\cite{tsamardinos2006max} and is more scalable to the high-dimensional datasets than the implementations in \emph{CovSel}.  
 
 Another area of research in causal inference focuses on using asymptotic variance to identify valid or optimal adjustment sets for estimating average causal effects with causal linear models. Several recent works have explored this approach, including Rotnitzky and Smucler~\cite{rotnitzky2020efficient}, Witte \etal~\cite{witte2020efficient}, and Henckel \etal~\cite{henckel2022graphical}.
Henckel \etal~\cite{henckel2022graphical} proposed a graphical criterion to compare the asymptotic variances of different valid adjustment sets for causal linear models. This criterion can be used to develop two graphical criteria: a variance-decreasing pruning procedure for any given valid adjustment set, and a graphical characterisation of a valid adjustment set that provides the optimal asymptotic variance among all valid adjustment sets. These results are applicable to DAGs, CPDAGs and maximal PDAGs. Building on this work, Rotnitzky and Smucler~\cite{rotnitzky2020efficient} discussed several new graphical criteria for  {nonparametric} causal graphical models when treatment effects were estimated using  nonparametrically adjusted estimators of the interventional means. They also proposed a new graphical criterion for discovering the optimal adjustment set among the minimal adjustment sets.
Witte \etal~\cite{witte2020efficient} concluded that the optimal valid adjustment set (O-set) was characterised graphically as the set of parent nodes of outcome $Y$ that yielded the smallest asymptotic variance. These approaches have the potential to improve the accuracy and efficiency of causal effect estimation in various settings.

 \subsection{Methods Considering Latent Variables Which Are Not Between $W$ and $Y$}
 \label{subsec:latentvariables}
 When there are latent variables in data, an ancestral graph is usually used to represent the causal relationships among measured variables. From the data, only an equivalence class of MAGs, \ie a PAG, can be recovered. The generalised back-door criterion (or generalised adjustment criterion) can be used in a MAG (PAG) to determine a proper adjustment set when there exists such a proper adjustment set~\cite{maathuis2015generalized,perkovic2018complete}. In this section, we will review the data-driven methods considering latent variables and finding proper adjustment sets based on the graphical criteria for the causal effect estimation.

 \subsubsection{LV-IDA}
 \label{subsec:lvida}
 Similar to IDA, LV-IDA  (latent variable IDA) by Malinsky and Spirtes~\cite{malinsky2017estimating} considers latent variables and is based on the generalised back-door criterion. There are also two versions of LV-IDA algorithms, the global version (referred to as the global LV-IDA algorithm) and the local version (referred to as the LV-IDA algorithm). 
  
 The global LV-IDA first recovers a PAG from data with latent variables, using a structure learning algorithm such as FCI~\cite{spirtes2000causation}, or RFCI~\cite{colombo2012learning}. Then it enumerates the MAGs encoded in the PAG. In each of the MAGs, LV-IDA searches for a proper adjustment set for each enumerated MAG according to the generalised back-door criterion and provides an estimate of the causal effect using the adjustment set found in the enumerated MAG. In the end, a multiset of causal effects is returned by LV-IDA. 
 
 Enumerating all MAGs encoded by a PAG is computationally expensive, and becomes impossible for  moderately sized PAGs (the number of nodes is with up to $20$) even using the ZML algorithm~\cite{zhang2008completeness} to reduce the search space. To address the efficiency problem, a local search algorithm has been proposed by using the $possible$-$D$-$SEP(W, Y, \mathcal{G})$ set (\ie the possible $d$-separated set relative to $(W, Y)$ in $\mathcal{G}$ and abbreviated as $pds(W, Y, \mathcal{G})$)~\cite{spirtes2000causation,colombo2012learning,malinsky2017estimating} defined below. 
 
 \begin{definition}[$pds(W, Y, \mathcal{G})$]
 	A node $X$ is in the set $pds(W, Y, \mathcal{G})$ if and only if there is a path between $X$ and $W$ in $\mathcal{G}$ such that for every sub-path $\left\langle X_i, X_k, X_j\right\rangle$ on this path either $X_k$ is a collider, or $\left\langle X_i, X_k, X_j\right\rangle$ is a triangle in $\mathcal{G}$ (i.e. each pair of variables in $\left\langle X_i, X_k, X_j\right\rangle$ are adjacent).
 \end{definition}

 It is worth mentioning that $pds(W, Y, \mathcal{G})$ reduces the size of the original PAG (\ie a sub-graph of the learned PAG) since only using this set would be sufficient to check the conditions of generalised back-door criterion \wrt the pair of variables $(W, Y)$. 
 
 Instead of enumerating all the MAGs from the complete PAG learned from data, the LV-IDA enumerates the MAGs represented by a sub-graph of the complete PAG constructed based on the set $pds(W, Y, \mathcal{G})$.  For each of the MAGs enumerated from the subgraph, the generalised back-door criterion is used to determine a proper adjustment set, and similar to LV-IDA, the final output of the local version of LV-IDA is a multiset of estimated causal effects.
 
 LV-IDA has  demonstrated its ability to handle high-dimensional data with thousands of variables when the underlying causal MAG is sparse. The performance of both versions LV-IDA relies on the accuracy of the learned PAGs and the locations of latent variables. Learning an accurate PAG from data with latent confounders is still a challenging problem in causal discovery~\cite{zhang2008completeness,spirtes2010introduction,glymour2019review,cheng2020causal,vowels2021d}.

 \subsubsection{Causal Effect Estimation Based on SAT Solvers (CE-SAT)}
 \label{subsubsec:CE-SAT}
 Hyttinen \etal~\cite{hyttinen2015calculus} developed a data-driven method, \underline{C}ausal effect \underline{E}stimation based on \underline{SAT}isfiability (SAT) solver (CE-SAT), for causal effect estimation from data with latent variables. CE-SAT requires learning a PAG from data. Instead of enumerating all MAGs encoded in the learned PAG, CE-SAT works on subsets of a Markov equivalence class (\ie a sub-equivalence class) with different sets of constraint solvers~\cite{hyttinen2014constraint}. Some MAGs in the equivalent class satisfy the same set of $m$-separation/$m$-connection constraints\footnote{Note that in the original paper of CE-SAT~\cite{hyttinen2015calculus}, the terms $d$-separation and $d$-connection were used. However, we think it is more appropriate to use the terms $m$-separation and $m$-connection here since CE-SAT requires a PAG as input, rather than a CPDAG.}  
 which guarantee the satisfaction of the same set of do-calculus rules~\cite{pearl2009causality} (a rule set that is a more general criterion of causal effect identifiability than the (generalised) back-door criterion.). When a set of MAGs satisfies the same set of $do$-calculus rules, they have the same do-free expression~\cite{pearl2009causality} for obtaining the causal effect and hence the same estimated causal effect from a dataset based on this set of MAGs. CE-SAT translates $m$-separation constraints into a logical representation, and uses an SAT solver to find a sub-equivalence class satisfying the same do-calculus rules for preventing repeatedly estimating the same causal effect based on the sub-equivalence class. CE-SAT avoids enumerating all MAGs in the PAG. CE-SAT also returns a multiset of causal effects from a dataset but the size of the multiset can be much smaller than that returned by LV-IDA. The process of finding sub-equivalence classes satisfying the same do-calculus rules does not scale well with the number of variables, and it has been shown that CE-SAT only works with datasets  of a dozen of variables.

 \subsubsection{Efficient Bound Estimation with Local Search}
 \label{subsubsec:DICE}
 Cheng et al.~\cite{cheng2020causal} proposed a \underline{D}ata dr\underline{I}ven \underline{C}ausal \underline{E}ffect estimation algorithm (DICE) to achieve efficient causal effect estimation from data with latent variables. Based on the generalised back-door criterion, the authors have proved that there exists at least an adjustment set in the local causal structures around $W$ and $Y$ (i.e. the adjacent nodes of $W$ and $Y$ in the PAG)  under the assumption of pretreatment variables and the assumption of no latent confounders between $W$ and $Y$. With this conclusion, there is no need to learn the complete PAG but to find the adjacent nodes of $W$ and $Y$ locally, which significantly reduces the time complexity. DICE employs a local causal structure learning method, such as PC-Select~\cite{kalisch2012causal}, to recover the local structures around $W$ and $Y$, and then enumerates all subsets of the adjacent variables of $W$ and $Y$ that are the possible adjustment sets. DICE obtains the set of possible causal effects by adjusting for the possible adjustment sets. DICE is a bound estimator and is faster than LV-IDA and CE-SAT for bound estimation \wrt the causal effect of $W$ on $Y$ due to  its localised search approach.

 \subsubsection{Causal Effect Estimation by the Generalised Adjustment Criterion (GAC)}
 \label{subsec:GAC}
 Perkovi{\'c} et al.~\cite{perkovic2018complete} developed a data-driven method, denoted as GAC, using the generalised adjustment criterion to identify an adjustment set from a given PAG. 
 
 \begin{definition}[Generalised Adjustment Criterion~\cite{perkovic2018complete}]
 	In a given MAG or PAG $\mathcal{G}$, the set $\mathbf{Z}$ satisfies the generalised adjustment criterion w.r.t. $(W, Y)$ if (i). $\mathcal{G}$ is adjustment amenable w.r.t. the pair of variables $(W, Y)$, and (ii). $\mathbf{Z}\cap Forb(W, Y, \mathcal{G})= \emptyset$, and (iii). all definite status non-causal paths between $W$ and $Y$ are blocked (i.e. $m$-separated) by $\mathbf{Z}$ in $\mathcal{G}$.
 \end{definition}  
 
 In the above definition,  $Forb(W, Y, \mathcal{G})$ is the set of all nodes lying on all causal paths between $W$ and $Y$. The first condition, i.e. $\mathcal{G}$ is adjustment amenable means that the causal effect of $W$ on $Y$ is identifiable, and the second condition denotes that $\mathbf{Z}$ does not contain a node lying on any causal path between $W$ and $Y$, \ie $\forall Z\in\mathbf{Z}$, $Z\notin Forb(W, Y, \mathcal{G})$. The third condition is used to read off a set $\mathbf{Z}$ from the given MAG or PAG $\mathcal{G}$ that blocks all definite status non-causal paths between $W$ and $Y$.   
 
 GAC needs to learn a PAG from data using a structure learning algorithm, such as FCI~\cite{spirtes2000causation} or RFCI~\cite{colombo2012learning}. From the input PAG, GAC enumerates $m$-separating sets of the pair $(W, Y)$ from each MAG encoded in the PAG. Then based on the  $m$-separating sets, an adjustment set is identified using the generalised adjustment criterion. The performance of the data-driven GAC algorithm relies on the correctness of the learned PAG. GAC is similar to LV-IDA in time efficiency since both require a PAG learned from data.  
 
 To improve the efficiency of finding an adjustment set, a fast version of the GAC uses the set of all possible ancestors of $W$ and $Y$, excluding the forbidden set as a valid adjustment set that can be obtained in linear time. However, the set of the possible ancestors of $W$ and $Y$ may be large and hence the quality of the estimated causal effect is low. Note that the fast version of the GAC is faster than LV-IDA.

 \subsubsection{The EHS Algorithm}
 \label{subsubsec:threerules}
 Entner, Hoyer, and Spirtes~\cite{entner2013data} proposed to use two inference rules to solve the problem of inferring whether a treatment variable $W$ has a causal effect on the outcome $Y$ of interest; and if it has, determining an adjustment set from data by conducting statistical tests. The two inference rules support data-driven covariate selection for estimating the causal effect of $W$ on $Y$ directly from data with latent variables. We call the two inference rules EHS conditions and the method EHS algorithm, named after the authors' names. EHS is different from the previous methods since it does not need a complete or local causal structure, but finds an adjustment set directly from data by conducting a set of conditional independence tests. The causal effect is unique if an adjustment set can be found from the data.
 
 EHS conditions need the pretreatment variable assumption. EHS algorithm searches for a variable $S$ and a set $\mathbf{Z}$ simultaneously such that (1) $ S\not\!\perp\!\!\!\perp Y\mid \mathbf{Z}$ and  $S\ci Y\mid \mathbf{Z}\cup \{W\}$  or (2) $S\not\!\perp\!\!\!\perp W\mid \mathbf{Z}$ and  $S\ci Y\mid \mathbf{Z}$. If (1) is satisfied, $\mathbf{Z}$ is an adjustment set and is used for causal effect estimation; if (2) is satisfied, the causal effect of $W$ on $Y$ is zero; and if neither is satisfied, we   {do not} know whether the causal effect can be estimated or not from the dataset. The EHS conditions are sound and complete for querying   {the causal effect} between $W$ and $Y$. However, the EHS algorithm~\cite{entner2013data} is extremely inefficient since it globally searches for both $S$ and $\mathbf{Z}$, and the time complexity is $\mathbf{O}(2^{|\mathbf{X}|})$, where $\mathbf{X}$ includes all measured variables in the system except $W$ and $Y$. Hence, the EHS algorithm only works for a small number of variables.

 \subsubsection{GAC Guided Conditional Independent Tests for Causal Effect Estimation}
 \label{subsubsec:davs}
 The reliability of the GAG algorithm depends on the correctness of the PAG learned from data, but the PAG learned may be imprecise because of the high complexity of graph discovery and the uncertainty of edge orientation. On the other hand, EHS makes use of statistical tests that can obtain a more reliable adjustment set than reading an adjustment set from a PAG learned from data. However, EHS requires the pretreatment variable assumption such that it will fail to determine a valid adjustment set when there is a mediation variable between $W$ and $Y$. For example, in a MAG, there is a causal path $W\rightarrow X\rightarrow Y$. In this case, $X$ is a child node of $W$ and 
 EHS will erroneously include $X$ in the adjustment set.  The estimated causal effect by EHS will be biased due to that $X$ is a mediator.  
 
 Cheng \etal~\cite{cheng2022toward} developed the \underline{D}ata-driven \underline{A}djustment \underline{S}earch (DAVS) algorithm for unique causal effect estimation by taking the advantages of GAC and EHS. The authors developed a criterion which uses a conditional independence test to discover an adjustment set after finding candidate adjustment sets based on the generalised adjustment criterion. DAVS does not require the pretreatment variable assumption. The proposed criterion uses a COSO (Cause Or a Spouse of the treatment Only) variable which is not adjacent to $Y$. A COSO variable is often known by domain experts since it is common for domain experts to know a direct cause of the treatment variable and that the direct cause is not a direct cause of $Y$. Even with no such prior knowledge, a COSO variable can be identified from data by conducting statistic tests for adjacencies. A COSO variable is not an IV~\cite{baiocchi2014instrumental,martens2006instrumental,hernan2006instruments} introduced in Definition~\ref{def:standardIV}. A COSO variable is proposed to enable   {a} data-driven confounding adjustment method for causal effect estimation, while an IV is employed to address the latent confounders between $W$ and $Y$ (Introduced in Section~\ref{subsec:latentvariables_wy}). Moreover, a COSO variable can be found via statistical tests in data whereas an IV should be nominated   {using} domain knowledge.

 Following the GAC, an adjustment set is in a possible $m$-separating set (which contains all variables possibly $m$-separating all the paths between $W$ and $Y$), but not in the forbidden set (which contains all possible descendant variables of $W$ lying on the possible direct paths between $W$ and $Y$). Note that ``possible'' here is caused by the uncertainty of the edge orientation in a learned PAG as introduced in Section~\ref{subsec:markoveqdags}. Both possible $m$-separating set and forbidden set can be read from the learned PAG. To determine a unique causal effect of $W$ on $Y$, DAVS determines if a set $\mathbf{Z}$ is an adjustment set $\mathbf{Z}$ by testing if $Q\ci Y|\mathbf{Z}\cup \{W\}$ where $Q$ is a COSO variable. DAVS adopts the \emph{Apriori pruning strategy}~\cite{agrawal1994fast} for searching for the minimal adjustment sets satisfying the test, and then estimates causal effects using the identified minimal adjustment sets. Note that, the criterion using a COSO variable finds an adjustment set when there is a causal relationship between $W$ and $Y$, but it cannot determine whether there is a causal relationship between $W$ and $Y$ (i.e. when an adjustment set is not found, it is unsure that there is no causal relationship between $W$ and $Y$ or the relationship is unidentifiable using data.). DAVS is similar to LV-IDA in time efficiency but much faster than EHS.

 \subsubsection{Local search for efficient discovery of Adjustment Sets}
 To improve the efficiency of EHS and DAVS, Cheng \etal~\cite{cheng2020local} proposed a fully local search algorithm, \underline{C}ausal \underline{E}ffect \underline{E}stimation by \underline{L}ocal \underline{S}earch (CEELS). CEELS makes use of the result from DICE (see Section~\ref{subsubsec:DICE}) that at least one adjustment set exists in the adjacent set of $(W, Y)$ and this ensures that the search for an adjustment set can be done locally. As DAVS,  CEELS also makes use of a COSO variable given by domain experts or learned from data.
 
 Compared to EHS, CEELS searches for a proper adjustment set by using EHS conditions with the following improvements: a COSO variable is used, and the search for an adjustment set $\mathbf{Z}$ is restricted to the adjacent variables of $W$ and $Y$. The search space of CEELS is within the local structure around $W$ and $Y$, and it is significantly smaller than the search space of EHS. CEELS improves DICE by finding a proper adjustment set for obtaining a unique (instead of a bound) estimation of the causal effect of $W$ on $Y$. In addition to the local search, as with DAVS, CEELS employs a frequent pattern mining approach (\ie the \emph{Apriori} pruning)~\cite{agrawal1994fast} to reduce the search space to find the minimal adjustment set. Different from DAVS, CEELS is a pure local search for unique causal effect estimation and has a higher efficiency than DAVS. CEELS is significantly faster than EHS and DAVS, and scales well with high-dimensional datasets. CEELS is also faster than LV-IDA. However, CEELS may miss an adjustment set in local search while EHS can find it using the global search, and the work in~\cite{cheng2020local} shows that the missing rate is small in simulation studies.

 \subsection{Methods Considering Latent Variables Affecting Both $W$ and $Y$}
 \label{subsec:latentvariables_wy}
 When there are latent confounders, i.e. latent variables affecting both $W$ and $Y$, the causal effect of $W$ on $Y$ is not identifiable via confounding adjustment. Instrumental variable (IV) approach is a powerful method for inferring the causal effect of $W$ on $Y$ from data with latent confounders even when there exist latent confounders affecting both $W$ and $Y$. If there exists a valid IV, the causal effect of $W$ on $Y$ can be recovered from data with latent confounders. It is challenging to determine an IV directly from data~\cite{pearl1995testability,hernan2006instruments}. Traditionally, a valid IV is nominated based on domain knowledge. However, recently, research has emerged to work towards data-driven methods related to the IV approach. In this section, we review the emerging research on the IV-based approach. 
 
 \subsubsection{sisVIVE and Its Extension}
 \label{subsubsec:sisVIVE}
 The three conditions of a standard IV introduced in Section~\ref{Subsec:IV} are untestable from data, therefore it is very difficult to discover a valid IV from data with latent variables. Instead of discovering a valid IV from data, Kang \etal~\cite{kang2016instrumental} proposed a practical algorithm, called \textit{\underline{s}ome \underline{i}nvalid \underline{s}ome \underline{V}alid \underline{IV} \underline{E}stimator} (sisVIVE), to estimate the causal effects of $W$ on $Y$ from data with latent variables as long as more than 50\% of candidate instruments are valid. The assumption is referred to as majority validation of IVs, \ie at least half of the covariates are valid IVs. sisVIVE is able to obtain an unbiased estimation of causal effect without knowing which covariate is a valid IV, and can be applied in Mendelian randomisation~\cite{kang2016instrumental,hartford2021valid}. However, the estimation may be biased when the assumption of majority validation   {of IVs} is violated. Hence, it would be better to check the majority validation assumption before using sisVIVE. 
 
 An extended version of the sisVIVE algorithm, modeIV, has been developed by Hartford \etal~\cite{hartford2021valid}. ModeIV is a robust IV technique when the model relationship between an IV $\mathbf{S}$ and $Y$, \ie the  conditional expectation $\mathbf{E}(Y\mid W, S)$ is valid and at least half of IVs are valid (\ie model validity holds).  The modeIV allows the estimation of non-linear causal effects from data by employing the recently developed deep learning IV estimator in~\cite{hartford2017deep}. The advantage of modeIV is its ability in removing most of the bias introduced by invalid IVs. When the assumption of model validity is violated, modeIV will not work. 
  
 \subsubsection{IV.tetrad}  A CIV has more relaxed conditions than a standard IV~\cite{brito2002generalized} and allows a variable in the covariate set to be a valid IV conditioning on a set of measured variables. IV.tetrad~\cite{silva2017learning} is the first data-driven CIV-based method for discovering a pair of valid CIVs from data directly. IV.tetrad requires that there exist two valid CIVs, denoted as $S_i$ and $S_j$ in the covariate set and the conditioning sets of $S_i$ and $S_j$ equal to the set of remaining covariates $\mathbf{Z}= \mathbf{X}\setminus\{S_i, S_j\}$ (i.e. the original covariate set $\mathbf{X}$ excluding the pair of CIVs $\{S_i, S_j\}$). Moreover, the assumption of linear non-Gaussian causal model is needed for the validity of IV.tetrad. For the two valid CIVs, we have $\beta_{wy} = \sigma_{s_{i}*y*\mathbf{z}}/\sigma_{s_{i}*w*\mathbf{z}} = \sigma_{s_{j}*y*\mathbf{z}}/\sigma_{s_{j}*w*\mathbf{z}}$, where $\sigma_{s_{i}*y*\mathbf{z}}$ is the causal effect of $S_i$ on $Y$ conditioning on $\mathbf{Z}$ and $\sigma_{s_{i}*w*\mathbf{z}}$ is the causal effect of $S_i$ on $W$ conditioning on $\mathbf{Z}$ ($\sigma_{s_{j}*y*\mathbf{z}}$ and $\sigma_{s_{j}*w*\mathbf{z}}$ are the same.). Thus, the tetrad constraint, i.e. $\sigma_{s_{i}*y*\mathbf{z}}\sigma_{s_{j}*w*\mathbf{z}} - \sigma_{s_{i}*w*\mathbf{z}}\sigma_{s_{j}*y*\mathbf{z}} = 0$ can be obtained. IV.tetrad uses the tetrad constraint to check the validity of a pair of covariates $\{S_i, S_j\}$ in $\mathbf{X}$ conditioning on $\mathbf{Z} = \mathbf{X}\setminus\{S_i, S_j\}$. Note that the tetrad constraint is a necessary condition for discovering a pair of valid CIVs, because a pair of non-valid CIVs can pass the tetrad constraint. It means that the tetrad constraint cannot exactly identify a pair of valid CIVs from data, \ie IV.tetrad provides a bound estimation of causal effects. 
 
 IV.tetrad aims to discover valid CIVs from data directly, and it is the first data-driven algorithm to discover CIVs without domain knowledge. However, IV.tetrad may fail in obtaining an unbiased estimation when the set of remaining covariates $\mathbf{X}\setminus\{S_i, S_j\}$ contains a collider between $S_i$ (or $S_j$) and $Y$, \ie there exists an $M$-bias~\cite{greenland2003quantifying,pearl2009myth}.

 \subsubsection{ {Data-Driven AIV Estimator}} 
As discussed in Section~\ref{Subsec:IV}, a CIV may produce a misleading conclusion~\cite{van2015efficiently} and an AIV remedies this limitation. The graphical criterion given in Definition~\ref{def:AIV} for finding a conditioning set relative to a given AIV needs a complete causal DAG representing the causal relationships of both measured and unmeasured variables~\cite{brito2002generalized,van2015efficiently}. The requirement of a complete causal DAG makes it impossible to discover a conditioning set directly from data \wrt a given AIV. Cheng \etal~\cite{cheng2022ancestral} studied a novel graphical property of an AIV in a MAG to discover from data a proper conditioning set that instrumentalises a given AIV.  {Using} the graphical property, they proposed a data-driven method, \underline{A}ncestral \underline{IV} estimator \underline{i}n \underline{P}AG (\emph{AIViP} for short) to discover a conditioning set $\mathbf{Z}$ in a PAG  that instrumentalises a given AIV $S$ in the underlying causal DAG. \emph{AIViP} employs a causal structure learning method, \ie RFCI~\cite{colombo2012learning} to discover a PAG from data with latent variables, and then constructs a manipulated PAG to discover the conditioning set $\mathbf{Z}$. Finally, \emph{AIViP} uses a TSLS method to calculate the causal effect of $W$ on $Y$ by using the given AIV $S$ and the conditioning set $\mathbf{Z}$. The quality of \emph{AIViP} for finding the correct conditioning sets relies on the quality of the PAG discovered from data with latent confounders.

 \begin{table}[t]
	\centering
	\footnotesize
	\caption{A summary of the properties and characteristics of the data-driven causal effect estimation methods. `Causal sufficiency' refers to Definition~\ref{Causalsufficiency}, `Amenability' refers to Definition~\ref{amenability}, `Pretreatment' denotes the pretreatment variable assumption, `Bound' means that the estimator produces multiset estimated causal effects (`$\times$' indicates a unique estimate), and `Time-complexity' is based on the time complexity of the causal structure learning part of a method (`+' denotes global causal structure learning (exponential time complexity with the number of variables) and `-'  indicates local structure learning (polynomial time complexity with the number of variables)).}
	\begin{tabular}{|c|c|c|c|c|c|c|}
		\hline
		Methods      & Causal sufficiency & Amenability &  Linearity  & Pretreatment  & Bound&  Time-complexity    \\ \hline
		IDA~\cite{maathuis2009estimating}	         &   $\checkmark$      &  $\checkmark$   &    $\checkmark$  &   $\times$       & $\checkmark$    &  $+$           \\ \hline
		Semi-local IDA~\cite{perkovic2017interpreting}  &    $\checkmark$  & $\checkmark$     &   $\checkmark$    &   $\times$ &   $\checkmark$ &  $+$          \\ \hline
		DIDA~\cite{fang2020ida}      &   $\checkmark$    & $\checkmark$     &    $\checkmark$    &   $\times$  & $\checkmark$ &  $+$             \\ \hline
		DIDN ~\cite{fang2020ida}     &   $\checkmark$    &   $\checkmark$   &    $\checkmark$    &   $\times$  & $\checkmark$ &  $+$          \\ \hline
		CovSel~\cite{haggstrom2015covsel}    &  $\checkmark$   &$\checkmark$   &   $\checkmark$  &   $\checkmark$ &   $\times$  &     $-$       \\ \hline
		CovSelHigh~\cite{haggstrom2018data}   & $\checkmark$ &      $\checkmark$  &   $\checkmark$  &  $\checkmark$  &  $\times$ &   $-$      \\ \hline
		LV-IDA~\cite{malinsky2017estimating}     &   $\times$         &  $\checkmark$  &    $\checkmark$    &   $\times$  & $\checkmark$&  $+$              \\ \hline
		CE-SAT~\cite{hyttinen2015calculus}    &    $\times$ &$\checkmark$   &    $\times$ &  $\times$  &   $\checkmark$ &  $+$              \\ \hline
		DICE~\cite{cheng2020causal}         & $\times$  &   $\checkmark$     &  $\times$   &  $\checkmark$   &   $\checkmark$ &      $-$           \\ \hline
		EHS~\cite{entner2013data}          & $\times$  &   $\checkmark$     &  $\times$   & $\checkmark$    &   $\checkmark$& $+$                  \\ \hline
		GAC~\cite{perkovic2018complete}        & $\times$ &$\checkmark$        &   $\times$  &   $\times$ &     $\times$ &     $+$           \\ \hline
		DAVS~\cite{cheng2022toward}     & $\times$  &    $\checkmark$    & $\times$    & $\times$    &  $\times$  &      $+$             \\ \hline
		CEELS~\cite{cheng2020local}               &  $\times$  & $\checkmark$    &   $\times$  &  $\checkmark$  &  $\times$ &    $-$       \\ \hline
		sisVIVE~\cite{kang2016instrumental}      &  $\times$ &     $\times$   &     $\checkmark$ & $\checkmark$  &     $\checkmark$  &     $-$          \\ \hline 
		modeIV~\cite{hartford2021valid}     & $\times$  &      $\times$  & $\times$     &  $\checkmark$   &     $\checkmark$  &  $-$             \\ \hline
		$AIViP$ \cite{cheng2022ancestral}       & $\times$ &$\times$      & $\checkmark$    &$\checkmark$    &   $\times$ &    $+$            \\ \hline
		IV.tetrad~\cite{silva2017learning}     &  $\times$   &    $\times$   & $\checkmark$    &    $\checkmark$  &   $\checkmark$ & $-$        \\ \hline
		AIV.GT~\cite{cheng2022discovering}    & $\times$    &  $\times$      & $\checkmark$    &  $\checkmark$  &    $\checkmark$  &    $-$          \\ \hline
	\end{tabular}
	\label{tab:002}
\end{table}  
 
 \subsubsection{  {AIV Generalised Tetrad}} Cheng \etal~\cite{cheng2022discovering} extended the tetrad constraint~\cite{silva2017learning} to a generalised tetrad condition:  $\sigma_{s_{i}*y*\mathbf{z}_i}\sigma_{s_{j}*w*\mathbf{z}_j} - \sigma_{s_{i}*w*\mathbf{z}_i}\sigma_{s_{j}*y*\mathbf{z}_j} = 0$ where $\mathbf{Z}_i$ and $\mathbf{Z}_j$, the conditioning sets of the pair of AIVs $S_i$ and $S_j$,  $\mathbf{Z}_i \subseteq \mathbf{X}\setminus\{S_i\}$ and $\mathbf{Z}_j \subseteq \mathbf{X}\setminus\{S_j\}$) do not need to be equal. Under the Oracle test, $\mathbf{Z}_i$ and $\mathbf{Z}_j$ found in a learned PAG do not contain a collider. Hence, the generalised tetrad condition avoids the $M$-bias~\cite{greenland2003quantifying,pearl2009myth} in causal effect estimation using the condition. Furthermore, the authors have proved that the generalised tetrad condition can be used to discover a pair of valid AIVs if there is a pair of AIVs in the data. Based on the graphical property of an AIV in a MAG, Cheng et al.~\cite{cheng2022ancestral} found that each directed AIV is in the set $Adj(Y)\setminus\{W\}$ (the set of adjacent nodes of $Y$ excluding $Y$) in a MAG. Using the finding and the generalised tetrad condition, a data-driven algorithm, Ancestral IV based on the Generalised Tetrad condition (AIV.GT)~\cite{cheng2022discovering}, is developed for estimating the causal effects from data with latent variables. AIV.GT finds a pair of direct AIVs and their corresponding conditioning sets using the generalised tetrad condition and the search space is confined in the set $Adj(Y)\setminus\{W\}$. Same to IV.tetrad, AIV.GT also provides a bound estimation of causal effects, but AIV.GT is faster since it uses a local search.

\subsection{Summary}
\label{subsub:summary}
 Table~\ref{tab:002} provides a summary of the properties and characteristics of the data-driven causal effect estimation methods listed in Table~\ref{tab:methods}. The methods listed in Table~\ref{tab:002} are used to estimate ACEs from observational data, and therefore, they can be widely applied in fields such as economics~\cite{imbens2015causal,heckman2008econometric}, epidemiology~\cite{sauer2013review,textor2016robust}, healthcare~\cite{robins1986new}, and bioinformatics~\cite{maathuis2010predicting,zisoulis2012autoregulation}. Note that some of these methods are parametric and some are nonparametric~\cite{pearl2009causal,correa2017causal,jaber2019causal,jaber2019identification}. A parametric method assumes  a functional form of the underlying data generation process, such as linearity, whereas a nonparametric method does not make such an assumption.  As shown in Table~\ref{tab:002}, most methods are parametric, and only a few methods are nonparametric, including CET-SAT, DICE, EHS, GAC, DAVS, CEELS and modelIV. It is worth mentioning that in some cases, the causal effect may not be nonparametrically identifiable, and the best option is to derive a bound estimation by conducting sensitive analysis~\cite{robins2000sensitivity,scharfstein2021semiparametric,christopher2002identification,tortorelli1994design}. Researchers need to carefully consider these properties/characteristics in light of their specific research question, data, and resources, to select  an appropriate method for their application.

\section{Performance and Complexity Evaluation}
\label{sec:005} 
 In this section, we design simulation experiments to illustrate the performance of the estimators we have reviewed for estimating $ACE(W, Y)$ in terms of the two challenges posed by latent confounders. The primary motivation behind this simulation experimental design is that in real-world applications, latent confounders are widespread within the data. Hence, we consider two different scenarios: data with latent confounders which are not between $(W, Y)$, and data with latent confounders, one of which is between $(W, Y)$. Furthermore, we conduct simulation studies to evaluate the time complexity of these estimators by varying the numbers of samples and variables.

\subsection{Performance Evaluation on Synthetic Datasets}
\begin{figure}[t]
	\centering
	\includegraphics[scale=0.33]{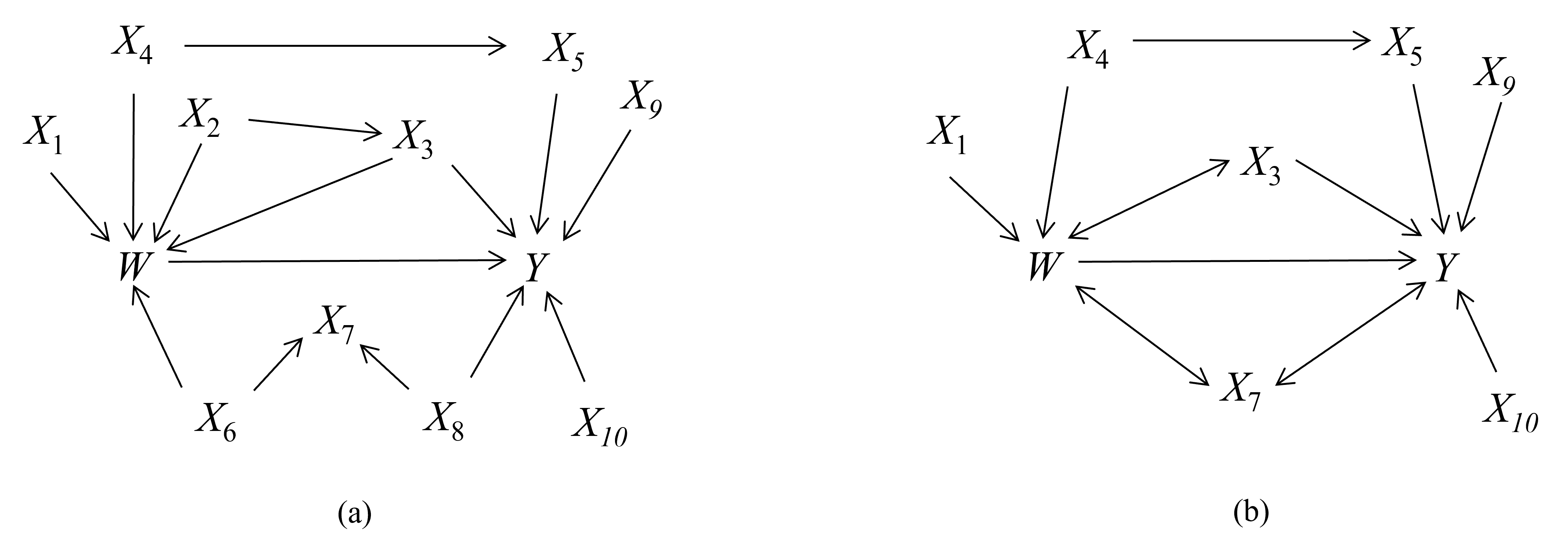}
	\caption{ {(a) A causal DAG for synthetic datasets generation, and (b) its corresponding MAG, where $\{X_2, X_6, X_8\}$ are removed.}}
	\label{fig:DAGgeneration}
\end{figure}

	 \paragraph{Data Generation} We generate two groups of synthetic datasets to assess these data-driven causal effect estimators reviewed in this survey. Following the data generation procedure described in~\cite{witte2019covariate}, we first utilise the causal DAG shown in Fig.~\ref{fig:DAGgeneration} (a) to generate synthetic datasets without latent confounders. In this causal DAG, there are four back-door paths between the treatment $W$ and outcome $Y$, indicating sparse structures around $W$ and $Y$. To obtain synthetic datasets with latent confounders that are not between $(W, Y)$, following the corresponding MAG in Fig.~\ref{fig:DAGgeneration} (b), we remove the variables $\{X_2, X_6, X_8\}$ from datasets without latent confounders and treat them as latent confounders. These datasets are named as Group 1 datasets. Additionally, we remove the variable $\{X_3\}$ from Group 1 datasets to obtain the synthetic datasets with latent confounders, i.e. the latent confounder $X_3$ affects both $W$ and $Y$. These synthetic datasets are called Group 2 datasets. Both groups of synthetic datasets are used to evaluate the performance of estimators reviewed in this survey.

 To approximate real-world scenarios, we generate 90 additional random variables as noise variables in the synthetic datasets. These variables are correlated with each other but not with the variables in the causal DAG shown in Fig.~\ref{fig:DAGgeneration} (a). In our simulation studies, we vary the sample sizes, generating datasets with 2k, 4k, 6k, 8k, 10k, and 100k samples. The ground-truth $ACE(W, Y)$ for all synthetic datasets is 0.5. To mitigate potential biases resulting from the data generation procedure, we repeat the process 20 times, generating 20 datasets for each sample size.

 \paragraph{Parameters Setting \& methods exclusion} In our simulation studies, the significance level $\alpha$ is set to 0.05 for all estimators involved. For conditional independence tests, we use \emph{gaussCItest} in the {R} package \emph{pcalg}~\cite{kalisch2012causal} to implement the estimator involved. CovSelHigh~\cite{haggstrom2018data}  has the implementations of five criteria: all causes of $W$, all causes of $Y$, all causes of $W$ excluding causes of $Y$,  all causes of $Y$ excluding causes of $W$, and the union of all causes of $W$ and all causes of $Y$, as the adjustment set respectively. Following the notation in~\cite{de2011covariate}, the identified adjustment sets are denoted as $\hat{\mathbf{X}}_{\rightarrow W}$, $\hat{\mathbf{X}}_{\rightarrow Y}$, $\hat{\mathbf{Q}}_{\rightarrow W}$, $\hat{\mathbf{Z}}_{\rightarrow Y}$ and $\hat{\mathbf{X}}_{\rightarrow W,Y} = (\hat{\mathbf{X}}_{\rightarrow W} \cup \hat{\mathbf{X}}_{\rightarrow Y})$ respectively. DIDA and DIDN are not included since both estimators require directed causal information~\cite{fang2020ida}. EHS~\cite{entner2013data}, CE-SAT~\cite{hyttinen2015calculus} and CovlSet~\cite{haggstrom2015covsel} are not evaluated since they have a high time-complexity and do not work in the synthetic datasets. For bound estimators, IDA, semi-local IDA, LV-IDA, DICE, IV.tetrad and AIV.GT, we report the mean of the bound estimations as their final results.

 \paragraph{Methods classification} Following the classification of the methods  in Section~\ref{subsec:methodsclass} regarding the challenges of latent confounders, the methods are divided into three types: Type I, requiring causal sufficiency: IDA, semi-local IDA,  $\hat{\mathbf{X}}_{\rightarrow W}$, $\hat{\mathbf{X}}_{\rightarrow Y}$, $\hat{\mathbf{Q}}_{\rightarrow W}$, $\hat{\mathbf{Z}}_{\rightarrow Y}$ and $\hat{\mathbf{X}}_{\rightarrow W,Y}$; Type II handling data with latent confounders which are not between $(W, Y)$: LV-IDA, DICE, DAVS and CEELS; and Type III, IV-based estimators: sisVIVE, IV.tetrad, $AIViP$ and AIV.GT.

 \paragraph{Evaluation metrics}  We use the estimation error $\epsilon_{ACE} = \hat{ACE} -ACE$ to evaluate the performance of these estimators, where $\hat{ACE}$ is the estimated causal effect of $W$ on $Y$. For the three groups of synthetic datasets, the ground-truth $ACE(W, Y)$ is 0.5.  For each sample size, we report the Mean and standard deviation (STD) of $\epsilon_{ACE}$ as the final result over 20 synthetic datasets.

\begin{table}[t]
\caption{Estimation errors $\epsilon_{ACE}$ (Mean$\pm$STD) over 20 Group 1 synthetic datasets (with latent confounders which are not between $W$ and $Y$). The first set of estimators (Type I) requires causal sufficiency assumption, the second set of estimators (Type II) are able to handle data with latent confounders which are not between $W$ and $Y$, and the third set of estimators (Type III) are IV-based estimators. Note that semiIDA is short for Semi-local IDA. Methods marked with `*' are bound estimators.}
\begin{tabular}{|cc|cccccc|}
	\hline
	\multicolumn{2}{|c|}{\multirow{2}{*}{Methods}}                                            & \multicolumn{6}{c|}{Number of Samples}                                                                                                                           \\ \cline{3-8} 
	\multicolumn{2}{|c|}{}                                                                    & \multicolumn{1}{c|}{2k}    & \multicolumn{1}{c|}{4k}    & \multicolumn{1}{c|}{6k}    & \multicolumn{1}{c|}{8k}    & \multicolumn{1}{c|}{10k}   & 100k  \\ \hline
	\multicolumn{1}{|c|}{\multirow{7}{*}{Type I}}         & IDA*                                   & \multicolumn{1}{c|}{0.273$\pm$0.078} & \multicolumn{1}{c|}{0.303$\pm$0.077} & \multicolumn{1}{c|}{0.285$\pm$0.073} & \multicolumn{1}{c|}{0.290$\pm$0.064} & \multicolumn{1}{c|}{0.284$\pm$0.058} & 0.274$\pm$0.015 \\ \cline{2-8} 
	\multicolumn{1}{|c|}{}                            & semiIDA*                               & \multicolumn{1}{c|}{0.272$\pm$0.111} & \multicolumn{1}{c|}{0.252$\pm$0.082} & 
	\multicolumn{1}{c|}{0.279$\pm$0.039} &
	\multicolumn{1}{c|}{0.288$\pm$0.043} & 
	\multicolumn{1}{c|}{0.292$\pm$0.038} & 0.215$\pm$0.887 \\ \cline{2-8} 
	\multicolumn{1}{|c|}{}                            & $\hat{\mathbf{X}}_{\rightarrow W}$    & \multicolumn{1}{c|}{0.329$\pm$0.019} & \multicolumn{1}{c|}{0.242$\pm$0.024} & \multicolumn{1}{c|}{0.291$\pm$0.027} & \multicolumn{1}{c|}{0.279$\pm$0.021} & \multicolumn{1}{c|}{0.287$\pm$0.016} & 0.282$\pm$0.013 \\ \cline{2-8} 
	\multicolumn{1}{|c|}{}                            & $\hat{\mathbf{Q}}_{\rightarrow W}$    & \multicolumn{1}{c|}{0.286$\pm$0.024} & \multicolumn{1}{c|}{0.203$\pm$0.018} & \multicolumn{1}{c|}{0.258$\pm$0.015} & \multicolumn{1}{c|}{0.241$\pm$0.019} & \multicolumn{1}{c|}{0.241$\pm$0.014} & 0.247$\pm$0.016 \\ \cline{2-8} 
	\multicolumn{1}{|c|}{}                            & $\hat{\mathbf{X}}_{\rightarrow Y}$    & \multicolumn{1}{c|}{0.265$\pm$0.018} & \multicolumn{1}{c|}{0.183$\pm$0.015} & \multicolumn{1}{c|}{0.261$\pm$0.018} & \multicolumn{1}{c|}{0.224$\pm$0.013} & \multicolumn{1}{c|}{0.226$\pm$0.017} & 0.224$\pm$0.012 \\ \cline{2-8} 
	\multicolumn{1}{|c|}{}                            & $\hat{\mathbf{Z}}_{\rightarrow Y}$    & \multicolumn{1}{c|}{0.279$\pm$0.018} & \multicolumn{1}{c|}{0.187$\pm$0.016} & \multicolumn{1}{c|}{0.253$\pm$0.019} & \multicolumn{1}{c|}{0.227$\pm$0.012} & \multicolumn{1}{c|}{0.224$\pm$0.013} & 0.225$\pm$0.011 \\ \cline{2-8} 
	\multicolumn{1}{|c|}{}                            & $\hat{\mathbf{X}}_{\rightarrow W, Y}$ & \multicolumn{1}{c|}{0.316$\pm$0.018} & \multicolumn{1}{c|}{0.241$\pm$0.015} & \multicolumn{1}{c|}{0.310$\pm$0.021} & \multicolumn{1}{c|}{0.278$\pm$0.023} & \multicolumn{1}{c|}{0.277$\pm$0.019} & 0.277$\pm$0.013 \\ \hline
	\multicolumn{1}{|c|}{\multirow{6}{*}{Type II}}  & GAC                                   & \multicolumn{1}{c|}{0.847$\pm$0.550} & \multicolumn{1}{c|}{1.235$\pm$0.384} & \multicolumn{1}{c|}{1.250$\pm$0.225} & \multicolumn{1}{c|}{1.285$\pm$0.272} & \multicolumn{1}{c|}{1.174$\pm$0.403} & 1.018$\pm$0.491 \\ \cline{2-8} 
	\multicolumn{1}{|c|}{}                            & LV-IDA*                                & \multicolumn{1}{c|}{0.539$\pm$0.063} & \multicolumn{1}{c|}{0.442$\pm$0.065} & \multicolumn{1}{c|}{0.487$\pm$0.119} & \multicolumn{1}{c|}{0.492$\pm$0.023} & \multicolumn{1}{c|}{0.437$\pm$0.132} & 0.133$\pm$0.101 \\ \cline{2-8} 
	\multicolumn{1}{|c|}{}                            & DICE*                                  & \multicolumn{1}{c|}{0.481$\pm$0.096} & \multicolumn{1}{c|}{0.569$\pm$0.059} & \multicolumn{1}{c|}{0.477$\pm$0.044} & \multicolumn{1}{c|}{0.511$\pm$0.053} & \multicolumn{1}{c|}{0.505$\pm$0.045} & 0.553$\pm$0.034 \\ \cline{2-8} 
	\multicolumn{1}{|c|}{}                            & DAVS                                  & \multicolumn{1}{c|}{0.634$\pm$0.158} & \multicolumn{1}{c|}{0.528$\pm$0.181} & \multicolumn{1}{c|}{0.236$\pm$0.245} & \multicolumn{1}{c|}{0.240$\pm$0.209} & \multicolumn{1}{c|}{0.235$\pm$0.355} & 0.189$\pm$0.034 \\ \cline{2-8} 
	\multicolumn{1}{|c|}{}                            & CEELS                                 & \multicolumn{1}{c|}{0.621$\pm$0.267} & \multicolumn{1}{c|}{0.615$\pm$0.269} & \multicolumn{1}{c|}{0.253$\pm$0.204} & \multicolumn{1}{c|}{0.215$\pm$0.121} & \multicolumn{1}{c|}{0.236$\pm$0.222} & 0.108$\pm$0.024 \\ \hline
	\multicolumn{1}{|c|}{\multirow{4}{*}{Type III}}   & sisVIVE                               & \multicolumn{1}{c|}{0.488$\pm$0.174} & \multicolumn{1}{c|}{0.778$\pm$0.364} & \multicolumn{1}{c|}{0.619$\pm$0.278} & \multicolumn{1}{c|}{0.686$\pm$0.553} & \multicolumn{1}{c|}{0.767$\pm$0.342} & 0.642$\pm$0.368 \\ \cline{2-8} 
	\multicolumn{1}{|c|}{}                            & IV.tetrad*                             & \multicolumn{1}{c|}{0.299$\pm$0.197} & \multicolumn{1}{c|}{0.243$\pm$0.183} & \multicolumn{1}{c|}{0.324$\pm$0.265} & \multicolumn{1}{c|}{0.372$\pm$0.244} & \multicolumn{1}{c|}{0.298$\pm$0.272} & 0.311$\pm$0.214 \\ \cline{2-8} 
	\multicolumn{1}{|c|}{}                            & $AIViP$                                 & \multicolumn{1}{c|}{0.228$\pm$0.222} & \multicolumn{1}{c|}{0.242$\pm$0.164} & \multicolumn{1}{c|}{0.135$\pm$0.124} & \multicolumn{1}{c|}{0.162$\pm$0.099} & \multicolumn{1}{c|}{0.153$\pm$0.125} & 0.109$\pm$0.053 \\ \cline{2-8} 
	\multicolumn{1}{|c|}{}                            & AIV.GT*                                 & \multicolumn{1}{c|}{0.215$\pm$0.270} & \multicolumn{1}{c|}{0.342$\pm$0.276} & \multicolumn{1}{c|}{0.308$\pm$0.250} & \multicolumn{1}{c|}{0.268$\pm$0.223} & \multicolumn{1}{c|}{0.240$\pm$0.247} & 0.355$\pm$0.234 \\ \hline
\end{tabular}
\label{tab:latentC004}
\end{table}

 \paragraph{Results} We report all results on both groups of datasets  in Tables~\ref{tab:latentC004} and~\ref{tab:lantC005}, and discuss the performance of the different types of methods in detail below.  

 With Group 1 datasets (with latent confounders but not between $W$ and $Y$), methods in Type II and III are expected to perform well since data sets satisfy their assumptions. However, from Table 3 methods in Type I and III perform well overall, and the performance of methods in Type II varies. For methods in Type I, the data sets do not satisfy their causal sufficiency requirement (no latent variables). However, their performance is relatively well since the latent variables not between $(W, Y)$ may not bias the causal effects very much and their results can be still useful. For methods in Type II, the data sets satisfy their requirement (latent variables not between $(W, Y)$). However, they perform well only when the data set is large. The lowest biased estimates are achieved by GAC and CEELS in data sets of 100k. There is a clear trend that bias is reducing with the increase in data set sizes. This is because PAGs are difficult to learn correctly from data due and learning correct PAG needs big data sets. In contrast, learning CPDAGs is much easier and needs smaller data sets. For methods in Type III, the data sets satisfy their requirement (except sisVIVE) since they are supposed to handle all data types as long as IVs (AIVs) are identified. sisVIVE requires more than half of the variables to be valid IVs, while there are only two valid IVs, i.e. ${X_1, X_4}$ in the data. Both IV.tetrad and AIC.GT produces bound estimations and hence their mean biases are moderate. 

Note that identifying an IV is quite challenging in many applications. In this simulation, we aim at many applications where IVs are few. However, when genetic variants are considered as IVs to estimate the causal effect of a modifiable factor on a disease, i.e. Mendelian randomisation (MR) analysis~\cite{kang2016instrumental,brumpton2020avoiding}, many genetic variants can be used as IVs and hence sisVIVE has many applications. See Section~\ref{subsec:disbound} for details.

 With Group 2 datasets (with a latent confounder between $W$ and $Y$), methods in Type III are supposed to work well and the requirements of other methods are not satisfied. From Table 4, all methods in Type I have large biases because of the latent confounder between $W$ and $Y$. In comparing results in Table 3, a latent confounder between $W$ and $Y$ biases methods of Type I more. Therefore, we can see that the unconfoundedness (no latent variable between $W$ and $Y$) is the essential assumption for most methods in causal effect estimation.  All methods in Type II do not work on the data since the edge $W \to Y$ is invisible and they terminate themselves. For methods in Type III, $AIViP$ and AIV.GT performed well as expected.  AIV.GT is a bound estimator and hence its biases are moderate. sisVIVE does not perform well because its requirement of more than half of the variables being valid IVs is not satisfied. IV.tetrad shows worse results in Group 2 datasets than those in Group 1 data sets because fewer variables pass the IV.tetrad test and yield a larger bound estimation.
 
\begin{table}[t]
\caption{Estimation errors $\epsilon_{ACE}$ (Mean$\pm$STD) over 20 Group 2 synthetic datasets (with latent confounders, one of which is between $W$ and $Y$). The first set of estimators (Type I) requires causal sufficiency assumption, the second set of estimators (Type II) are able to handle data with latent confounders which are not between $W$ and $Y$, and the third set of estimators (Type III) are IV-based estimators. Methods marked with `*' are bound estimators. Note that semiIDA is short for Semi-local IDA and `$-$' denotes N/A returned by the estimator on these datasets since edge $W \rightarrow Y$ is invisible and the methods do not work on the datasets. }
\begin{tabular}{|cc|cccccc|}
	\hline
	\multicolumn{2}{|c|}{\multirow{2}{*}{Methods}}                                            & \multicolumn{6}{c|}{Number of Samples}                                                                                                                           \\ \cline{3-8} 
	\multicolumn{2}{|c|}{}                                                                    & \multicolumn{1}{c|}{2k}    & \multicolumn{1}{c|}{4k}    & \multicolumn{1}{c|}{6k}    & \multicolumn{1}{c|}{8k}    & \multicolumn{1}{c|}{10k}   & 100k  \\ \hline
	\multicolumn{1}{|c|}{\multirow{7}{*}{Type I}}         & IDA*                                   & \multicolumn{1}{c|}{0.914$\pm$0.070} & \multicolumn{1}{c|}{0.874$\pm$0.057} & \multicolumn{1}{c|}{0.891$\pm$0.081} & \multicolumn{1}{c|}{0.885$\pm$0.055} & \multicolumn{1}{c|}{0.898$\pm$0.055} & 0.896$\pm$0.014 \\ \cline{2-8} 
	\multicolumn{1}{|c|}{}                            & semiIDA*                               & \multicolumn{1}{c|}{0.914$\pm$0.127} & \multicolumn{1}{c|}{0.919$\pm$0.077} & \multicolumn{1}{c|}{0.892$\pm$0.038} & \multicolumn{1}{c|}{0.878$\pm$0.042} &
	\multicolumn{1}{c|}{0.885$\pm$0.034} & 0.887$\pm$0.014 \\ \cline{2-8} 
	\multicolumn{1}{|c|}{}                            & $\hat{\mathbf{X}}_{\rightarrow W}$    & \multicolumn{1}{c|}{0.858$\pm$0.026} & \multicolumn{1}{c|}{0.932$\pm$0.021} & \multicolumn{1}{c|}{0.870$\pm$0.018} & \multicolumn{1}{c|}{0.899$\pm$0.016} & \multicolumn{1}{c|}{0.890$\pm$0.012} & 0.893$\pm$0.013 \\ \cline{2-8} 
	\multicolumn{1}{|c|}{}                            & $\hat{\mathbf{Q}}_{\rightarrow W}$    & \multicolumn{1}{c|}{0.837$\pm$0.023} & \multicolumn{1}{c|}{0.925$\pm$0.028} & \multicolumn{1}{c|}{0.869$\pm$0.021} & \multicolumn{1}{c|}{0.899$\pm$0.015} & \multicolumn{1}{c|}{0.890$\pm$0.014} & 0.893$\pm$0.013 \\ \cline{2-8} 
	\multicolumn{1}{|c|}{}                            & $\hat{\mathbf{X}}_{\rightarrow Y}$    & \multicolumn{1}{c|}{0.832$\pm$0.018} & \multicolumn{1}{c|}{0.906$\pm$0.024} & \multicolumn{1}{c|}{0.845$\pm$0.021} & \multicolumn{1}{c|}{0.895$\pm$0.018} & \multicolumn{1}{c|}{0.892$\pm$0.015} & 0.896$\pm$0.012 \\ \cline{2-8} 
	\multicolumn{1}{|c|}{}                            & $\hat{\mathbf{Z}}_{\rightarrow Y}$    & \multicolumn{1}{c|}{0.823$\pm$0.019} & \multicolumn{1}{c|}{0.904$\pm$0.021} & \multicolumn{1}{c|}{0.855$\pm$0.017} & \multicolumn{1}{c|}{0.895$\pm$0.015} & \multicolumn{1}{c|}{0.888$\pm$0.011} & 0.896$\pm$0.012 \\ \cline{2-8} 
	\multicolumn{1}{|c|}{}                            & $\hat{\mathbf{X}}_{\rightarrow W, Y}$ & \multicolumn{1}{c|}{0.867$\pm$0.023} & \multicolumn{1}{c|}{0.934$\pm$0.021} & \multicolumn{1}{c|}{0.857$\pm$0.019} & \multicolumn{1}{c|}{0.896$\pm$0.015} & \multicolumn{1}{c|}{0.892$\pm$0.013} & 0.896$\pm$0.012 \\ \hline
	\multicolumn{1}{|c|}{\multirow{6}{*}{Type II}}                 & GAC                                      & \multicolumn{1}{c|}{$-$} & \multicolumn{1}{c|}{$-$} & \multicolumn{1}{c|}{$-$} & \multicolumn{1}{c|}{$-$} & \multicolumn{1}{c|}{$-$} & $-$ \\ \cline{2-8} 
	\multicolumn{1}{|c|}{}                            & LV-IDA*         & \multicolumn{1}{c|}{$-$} & \multicolumn{1}{c|}{$-$} & \multicolumn{1}{c|}{$-$} & \multicolumn{1}{c|}{$-$} & \multicolumn{1}{c|}{$-$} & $-$                            \\ \cline{2-8} 
	\multicolumn{1}{|c|}{}                              & DICE*       & \multicolumn{1}{c|}{$-$} & \multicolumn{1}{c|}{$-$} & \multicolumn{1}{c|}{$-$} & \multicolumn{1}{c|}{$-$} & \multicolumn{1}{c|}{$-$} & $-$ \\ \cline{2-8} 
	\multicolumn{1}{|c|}{}                            & DAVS                                  & \multicolumn{1}{c|}{$-$} & \multicolumn{1}{c|}{$-$} & \multicolumn{1}{c|}{$-$} & \multicolumn{1}{c|}{$-$} & \multicolumn{1}{c|}{$-$} & $-$ \\ \cline{2-8} 
	\multicolumn{1}{|c|}{}                            & CEELS                                 & \multicolumn{1}{c|}{$-$} & \multicolumn{1}{c|}{$-$} & \multicolumn{1}{c|}{$-$} & \multicolumn{1}{c|}{$-$} & \multicolumn{1}{c|}{$-$} & $-$ \\ \hline
	\multicolumn{1}{|c|}{\multirow{4}{*}{Type III}}   & sisVIVE                               & \multicolumn{1}{c|}{{0.609$\pm$0.311}} & \multicolumn{1}{c|}{0.732$\pm$0.335} & \multicolumn{1}{c|}{0.464$\pm$0.187} & \multicolumn{1}{c|}{0.640$\pm$0.466} & \multicolumn{1}{c|}{0.633$\pm$0.618} & 0.576$\pm$0.452 \\ \cline{2-8} 
	\multicolumn{1}{|c|}{}                            & IV.tetrad*                             & \multicolumn{1}{c|}{1.002$\pm$0.477} & \multicolumn{1}{c|}{0.974$\pm$0.383} & \multicolumn{1}{c|}{0.979$\pm$0.335} & \multicolumn{1}{c|}{0.891$\pm$0.415} & \multicolumn{1}{c|}{0.882$\pm$0.434} & 0.923$\pm$0.323 \\ \cline{2-8} 
	\multicolumn{1}{|c|}{}                            & $AIViP$                                 & \multicolumn{1}{c|}{0.228$\pm$0.222} & \multicolumn{1}{c|}{0.244$\pm$0.163} & \multicolumn{1}{c|}{0.141$\pm$0.127} & \multicolumn{1}{c|}{0.162$\pm$0.099} & \multicolumn{1}{c|}{0.153$\pm$0.125} & 0.109$\pm$0.053 \\ \cline{2-8} 
	\multicolumn{1}{|c|}{}                            & AIV.GT*                                 & \multicolumn{1}{c|}{0.270$\pm$0.247} & \multicolumn{1}{c|}{0.378$\pm$0.555} & \multicolumn{1}{c|}{0.346$\pm$0.376} & \multicolumn{1}{c|}{0.270$\pm$0.408} & \multicolumn{1}{c|}{0.295$\pm$0.445} &  0.347$\pm$0.297 \\ \hline
\end{tabular}
\label{tab:lantC005}
\end{table}

\subsection{ {Complexity Evaluation}}
 We conduct a simulation study to examine the time complexity of the reviewed estimators using synthetic datasets with latent confounders which are not between $W$ and $Y$. The R package \emph{pcalg}~\cite{kalisch2012causal} is utilised for generating the MAGs (Maximal Ancestral Graph) and synthetic datasets, following the procedure outlined in the literature~\cite{cheng2020local}. Specifically, we employ the functions \emph{randomDAG} and \emph{rmvdag}, where the parameter $prob$ for \emph{randomDAG} is set to 0.1, resulting in the generation of a sparse DAG. Subsequently, we obtain a sparse MAG by removing 10 nodes from the obtained DAG. Note that we pick up the MAG with $W\rightarrow Y$ visible for generating data with latent confounders such that all methods can be evaluated. This sparse DAG is used for generating synthetic data. To introduce latent confounders, the removed nodes from the sparse DAG are removed from the generated synthetic data. The computations of all estimators are conducted on a machine equipped with an Intel (R) Core (TM) i7-9700K CPU and 32 GB of RAM. 

\begin{figure}[t]
\centering
\includegraphics[scale=0.32]{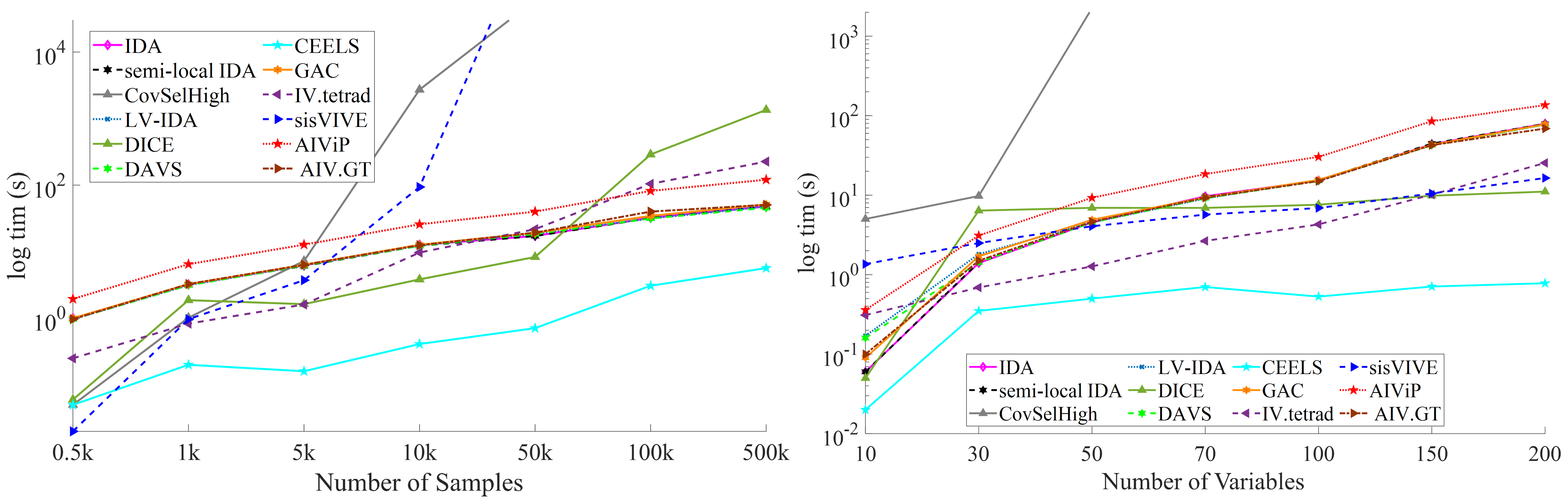}
\caption{ Running time of the data-driven estimators. We only show the runtime of CovSelHigh, since the five criteria rely on the same process of CPDAG learning. We use the logarithmic form of the running time for the Y-axis.}
\label{fig:runtimes}
\end{figure}

 To evaluate the scalability in terms of the number of samples, the number of measured variables is fixed at 50 and evaluated with different number of sample sizes: 0.5k, 5k, 10k, 50k, 100k, and 500k. Similarly, to assess the scalability regarding the number of variables, the
 sample size is fixed at 10k,  while the number of measured variables is varied as 10, 30, 50, 70, 100, 150, and 200. In all experiments, we repeat each experiment 10 times, and the mean running time of each estimator is recorded and visualised in Fig.~\ref{fig:runtimes}.

 From the left panel of Fig.~\ref{fig:runtimes}, we have the following observations on time efficiency and scalability regarding the number of samples: (1) Most methods have similar running times. IDA, semi-local IDA, LV-IDA, DAVS, GAC, $AIViP$ and AIV.GT exhibit similar time efficiency since they employ a global structural learning algorithm to learn a CPDAG or a PAG from the data first before estimating causal effects. These methods exhibit good scalability with sample sizes. (2) The running time of DICE and IV.tetrad is similar to the previous methods because both methods need to enumerate all the possible adjustment sets and possible IVs from a (possibly large) candidate set although they do not need a global causal structure (they use a local structure). (3) CEELS is the fastest algorithm among all since it employs a purely local search strategy for finding the adjustment set. (4) CovSelHigh and sisVIVE, do not scale well with the number of samples. They are efficient when the number of samples is small (fewer than 5K). Their run-time exceeds 2,000 seconds when the sample size is larger than 50k since both methods involve a large number of matrix manipulations (whose complexity is polynomial to the number of samples) to obtain the corresponding coefficient weights for samples.

 From the right panel of Fig.~\ref{fig:runtimes} with respect to the number of variables, we have the same observations of time efficiency and scalability of all methods as above except sisVIVE. sisVIVE employs the equivalent Lagrangian form for causal effect estimation so its time complexity does not increase significantly with the number of variables. Note that the time complexity of causal structure learning is ultimately exponential to the number of variables. The algorithms deal with variables in a few hundreds not thousands except the causal structure is extremely sparse.

\section{Discussions} 
 \label{sec:disc} 
 In this section, we present some discussions related to issues with data-driven causal effect estimation methods, and their applications and potential.

\subsection{Causal Graph Recovery Approach vs Conditional Independence Test based Approach }
\label{subsec:practicalcons}
 All reviewed methods apart from IV (and AIV) methods fall into two broad categories: Causal Graph Recovery Approach vs Conditional Independence Test based Approach. We compare the two types of methods as follows.

\paragraph{Causal Graph Recovery based Approach} This type of data-driven methods learns a causal structure (CPDAG/PAG) in observational data using a causal structure learning algorithm~\cite{scutari2010learning,kalisch2012causal,vowels2021d}, then identifies \textit{adjustment sets} for confounding adjustment in the learned CPDAG/PAG, and finally, uses the adjustment sets for causal effect estimation. Because of uncertainty in causal structure learning, the causal effect is a bond estimation.  Because this type of methods requires a discovered causal graph, the performance of these methods relies on the correctness of the recovered causal graph. The issues with the current causal structure learning methods, such as cumulative statistical errors and incorrect edge orientation affect the performance of causal effect estimation~\cite{cheng2020local,gradu2022valid}. IDA~\cite{maathuis2009estimating}, semi-local IDA~\cite{perkovic2017interpreting}, DIDA and DIDN~\cite{fang2020ida} can be used for a bound estimation when there are no latent variables. Using the pretreatment variables, the certainty of the estimation can be improved significantly from the bound estimation to unique estimation by CovSel~\cite{haggstrom2015covsel} and CovSelHigh~\cite{haggstrom2018data}. The time efficiency is also improved greatly since CovSel~\cite{haggstrom2015covsel} and CovSelHigh~\cite{haggstrom2018data} use local structure discovery. When there are latent variables but not in between $W$ and $Y$, LVIDA~\cite{malinsky2017estimating} and GAC~\cite{perkovic2018complete} can be used.

\paragraph{Conditional Independence Test based Approach} It is possible to find a confounding adjustment set using statistical tests when there are no latent variables between $W$ and $Y$ and bypass learning a CPDAG or a PAG from data with some knowledge. One way is to utilise a COSO variable (a cause of $W$ or a spouse of $W$ only, denoted by $Q$). The COSO variable can be given by domain knowledge as a direct cause of $W$ but not a direct cause of $Y$, or be discovered from data using statistical tests. Then, the triple $(W, Y, Q)$ is used in statistical tests to find an adjustment set from data directly. A direct search for an adjustment set based on the conditional independence tests is global (all variables apart from $W$ and $Y$ are candidates for an adjustment set), such as DAVS~\cite{cheng2022toward}. When pretreatment variables are known, the search can be local (only variables around $W$ and $Y$ are candidates for an adjustment set) as implemented in CEELS~\cite{cheng2020causal}. Alternatively, when pretreatment variables are known, there is no need to use a COSO variable but a direct search for an adjustment set using EHS conditions by a global search as implemented by EHS method~\cite{entner2013data}. EHS provide a bound estimation. Also, EHS method employs an exhaustive search and only deals with a small number of variables.

The conditional independence test approach, in general, is more accurate than the causal graph recovery approach since the uncertainty of a causal graph found in data is high due to the high complexity of the problem itself~\cite{chickering1996learning,chickering2002learning}. In some cases, the edge directions cannot be identified from data~\cite{spirtes2010introduction,pearl2009causal,malinsky2017estimating}. For example, if there is a causal structure $X_1 \to X_2 \rightarrow X_3 \leftarrow X_1$, the edge direction between $X_1$ and $X_3$ is not identifiable using data only. Note that one incorrect orientation of edge around $W$ or $Y$ may lead to an error in adjustment set identification. In contrast, if there are no errors in the statistical tests, the adjustment set determined by the conditional independence test approach is correct when the adjustment set is found.

\subsection{Considerations When Using Data-Driven Estimators}
One question is whether it is possible to obtain unique estimation without domain knowledge. The answer relies on the partial causal information we have in hand. In the following discussions, we assume that the faithfulness assumption is satisfied, and there are no errors in statistical tests.   

If there is a latent confounder between $W$ and $Y$, there is no algorithm to identify the causal effect directly from the data. If we know an AIV or two valid (unspecified) AIVs, $AIViP$~\cite{cheng2022ancestral} and AIV.GT~\cite{cheng2022discovering} can be used to recover the causal effect of $W$ on $Y$ from data alone. If there is no information about the number of valid IVs, it is better to use multiple IV based methods, such as AIV.GT~\cite{cheng2022discovering}, IV.Tetrad~\cite{silva2017learning}, sisVIVE~\cite{kang2016instrumental} and ModeIV~\cite{hartford2021valid}, to obtain a set of causal effects, and then choose a range of consistent estimates to draw some common-sense conclusions. Note that before using IV based estimators, it is better to analyse the assumptions required to obtain reliable conclusions.

If there is not a latent confounder between $W$ and $Y$, but there are latent variables elsewhere in a system,  EHS~\cite{entner2013data} and CEELS~\cite{cheng2020local} should be the first choice since they do not rely on learning PAG. Note that EHS is inefficient for a problem with a dozen or more variables since it uses an exhaustive search strategy. DICE~\cite{cheng2020causal} provides a bound estimate and does not rely on PAG. Hence, DICE can be used as the second choice. Other methodsGAC~\cite{perkovic2018complete}, DAVS~\cite{cheng2022toward}, and LV-IDA~\cite{malinsky2017estimating} use a learned DAG, and their results need to be taken with caution since learning a PAG needs a large data set and learned PAG may be inaccurate for small or moderate data sets. Any partial knowledge that constrains the edge directions for improving PAG accuracy will be helpful for reducing biases in causal effect estimation.

If there is not a latent variable between $W$ and $Y$, and there are no latent variables in a system, CovSel~\cite{haggstrom2015covsel} and CovSelHigh~\cite{haggstrom2018data} can be efficient and accurate for estimating the causal effects from data directly. When the data set is large, i.e. with a large number of samples and/or a large number of variables, IDA~\cite{maathuis2009estimating}, and Semi-local IDA~\cite{perkovic2017interpreting} can provide a reliable bound estimation.

A main challenge of causal effect estimation is the difficulty in justifying whether there exist latent variables in many real-world applications. If there is not any domain knowledge available, using several purely data-driven estimations to cross-check would be better than a single method in terms of obtaining a consistent conclusion. Furthermore, it is better to combine domain knowledge or the suggestions of a domain expert to obtain a reliable conclusion after using the data-driven causal effect estimation methods.

\subsection{Double Dipping}
\label{subsec:dd}
A risk for data-driven causal effect estimation is the bias caused by ``double dipping''. To estimate the causal effect of $W$ on $Y$ from observational data using graphical causal modelling, causal discovery methods are employed to recover the underlying causal structure, followed by a causal inference method. However, this approach faces a statistical challenge known as ``double dipping,'' which can invalidate the coverage guarantees of classical confidence intervals\cite{gradu2022valid}. This problem is caused by both learning the causal graph and estimating the causal effect on the same data and the data reuse may lead to a biased estimation of the causal effect. Gradu et al.\cite{gradu2022valid} have developed a randomising causal discovery method to remedy the ``double dipping'' bias. This is a new direction for bias mitigation in data-driven causal effect estimation. Readers should be aware of the future development in this space.   

\subsection{Applications \& Potential}
\label{subsec:disbound} 
\paragraph{Applications} The majority of methods discussed in this paper have been developed in recent years. Some relatively older methods (from around ten years ago) have been applied to biological, health, and medicinal research. We discuss some typical applications in the following.

Maathuis et al.~\cite{maathuis2010predicting} applied IDA to gene expression data obtained without interventions (i.e. observational data) to estimate the regulatory causal effect of a gene (repeat for 234 different genes) on the other 5360 genes. The top $k$ genes with the highest causal effects are evaluated against the top 5\% and 10\% genes with the most changes in their expression levels in the knock-off experiment, where each time one of the 5360 genes was knocked off and the changes (before and after the knock-off) of the expression levels of the other genes were measured. The ROC curves showed that the causal effect rank picks up more top 5\% and 10\% most changed genes in the experiments than regression coefficient (association) ranks by Lasso and elastic net~\cite{tibshirani1996regression} for all $k$ genes, respectively.  

Le et al.~\cite{le2013inferring} used IDA on observational gene expression data to find the target genes (i.e. genes regulated by) of the two microRNAs, miR200a and miR200b, and evaluated the results using experiments which knocked off miR200a and miR200b respectively. The top 20, 50 and 100 genes with the highest causal effects contain a significant proportion of genes known to be regulated by miR200a and miR200b in the evaluation experiments. Moreover, using the signs of causal effects to indicate an increase or decrease of gene expression after knocking down a microRNA, the signs of causal effects by IDA are 95\% and 94\% consistent with the top 100 genes ranked by their absolute causal effects with miR200a and miR200b respectively.

 Ren et al. \cite{rengraphical2023} meticulously examined and visualized the intricate relationships encompassing image features, patient demographics, and clinicopathological variables. Employing an innovative score-based directed graph termed "Grouped Greedy Equivalence Search" (GGES), they take into account existing knowledge to map these associations. By meticulously refining and handpicking causal variables, they compute the IDA scores \cite{maathuis2009estimating}, thereby effectively quantifying the impact of each variable on patients' postoperative survival. Leveraging these IDA scores, a predictive formula is derived using GGES, facilitating the projection of postoperative survival outcomes for individuals afflicted with esophageal cancer.

 Sun et al. \cite{sun2021identification} aim to identify potential prognostic genes within the prostate adenocarcinoma microenvironment and simultaneously estimate their causal effects. To determine the minimal sets of confounding covariates between candidate causative genes and the biochemical recurrence (BCR) status, the R package CovSel \cite{haggstrom2015covsel} was employed for screening all candidate causative genes and clinical covariates. In their experiments, validation is performed on five genes using another prostate adenocarcinoma cohort (GEO: GSE70770). These findings have the potential to contribute to an enhanced prognosis for prostate adenocarcinoma.

Sieswerda et al. \cite{sieswerda2023identifying} have developed a (causal) Bayesian Network (BN) through a structure learning method that expands upon the idea of identifying an adjustment set by learning the causal Bayesian Network structure~\cite{de2011covariate, haggstrom2018data}. Their emphasis is on elucidating causal relationships while disregarding non-causal associations in causal structure learning. This study showcases how structure learning, guided by clinical insights, can be harnessed to derive a causal model. The predictions of the developed BN indicate a treatment effect of 1 percentage point at 10 years in localized prostate cancer.

 Mendelian randomisation (MR) analysis, employing genetic variants as IVs for causal effect estimation in the presence of latent variables, has gained significant attention~\cite{kang2016instrumental,brumpton2020avoiding}. In MR studies, the reliability of conclusions heavily depends on the validity of the genetic variants as IVs~\cite{burgess2020robust}. sisVIVE does not require that all IVs are valid, and has been used in many MR studies~\cite{allard2015mendelian,brumpton2020avoiding}.

\paragraph{Potential}  Causal effect estimation based on graphical causal modelling provides valuable insights into the causal mechanisms underlying complex systems and is rapidly evolving into a mature scientific tool~\cite{pearl2018book,van2019separators}. In practice, causal effect estimation finds numerous applications in real-world scenarios, especially when dealing with latent variables. For example, Sobel and Lindquist define and estimate both ``point'' and ``cumulated'' effects for brain regions in functional magnetic resonance imaging (fMRI)~\cite{sobel2020estimating}. Runge et al.~\cite{runge2019inferring} present an overview of causal inference frameworks for Earth system sciences and identify promising generic application cases. Griffith et al.~\cite{griffith2020collider} explore the collider bias problem in the context of COVID-19 disease, and they effectively mitigate the bias using appropriate sampling strategies.

\subsection{Challenges in Evaluation}
\label{subsec:aande}
Due to the relative newness of the methods, currently there are limited applications of data-driven methods in real-world applications. Some typical examples of applications are discussed in the previous subsection.  Most other algorithms have been evaluated with some real-world datasets such as Job training~\cite{lalonde1986evaluating}, Sachs~\cite{sachs2005causal}, 401 (k)~\cite{verbeek2008guide} and Schoolingreturns~\cite{card1993using}. However, these real-world datasets normally have a small number of variables and/or data samples. The true potential of data-driven methods has not been fully evaluated. Another challenge for real-world evaluation is the lack of ground truth. In real-world datasets, there is not ground truth of causal effects nor the underlying causal DAG (or MAG) such that there is not a baseline to compare the performance of the data-driven causal effect estimation methods.

In most literature, synthetic datasets with ground truth causal effects and causal structures are generated
for evaluating the performance of data-driven causal effect estimation methods. The two
semi-synthetic datasets, IHDP~\cite{hill2011bayesian} and Twins~\cite{louizos2017causal} are also commonly used to evaluate data-driven causal effect
estimation methods. For real-world datasets without ground truths, the empirical estimates in the literature
are usually used as substitutes for the ground truths since the empirical estimates are likely to be consistent with domain knowledge.

It should be noted that it is impossible to use cross-validation to validate causal effect estimation and it is not possible to use prediction accuracy or AUC (Area Under the Curve) to justify the validity of the estimated causal effects~\cite{li2020accurate}.

\section{Conclusion}
\label{Sec:conclusion}
 The estimation of causal effects from observational data is a fundamental task in causal inference. It is critical to remove  confounding bias for obtaining unbiased causal effect estimation even in the presence of latent variables. Graphical causal models have become widely used in many areas to represent the underlying mechanisms that generate the data. By taking advantage of graphical causal modelling, many data-driven methods have been developed for causal effect estimation. In this survey, we have reviewed the theories and assumptions for data-driven causal effect estimation under graphical causal modelling framework.  We have identified three challenges, uncertainty in causal structure learning, time complexity and latent variable in causal effect estimation from observational data. We categorise methods by the ways they handle the challenges, and have provided a comprehensive review of the methods based on graphical causal modelling.   We have also discussed the practical issues of using the methods, offered guidance, assessed the performance of these methods, and examined their time complexity. We focus on the theories and assumptions for data-driven methods based on graphical causal modelling. We hope that the review will enable researchers and practitioners to understand the strengths and limitations of the methods from the theoretical and assumption viewpoint, and support more research in this challenging area.  


\section{Acknowledgements}
We thank the action editor and the reviewers for their valuable comments. We wish to acknowledge the support
from the Australian Research Council (under grant DP200101210).

\bibliographystyle{ACM-Reference-Format}
\bibliography{suvery2022}


\begin{thebibliography}{165}


\ifx \showCODEN    \undefined \def \showCODEN     #1{\unskip}     \fi
\ifx \showDOI      \undefined \def \showDOI       #1{#1}\fi
\ifx \showISBNx    \undefined \def \showISBNx     #1{\unskip}     \fi
\ifx \showISBNxiii \undefined \def \showISBNxiii  #1{\unskip}     \fi
\ifx \showISSN     \undefined \def \showISSN      #1{\unskip}     \fi
\ifx \showLCCN     \undefined \def \showLCCN      #1{\unskip}     \fi
\ifx \shownote     \undefined \def \shownote      #1{#1}          \fi
\ifx \showarticletitle \undefined \def \showarticletitle #1{#1}   \fi
\ifx \showURL      \undefined \def \showURL       {\relax}        \fi
\providecommand\bibfield[2]{#2}
\providecommand\bibinfo[2]{#2}
\providecommand\natexlab[1]{#1}
\providecommand\showeprint[2][]{arXiv:#2}

\bibitem[Abadie(2003)]%
        {abadie2003semiparametric}
\bibfield{author}{\bibinfo{person}{Alberto Abadie}.}
  \bibinfo{year}{2003}\natexlab{}.
\newblock \showarticletitle{Semiparametric instrumental variable estimation of
  treatment response models}.
\newblock \bibinfo{journal}{\emph{Journal of Econometrics}}
  \bibinfo{volume}{113}, \bibinfo{number}{2} (\bibinfo{year}{2003}),
  \bibinfo{pages}{231--263}.
\newblock


\bibitem[Abadie and Imbens(2016)]%
        {abadie2016matching}
\bibfield{author}{\bibinfo{person}{Alberto Abadie} {and}
  \bibinfo{person}{Guido~W Imbens}.} \bibinfo{year}{2016}\natexlab{}.
\newblock \showarticletitle{Matching on the estimated propensity score}.
\newblock \bibinfo{journal}{\emph{Econometrica}} \bibinfo{volume}{84},
  \bibinfo{number}{2} (\bibinfo{year}{2016}), \bibinfo{pages}{781--807}.
\newblock


\bibitem[Agrawal and Srikant(1994)]%
        {agrawal1994fast}
\bibfield{author}{\bibinfo{person}{Rakesh Agrawal} {and}
  \bibinfo{person}{Ramakrishnan Srikant}.} \bibinfo{year}{1994}\natexlab{}.
\newblock \showarticletitle{Fast Algorithms for Mining Association Rules in
  Large Databases}. In \bibinfo{booktitle}{\emph{The 20th International
  Conference on Very Large Data Bases}}. \bibinfo{pages}{487--499}.
\newblock


\bibitem[Ali et~al\mbox{.}(2009)]%
        {ali2009markov}
\bibfield{author}{\bibinfo{person}{R~Ayesha Ali}, \bibinfo{person}{Thomas~S
  Richardson}, \bibinfo{person}{Peter Spirtes}, {et~al\mbox{.}}}
  \bibinfo{year}{2009}\natexlab{}.
\newblock \showarticletitle{Markov equivalence for ancestral graphs}.
\newblock \bibinfo{journal}{\emph{The Annals of Statistics}}
  \bibinfo{volume}{37}, \bibinfo{number}{5B} (\bibinfo{year}{2009}),
  \bibinfo{pages}{2808--2837}.
\newblock


\bibitem[Aliferis et~al\mbox{.}(2010)]%
        {aliferis2010local}
\bibfield{author}{\bibinfo{person}{Constantin~F Aliferis},
  \bibinfo{person}{Alexander Statnikov}, \bibinfo{person}{Ioannis Tsamardinos},
  {et~al\mbox{.}}} \bibinfo{year}{2010}\natexlab{}.
\newblock \showarticletitle{Local causal and {M}arkov blanket induction for
  causal discovery and feature selection for classification part i: Algorithms
  and empirical evaluation}.
\newblock \bibinfo{journal}{\emph{Journal of Machine Learning Research}}
  \bibinfo{volume}{11}, \bibinfo{number}{Jan} (\bibinfo{year}{2010}),
  \bibinfo{pages}{171--234}.
\newblock


\bibitem[Allard et~al\mbox{.}(2015)]%
        {allard2015mendelian}
\bibfield{author}{\bibinfo{person}{C Allard}, \bibinfo{person}{V Desgagn{\'e}},
  {et~al\mbox{.}}} \bibinfo{year}{2015}\natexlab{}.
\newblock \showarticletitle{Mendelian randomization supports causality between
  maternal hyperglycemia and epigenetic regulation of leptin gene in newborns}.
\newblock \bibinfo{journal}{\emph{Epigenetics}} \bibinfo{volume}{10},
  \bibinfo{number}{4} (\bibinfo{year}{2015}), \bibinfo{pages}{342--351}.
\newblock


\bibitem[Angrist and Imbens(1995)]%
        {angrist1995two}
\bibfield{author}{\bibinfo{person}{Joshua~D Angrist} {and}
  \bibinfo{person}{Guido~W Imbens}.} \bibinfo{year}{1995}\natexlab{}.
\newblock \showarticletitle{Two-stage least squares estimation of average
  causal effects in models with variable treatment intensity}.
\newblock \bibinfo{journal}{\emph{J. Amer. Statist. Assoc.}}
  \bibinfo{volume}{90}, \bibinfo{number}{430} (\bibinfo{year}{1995}),
  \bibinfo{pages}{431--442}.
\newblock


\bibitem[Angrist et~al\mbox{.}(1996)]%
        {angrist1996}
\bibfield{author}{\bibinfo{person}{Joshua~D Angrist}, \bibinfo{person}{Guido~W
  Imbens}, {and} \bibinfo{person}{Donald~B Rubin}.}
  \bibinfo{year}{1996}\natexlab{}.
\newblock \showarticletitle{Identification of causal effects using instrumental
  variables}.
\newblock \bibinfo{journal}{\emph{J. Amer. Statist. Assoc.}}
  \bibinfo{volume}{91}, \bibinfo{number}{434} (\bibinfo{year}{1996}),
  \bibinfo{pages}{444--455}.
\newblock


\bibitem[Arellano and Bover(1995)]%
        {arellano1995another}
\bibfield{author}{\bibinfo{person}{Manuel Arellano} {and}
  \bibinfo{person}{Olympia Bover}.} \bibinfo{year}{1995}\natexlab{}.
\newblock \showarticletitle{Another look at the instrumental variable
  estimation of error-components models}.
\newblock \bibinfo{journal}{\emph{Journal of Econometrics}}
  \bibinfo{volume}{68}, \bibinfo{number}{1} (\bibinfo{year}{1995}),
  \bibinfo{pages}{29--51}.
\newblock


\bibitem[Athey and Imbens(2016)]%
        {athey2016recursive}
\bibfield{author}{\bibinfo{person}{Susan Athey} {and} \bibinfo{person}{Guido
  Imbens}.} \bibinfo{year}{2016}\natexlab{}.
\newblock \showarticletitle{Recursive partitioning for heterogeneous causal
  effects}.
\newblock \bibinfo{journal}{\emph{Proceedings of the National Academy of
  Sciences}} \bibinfo{volume}{113}, \bibinfo{number}{27}
  (\bibinfo{year}{2016}), \bibinfo{pages}{7353--7360}.
\newblock


\bibitem[Athey et~al\mbox{.}(2018)]%
        {athey2018approximate}
\bibfield{author}{\bibinfo{person}{Susan Athey}, \bibinfo{person}{Guido~W
  Imbens}, {and} \bibinfo{person}{Stefan Wager}.}
  \bibinfo{year}{2018}\natexlab{}.
\newblock \showarticletitle{Approximate residual balancing: debiased inference
  of average treatment effects in high dimensions}.
\newblock \bibinfo{journal}{\emph{Journal of the Royal Statistical Society:
  Series B (Statistical Methodology)}} \bibinfo{volume}{80},
  \bibinfo{number}{4} (\bibinfo{year}{2018}), \bibinfo{pages}{597--623}.
\newblock


\bibitem[Athey et~al\mbox{.}(2019)]%
        {athey2019generalized}
\bibfield{author}{\bibinfo{person}{Susan Athey}, \bibinfo{person}{Julie
  Tibshirani}, {and} \bibinfo{person}{Stefan Wager}.}
  \bibinfo{year}{2019}\natexlab{}.
\newblock \showarticletitle{Generalized random forests}.
\newblock \bibinfo{journal}{\emph{The Annals of Statistics}}
  \bibinfo{volume}{47}, \bibinfo{number}{2} (\bibinfo{year}{2019}),
  \bibinfo{pages}{1148--1178}.
\newblock


\bibitem[Baiocchi et~al\mbox{.}(2014)]%
        {baiocchi2014instrumental}
\bibfield{author}{\bibinfo{person}{Michael Baiocchi}, \bibinfo{person}{Jing
  Cheng}, {and} \bibinfo{person}{Dylan~S Small}.}
  \bibinfo{year}{2014}\natexlab{}.
\newblock \showarticletitle{Instrumental variable methods for causal
  inference}.
\newblock \bibinfo{journal}{\emph{Statistics in Medicine}}
  \bibinfo{volume}{33}, \bibinfo{number}{13} (\bibinfo{year}{2014}),
  \bibinfo{pages}{2297--2340}.
\newblock


\bibitem[Bang and Robins(2005)]%
        {bang2005doubly}
\bibfield{author}{\bibinfo{person}{Heejung Bang} {and} \bibinfo{person}{James~M
  Robins}.} \bibinfo{year}{2005}\natexlab{}.
\newblock \showarticletitle{Doubly robust estimation in missing data and causal
  inference models}.
\newblock \bibinfo{journal}{\emph{Biometrics}} \bibinfo{volume}{61},
  \bibinfo{number}{4} (\bibinfo{year}{2005}), \bibinfo{pages}{962--973}.
\newblock


\bibitem[Bareinboim et~al\mbox{.}(2014)]%
        {bareinboim2014recovering}
\bibfield{author}{\bibinfo{person}{Elias Bareinboim}, \bibinfo{person}{Jin
  Tian}, {and} \bibinfo{person}{Judea Pearl}.} \bibinfo{year}{2014}\natexlab{}.
\newblock \showarticletitle{Recovering from selection bias in causal and
  statistical inference}. In \bibinfo{booktitle}{\emph{The Twenty-Eighth AAAI
  Conference on Artificial Intelligence}}. \bibinfo{pages}{2410--2416}.
\newblock


\bibitem[Benkeser et~al\mbox{.}(2017)]%
        {benkeser2017doubly}
\bibfield{author}{\bibinfo{person}{David Benkeser}, \bibinfo{person}{Marco
  Carone}, \bibinfo{person}{MJ~Van~Der Laan}, {and} \bibinfo{person}{PB
  Gilbert}.} \bibinfo{year}{2017}\natexlab{}.
\newblock \showarticletitle{Doubly robust nonparametric inference on the
  average treatment effect}.
\newblock \bibinfo{journal}{\emph{Biometrika}} \bibinfo{volume}{104},
  \bibinfo{number}{4} (\bibinfo{year}{2017}), \bibinfo{pages}{863--880}.
\newblock


\bibitem[Bennett et~al\mbox{.}(2019)]%
        {bennett2019deep}
\bibfield{author}{\bibinfo{person}{Andrew Bennett}, \bibinfo{person}{Nathan
  Kallus}, {and} \bibinfo{person}{Tobias Schnabel}.}
  \bibinfo{year}{2019}\natexlab{}.
\newblock \showarticletitle{Deep generalized method of moments for instrumental
  variable analysis}. In \bibinfo{booktitle}{\emph{Advances in neural
  information processing systems}}. \bibinfo{pages}{3564--3574}.
\newblock


\bibitem[Bhattacharya and Nabi(2022)]%
        {bhattacharya2022testability}
\bibfield{author}{\bibinfo{person}{Rohit Bhattacharya} {and}
  \bibinfo{person}{Razieh Nabi}.} \bibinfo{year}{2022}\natexlab{}.
\newblock \showarticletitle{On testability of the front-door model via verma
  constraints}. In \bibinfo{booktitle}{\emph{Uncertainty in Artificial
  Intelligence}}. PMLR, \bibinfo{pages}{202--212}.
\newblock


\bibitem[Bhattacharya et~al\mbox{.}(2020)]%
        {bhattacharya2020semiparametric}
\bibfield{author}{\bibinfo{person}{Rohit Bhattacharya}, \bibinfo{person}{Razieh
  Nabi}, {and} \bibinfo{person}{Ilya Shpitser}.}
  \bibinfo{year}{2020}\natexlab{}.
\newblock \showarticletitle{Semiparametric Inference For Causal Effects In
  Graphical Models With Hidden Variables}.
\newblock \bibinfo{journal}{\emph{Stat}}  \bibinfo{volume}{1050}
  (\bibinfo{year}{2020}), \bibinfo{pages}{27}.
\newblock


\bibitem[Bowden and Turkington(1990)]%
        {bowden1990instrumental}
\bibfield{author}{\bibinfo{person}{Roger~J Bowden} {and}
  \bibinfo{person}{Darrell~A Turkington}.} \bibinfo{year}{1990}\natexlab{}.
\newblock \bibinfo{booktitle}{\emph{Instrumental {V}ariables}}.
  Vol.~\bibinfo{volume}{8}.
\newblock \bibinfo{publisher}{Cambridge university press}.
\newblock


\bibitem[Brito and Pearl(2002)]%
        {brito2002generalized}
\bibfield{author}{\bibinfo{person}{Carlos Brito} {and} \bibinfo{person}{Judea
  Pearl}.} \bibinfo{year}{2002}\natexlab{}.
\newblock \showarticletitle{Generalized instrumental variables}. In
  \bibinfo{booktitle}{\emph{The Conference on Uncertainty in Artificial
  Intelligence}}. \bibinfo{pages}{85--93}.
\newblock


\bibitem[Brumpton et~al\mbox{.}(2020)]%
        {brumpton2020avoiding}
\bibfield{author}{\bibinfo{person}{Ben Brumpton}, \bibinfo{person}{Eleanor
  Sanderson}, {et~al\mbox{.}}} \bibinfo{year}{2020}\natexlab{}.
\newblock \showarticletitle{Avoiding dynastic, assortative mating, and
  population stratification biases in Mendelian randomization through
  within-family analyses}.
\newblock \bibinfo{journal}{\emph{Nature communications}} \bibinfo{volume}{11},
  \bibinfo{number}{1} (\bibinfo{year}{2020}), \bibinfo{pages}{3519}.
\newblock


\bibitem[Burgess et~al\mbox{.}(2020)]%
        {burgess2020robust}
\bibfield{author}{\bibinfo{person}{Stephen Burgess},
  \bibinfo{person}{Christopher~N Foley}, \bibinfo{person}{Elias Allara},
  \bibinfo{person}{James~R Staley}, {and} \bibinfo{person}{Joanna~MM Howson}.}
  \bibinfo{year}{2020}\natexlab{}.
\newblock \showarticletitle{A robust and efficient method for Mendelian
  randomization with hundreds of genetic variants}.
\newblock \bibinfo{journal}{\emph{Nature communications}} \bibinfo{volume}{11},
  \bibinfo{number}{1} (\bibinfo{year}{2020}), \bibinfo{pages}{376}.
\newblock


\bibitem[Card(1993)]%
        {card1993using}
\bibfield{author}{\bibinfo{person}{David Card}.}
  \bibinfo{year}{1993}\natexlab{}.
\newblock \showarticletitle{Using Geographic Variation in College Proximity to
  Estimate the Return to Schooling}. In
  \bibinfo{booktitle}{\emph{Econometrica}}, Vol.~\bibinfo{volume}{69}.
  Citeseer, \bibinfo{pages}{1127--1160}.
\newblock


\bibitem[Cheng et~al\mbox{.}(2020)]%
        {cheng2020causal}
\bibfield{author}{\bibinfo{person}{Debo Cheng}, \bibinfo{person}{Jiuyong Li},
  \bibinfo{person}{Lin Liu}, {et~al\mbox{.}}} \bibinfo{year}{2020}\natexlab{}.
\newblock \showarticletitle{Causal query in observational data with hidden
  variables}. In \bibinfo{booktitle}{\emph{24th European Conference on
  Artificial Intelligence}}. \bibinfo{pages}{2551--2558}.
\newblock


\bibitem[Cheng et~al\mbox{.}(2022a)]%
        {cheng2022ancestral}
\bibfield{author}{\bibinfo{person}{Debo Cheng}, \bibinfo{person}{Jiuyong Li},
  \bibinfo{person}{Lin Liu}, {et~al\mbox{.}}} \bibinfo{year}{2022}\natexlab{a}.
\newblock \showarticletitle{Ancestral Instrument Method for Causal Inference
  without Complete Knowledge}. In \bibinfo{booktitle}{\emph{International Joint
  Conference on Artificial Intelligence}}. \bibinfo{pages}{4843--4849}.
\newblock


\bibitem[Cheng et~al\mbox{.}(2022b)]%
        {cheng2022sufficient}
\bibfield{author}{\bibinfo{person}{Debo Cheng}, \bibinfo{person}{Jiuyong Li},
  \bibinfo{person}{Lin Liu}, {et~al\mbox{.}}} \bibinfo{year}{2022}\natexlab{b}.
\newblock \showarticletitle{Sufficient dimension reduction for average causal
  effect estimation}.
\newblock \bibinfo{journal}{\emph{Data Mining and Knowledge Discovery}}
  \bibinfo{volume}{36}, \bibinfo{number}{3} (\bibinfo{year}{2022}),
  \bibinfo{pages}{1174--1196}.
\newblock


\bibitem[Cheng et~al\mbox{.}(2022c)]%
        {cheng2022toward}
\bibfield{author}{\bibinfo{person}{Debo Cheng}, \bibinfo{person}{Jiuyong Li},
  \bibinfo{person}{Lin Liu}, {et~al\mbox{.}}} \bibinfo{year}{2022}\natexlab{c}.
\newblock \showarticletitle{Toward Unique and Unbiased Causal Effect Estimation
  From Data With Hidden Variables}.
\newblock \bibinfo{journal}{\emph{IEEE Transactions on Neural Networks and
  Learning Systems}} \bibinfo{volume}{34}, \bibinfo{number}{11}
  (\bibinfo{year}{2022}), \bibinfo{pages}{1--13}.
\newblock


\bibitem[Cheng et~al\mbox{.}(2023a)]%
        {cheng2022discovering}
\bibfield{author}{\bibinfo{person}{Debo Cheng}, \bibinfo{person}{Jiuyong Li},
  \bibinfo{person}{Lin Liu}, {et~al\mbox{.}}} \bibinfo{year}{2023}\natexlab{a}.
\newblock \showarticletitle{Discovering Ancestral Instrumental Variables for
  Causal Inference from Observational Data}.
\newblock \bibinfo{journal}{\emph{IEEE Transactions on Neural Networks and
  Learning Systems}} (\bibinfo{year}{2023}), \bibinfo{pages}{1--11}.
\newblock


\bibitem[Cheng et~al\mbox{.}(2023b)]%
        {cheng2020local}
\bibfield{author}{\bibinfo{person}{Debo Cheng}, \bibinfo{person}{Jiuyong Li},
  \bibinfo{person}{Lin Liu}, {et~al\mbox{.}}} \bibinfo{year}{2023}\natexlab{b}.
\newblock \showarticletitle{Local Search for Efficient Causal Effect
  Estimation}.
\newblock \bibinfo{journal}{\emph{IEEE Transactions on Knowledge and Data
  Engineering}} \bibinfo{volume}{35}, \bibinfo{number}{9}
  (\bibinfo{year}{2023}), \bibinfo{pages}{8823--8837}.
\newblock
\urldef\tempurl%
\url{https://doi.org/10.1109/TKDE.2022.3218131}
\showDOI{\tempurl}


\bibitem[Chickering(1996)]%
        {chickering1996learning}
\bibfield{author}{\bibinfo{person}{David~Maxwell Chickering}.}
  \bibinfo{year}{1996}\natexlab{}.
\newblock \showarticletitle{Learning {B}ayesian networks is {NP}-complete}.
\newblock In \bibinfo{booktitle}{\emph{Learning from {D}ata}}.
  \bibinfo{publisher}{Springer}, \bibinfo{pages}{121--130}.
\newblock


\bibitem[Chickering(2002)]%
        {chickering2002learning}
\bibfield{author}{\bibinfo{person}{David~Maxwell Chickering}.}
  \bibinfo{year}{2002}\natexlab{}.
\newblock \showarticletitle{Learning equivalence classes of {B}ayesian-network
  structures}.
\newblock \bibinfo{journal}{\emph{Journal of Machine Learning Research}}
  \bibinfo{volume}{2}, \bibinfo{number}{Feb} (\bibinfo{year}{2002}),
  \bibinfo{pages}{445--498}.
\newblock


\bibitem[Christopher~Frey and Patil(2002)]%
        {christopher2002identification}
\bibfield{author}{\bibinfo{person}{H Christopher~Frey} {and}
  \bibinfo{person}{Sumeet~R Patil}.} \bibinfo{year}{2002}\natexlab{}.
\newblock \showarticletitle{Identification and review of sensitivity analysis
  methods}.
\newblock \bibinfo{journal}{\emph{Risk analysis}} \bibinfo{volume}{22},
  \bibinfo{number}{3} (\bibinfo{year}{2002}), \bibinfo{pages}{553--578}.
\newblock


\bibitem[Chu et~al\mbox{.}(2001)]%
        {chu2001semi}
\bibfield{author}{\bibinfo{person}{Tianjiao Chu}, \bibinfo{person}{Richard
  Scheines}, {and} \bibinfo{person}{Peter Spirtes}.}
  \bibinfo{year}{2001}\natexlab{}.
\newblock \showarticletitle{Semi-instrumental variables: a test for instrument
  admissibility}. In \bibinfo{booktitle}{\emph{The Conference on Uncertainty in
  Artificial Intelligence}}. \bibinfo{pages}{83--90}.
\newblock


\bibitem[Cole and Hern{\'a}n(2008)]%
        {cole2008constructing}
\bibfield{author}{\bibinfo{person}{Stephen~R Cole} {and}
  \bibinfo{person}{Miguel~A Hern{\'a}n}.} \bibinfo{year}{2008}\natexlab{}.
\newblock \showarticletitle{Constructing inverse probability weights for
  marginal structural models}.
\newblock \bibinfo{journal}{\emph{American journal of epidemiology}}
  \bibinfo{volume}{168}, \bibinfo{number}{6} (\bibinfo{year}{2008}),
  \bibinfo{pages}{656--664}.
\newblock


\bibitem[Colombo et~al\mbox{.}(2012)]%
        {colombo2012learning}
\bibfield{author}{\bibinfo{person}{Diego Colombo}, \bibinfo{person}{Marloes~H
  Maathuis}, \bibinfo{person}{Markus Kalisch}, {and} \bibinfo{person}{Thomas~S
  Richardson}.} \bibinfo{year}{2012}\natexlab{}.
\newblock \showarticletitle{Learning high-dimensional directed acyclic graphs
  with latent and selection variables}.
\newblock \bibinfo{journal}{\emph{The Annals of Statistics}}
  \bibinfo{volume}{40}, \bibinfo{number}{1} (\bibinfo{year}{2012}),
  \bibinfo{pages}{294--321}.
\newblock


\bibitem[Cornfield et~al\mbox{.}(1959)]%
        {cornfield1959smoking}
\bibfield{author}{\bibinfo{person}{Jerome Cornfield}, \bibinfo{person}{William
  Haenszel}, {et~al\mbox{.}}} \bibinfo{year}{1959}\natexlab{}.
\newblock \showarticletitle{Smoking and lung cancer: recent evidence and a
  discussion of some questions}.
\newblock \bibinfo{journal}{\emph{Journal of the National Cancer institute}}
  \bibinfo{volume}{22}, \bibinfo{number}{1} (\bibinfo{year}{1959}),
  \bibinfo{pages}{173--203}.
\newblock


\bibitem[Correa and Bareinboim(2017)]%
        {correa2017causal}
\bibfield{author}{\bibinfo{person}{Juan~D Correa} {and} \bibinfo{person}{Elias
  Bareinboim}.} \bibinfo{year}{2017}\natexlab{}.
\newblock \showarticletitle{Causal effect identification by adjustment under
  confounding and selection biases}. In \bibinfo{booktitle}{\emph{The
  Thirty-First AAAI Conference on Artificial Intelligence}}.
  \bibinfo{pages}{3740--3746}.
\newblock


\bibitem[De~Luna et~al\mbox{.}(2011)]%
        {de2011covariate}
\bibfield{author}{\bibinfo{person}{Xavier De~Luna}, \bibinfo{person}{Ingeborg
  Waernbaum}, {and} \bibinfo{person}{Thomas~S Richardson}.}
  \bibinfo{year}{2011}\natexlab{}.
\newblock \showarticletitle{Covariate selection for the nonparametric
  estimation of an average treatment effect}.
\newblock \bibinfo{journal}{\emph{Biometrika}} \bibinfo{volume}{98},
  \bibinfo{number}{4} (\bibinfo{year}{2011}), \bibinfo{pages}{861--875}.
\newblock


\bibitem[Deaton and Cartwright(2018)]%
        {deaton2018understanding}
\bibfield{author}{\bibinfo{person}{Angus Deaton} {and} \bibinfo{person}{Nancy
  Cartwright}.} \bibinfo{year}{2018}\natexlab{}.
\newblock \showarticletitle{Understanding and misunderstanding randomized
  controlled trials}.
\newblock \bibinfo{journal}{\emph{Social Science \& Medicine}}
  \bibinfo{volume}{210} (\bibinfo{year}{2018}), \bibinfo{pages}{2--21}.
\newblock


\bibitem[Ding and VanderWeele(2016)]%
        {ding2016sensitivity}
\bibfield{author}{\bibinfo{person}{Peng Ding} {and} \bibinfo{person}{Tyler~J
  VanderWeele}.} \bibinfo{year}{2016}\natexlab{}.
\newblock \showarticletitle{Sensitivity analysis without assumptions}.
\newblock \bibinfo{journal}{\emph{Epidemiology (Cambridge, Mass.)}}
  \bibinfo{volume}{27}, \bibinfo{number}{3} (\bibinfo{year}{2016}),
  \bibinfo{pages}{368}.
\newblock


\bibitem[Duarte et~al\mbox{.}(2021)]%
        {duarte2021automated}
\bibfield{author}{\bibinfo{person}{Guilherme Duarte}, \bibinfo{person}{Noam
  Finkelstein}, \bibinfo{person}{Dean Knox}, \bibinfo{person}{Jonathan
  Mummolo}, {and} \bibinfo{person}{Ilya Shpitser}.}
  \bibinfo{year}{2021}\natexlab{}.
\newblock \showarticletitle{An automated approach to causal inference in
  discrete settings}.
\newblock \bibinfo{journal}{\emph{arXiv preprint arXiv:2109.13471}}
  (\bibinfo{year}{2021}).
\newblock


\bibitem[Entner et~al\mbox{.}(2013)]%
        {entner2013data}
\bibfield{author}{\bibinfo{person}{Doris Entner}, \bibinfo{person}{Patrik
  Hoyer}, {and} \bibinfo{person}{Peter Spirtes}.}
  \bibinfo{year}{2013}\natexlab{}.
\newblock \showarticletitle{Data-driven covariate selection for nonparametric
  estimation of causal effects}. In \bibinfo{booktitle}{\emph{Artificial
  Intelligence and Statistics}}. \bibinfo{pages}{256--264}.
\newblock


\bibitem[Evans and Richardson(2014)]%
        {evans2014markovian}
\bibfield{author}{\bibinfo{person}{Robin~J Evans} {and}
  \bibinfo{person}{Thomas~S Richardson}.} \bibinfo{year}{2014}\natexlab{}.
\newblock \showarticletitle{Markovian acyclic directed mixed graphs for
  discrete data}.
\newblock \bibinfo{journal}{\emph{The Annals of Statistics}}
  \bibinfo{volume}{42}, \bibinfo{number}{4} (\bibinfo{year}{2014}),
  \bibinfo{pages}{1452--1482}.
\newblock


\bibitem[Fang and He(2020)]%
        {fang2020ida}
\bibfield{author}{\bibinfo{person}{Zhuangyan Fang} {and}
  \bibinfo{person}{Yangbo He}.} \bibinfo{year}{2020}\natexlab{}.
\newblock \showarticletitle{IDA with {B}ackground {K}nowledge}. In
  \bibinfo{booktitle}{\emph{Conference on Uncertainty in Artificial
  Intelligence}}. PMLR, \bibinfo{pages}{270--279}.
\newblock


\bibitem[Funk et~al\mbox{.}(2011)]%
        {funk2011doubly}
\bibfield{author}{\bibinfo{person}{Michele~Jonsson Funk},
  \bibinfo{person}{Daniel Westreich}, \bibinfo{person}{Chris Wiesen},
  {et~al\mbox{.}}} \bibinfo{year}{2011}\natexlab{}.
\newblock \showarticletitle{Doubly robust estimation of causal effects}.
\newblock \bibinfo{journal}{\emph{American journal of epidemiology}}
  \bibinfo{volume}{173}, \bibinfo{number}{7} (\bibinfo{year}{2011}),
  \bibinfo{pages}{761--767}.
\newblock


\bibitem[Glymour et~al\mbox{.}(2019)]%
        {glymour2019review}
\bibfield{author}{\bibinfo{person}{C Glymour}, \bibinfo{person}{K Zhang}, {and}
  \bibinfo{person}{P Spirtes}.} \bibinfo{year}{2019}\natexlab{}.
\newblock \showarticletitle{Review of Causal Discovery Methods Based on
  Graphical Models.}
\newblock \bibinfo{journal}{\emph{Frontiers in Genetics}}  \bibinfo{volume}{10}
  (\bibinfo{year}{2019}), \bibinfo{pages}{524--524}.
\newblock


\bibitem[Glymour and Cooper(1999)]%
        {glymour1999computation}
\bibfield{author}{\bibinfo{person}{Clark~N Glymour} {and}
  \bibinfo{person}{Gregory~Floyd Cooper}.} \bibinfo{year}{1999}\natexlab{}.
\newblock \bibinfo{booktitle}{\emph{Computation, {C}ausation, and
  {D}iscovery}}.
\newblock \bibinfo{publisher}{AAAI Press}.
\newblock


\bibitem[Glymour et~al\mbox{.}(2008)]%
        {glymour2008methodological}
\bibfield{author}{\bibinfo{person}{M~Maria Glymour}, \bibinfo{person}{Jennifer
  Weuve}, {and} \bibinfo{person}{Jarvis~T Chen}.}
  \bibinfo{year}{2008}\natexlab{}.
\newblock \showarticletitle{Methodological challenges in causal research on
  racial and ethnic patterns of cognitive trajectories: measurement, selection,
  and bias}.
\newblock \bibinfo{journal}{\emph{Neuropsychology review}}
  \bibinfo{volume}{18}, \bibinfo{number}{3} (\bibinfo{year}{2008}),
  \bibinfo{pages}{194--213}.
\newblock


\bibitem[Gradu et~al\mbox{.}(2022)]%
        {gradu2022valid}
\bibfield{author}{\bibinfo{person}{Paula Gradu}, \bibinfo{person}{Tijana
  Zrnic}, \bibinfo{person}{Yixin Wang}, {and} \bibinfo{person}{Michael~I
  Jordan}.} \bibinfo{year}{2022}\natexlab{}.
\newblock \showarticletitle{Valid Inference after Causal Discovery}.
\newblock \bibinfo{journal}{\emph{arXiv preprint arXiv:2208.05949}}
  (\bibinfo{year}{2022}).
\newblock


\bibitem[Greene(2003)]%
        {greene2003econometric}
\bibfield{author}{\bibinfo{person}{William~H Greene}.}
  \bibinfo{year}{2003}\natexlab{}.
\newblock \bibinfo{booktitle}{\emph{Econometric {A}nalysis}}.
\newblock \bibinfo{publisher}{Pearson Education India}.
\newblock


\bibitem[Greenland(2003)]%
        {greenland2003quantifying}
\bibfield{author}{\bibinfo{person}{Sander Greenland}.}
  \bibinfo{year}{2003}\natexlab{}.
\newblock \showarticletitle{Quantifying biases in causal models: classical
  confounding vs collider-stratification bias}.
\newblock \bibinfo{journal}{\emph{Epidemiology}} \bibinfo{volume}{14},
  \bibinfo{number}{3} (\bibinfo{year}{2003}), \bibinfo{pages}{300--306}.
\newblock


\bibitem[Greenland et~al\mbox{.}(1999)]%
        {greenland1999causal}
\bibfield{author}{\bibinfo{person}{Sander Greenland}, \bibinfo{person}{Judea
  Pearl}, {and} \bibinfo{person}{James~M Robins}.}
  \bibinfo{year}{1999}\natexlab{}.
\newblock \showarticletitle{Causal diagrams for epidemiologic research}.
\newblock \bibinfo{journal}{\emph{Epidemiology}} \bibinfo{volume}{10},
  \bibinfo{number}{1} (\bibinfo{year}{1999}), \bibinfo{pages}{37--48}.
\newblock


\bibitem[Griffith et~al\mbox{.}(2020)]%
        {griffith2020collider}
\bibfield{author}{\bibinfo{person}{Gareth~J Griffith}, \bibinfo{person}{Tim~T
  Morris}, \bibinfo{person}{Matthew~J Tudball}, \bibinfo{person}{Annie
  Herbert}, \bibinfo{person}{Giulia Mancano}, \bibinfo{person}{Lindsey Pike},
  \bibinfo{person}{Gemma~C Sharp}, \bibinfo{person}{Jonathan Sterne},
  \bibinfo{person}{Tom~M Palmer}, \bibinfo{person}{George Davey~Smith},
  {et~al\mbox{.}}} \bibinfo{year}{2020}\natexlab{}.
\newblock \showarticletitle{Collider bias undermines our understanding of
  COVID-19 disease risk and severity}.
\newblock \bibinfo{journal}{\emph{Nature communications}} \bibinfo{volume}{11},
  \bibinfo{number}{1} (\bibinfo{year}{2020}), \bibinfo{pages}{5749}.
\newblock


\bibitem[Guo et~al\mbox{.}(2020)]%
        {guo2020survey}
\bibfield{author}{\bibinfo{person}{Ruocheng Guo}, \bibinfo{person}{Lu Cheng},
  \bibinfo{person}{Jundong Li}, {et~al\mbox{.}}}
  \bibinfo{year}{2020}\natexlab{}.
\newblock \showarticletitle{A survey of learning causality with data: Problems
  and methods}.
\newblock \bibinfo{journal}{\emph{ACM Computing Surveys (CSUR)}}
  \bibinfo{volume}{53}, \bibinfo{number}{4} (\bibinfo{year}{2020}),
  \bibinfo{pages}{1--37}.
\newblock


\bibitem[H{\"a}ggstr{\"o}m(2018)]%
        {haggstrom2018data}
\bibfield{author}{\bibinfo{person}{Jenny H{\"a}ggstr{\"o}m}.}
  \bibinfo{year}{2018}\natexlab{}.
\newblock \showarticletitle{Data-driven confounder selection via {M}arkov and
  {B}ayesian networks}.
\newblock \bibinfo{journal}{\emph{Biometrics}} \bibinfo{volume}{74},
  \bibinfo{number}{2} (\bibinfo{year}{2018}), \bibinfo{pages}{389--398}.
\newblock


\bibitem[H{\"a}ggstr{\"o}m et~al\mbox{.}(2015)]%
        {haggstrom2015covsel}
\bibfield{author}{\bibinfo{person}{Jenny H{\"a}ggstr{\"o}m},
  \bibinfo{person}{Emma Persson}, \bibinfo{person}{Ingeborg Waernbaum}, {and}
  \bibinfo{person}{Xavier de Luna}.} \bibinfo{year}{2015}\natexlab{}.
\newblock \showarticletitle{CovSel: An {R} package for covariate selection when
  estimating average causal effects}.
\newblock \bibinfo{journal}{\emph{Journal of Statistical Software}}
  \bibinfo{volume}{68}, \bibinfo{number}{1} (\bibinfo{year}{2015}),
  \bibinfo{pages}{1--20}.
\newblock


\bibitem[Hartford et~al\mbox{.}(2017)]%
        {hartford2017deep}
\bibfield{author}{\bibinfo{person}{Jason Hartford}, \bibinfo{person}{Greg
  Lewis}, \bibinfo{person}{Kevin Leyton-Brown}, {and} \bibinfo{person}{Matt
  Taddy}.} \bibinfo{year}{2017}\natexlab{}.
\newblock \showarticletitle{Deep IV: A flexible approach for counterfactual
  prediction}. In \bibinfo{booktitle}{\emph{International Conference on Machine
  Learning}}. PMLR, \bibinfo{pages}{1414--1423}.
\newblock


\bibitem[Hartford et~al\mbox{.}(2021)]%
        {hartford2021valid}
\bibfield{author}{\bibinfo{person}{Jason~S Hartford}, \bibinfo{person}{Victor
  Veitch}, \bibinfo{person}{Dhanya Sridhar}, {and} \bibinfo{person}{Kevin
  Leyton-Brown}.} \bibinfo{year}{2021}\natexlab{}.
\newblock \showarticletitle{Valid causal inference with (some) invalid
  instruments}. In \bibinfo{booktitle}{\emph{International Conference on
  Machine Learning}}. PMLR, \bibinfo{pages}{4096--4106}.
\newblock


\bibitem[Heckman(2008)]%
        {heckman2008econometric}
\bibfield{author}{\bibinfo{person}{James~J Heckman}.}
  \bibinfo{year}{2008}\natexlab{}.
\newblock \showarticletitle{Econometric causality}.
\newblock \bibinfo{journal}{\emph{International statistical review}}
  \bibinfo{volume}{76}, \bibinfo{number}{1} (\bibinfo{year}{2008}),
  \bibinfo{pages}{1--27}.
\newblock


\bibitem[Henckel et~al\mbox{.}(2022)]%
        {henckel2022graphical}
\bibfield{author}{\bibinfo{person}{Leonard Henckel}, \bibinfo{person}{Emilija
  Perkovi{\'c}}, {and} \bibinfo{person}{Marloes~H Maathuis}.}
  \bibinfo{year}{2022}\natexlab{}.
\newblock \showarticletitle{Graphical criteria for efficient total effect
  estimation via adjustment in causal linear models}.
\newblock \bibinfo{journal}{\emph{Journal of the Royal Statistical Society
  Series B}} \bibinfo{volume}{84}, \bibinfo{number}{2} (\bibinfo{year}{2022}),
  \bibinfo{pages}{579--599}.
\newblock


\bibitem[Hern{\'a}n and Robins(2006)]%
        {hernan2006instruments}
\bibfield{author}{\bibinfo{person}{Miguel~A Hern{\'a}n} {and}
  \bibinfo{person}{James~M Robins}.} \bibinfo{year}{2006}\natexlab{}.
\newblock \showarticletitle{Instruments for causal inference: an
  epidemiologist's dream?}
\newblock \bibinfo{journal}{\emph{Epidemiology}} \bibinfo{volume}{17},
  \bibinfo{number}{4} (\bibinfo{year}{2006}), \bibinfo{pages}{360--372}.
\newblock


\bibitem[Hern{\'a}n and Robins(2020)]%
        {hernan2020causal}
\bibfield{author}{\bibinfo{person}{Miguel~A Hern{\'a}n} {and}
  \bibinfo{person}{James~M Robins}.} \bibinfo{year}{2020}\natexlab{}.
\newblock \bibinfo{booktitle}{\emph{Causal {I}nference, {W}hat {I}f}}.
\newblock \bibinfo{publisher}{CRC Boca Raton, FL;}.
\newblock


\bibitem[Hill(2011)]%
        {hill2011bayesian}
\bibfield{author}{\bibinfo{person}{Jennifer~L Hill}.}
  \bibinfo{year}{2011}\natexlab{}.
\newblock \showarticletitle{Bayesian nonparametric modeling for causal
  inference}.
\newblock \bibinfo{journal}{\emph{Journal of Computational and Graphical
  Statistics}} \bibinfo{volume}{20}, \bibinfo{number}{1}
  (\bibinfo{year}{2011}), \bibinfo{pages}{217--240}.
\newblock


\bibitem[Hirano et~al\mbox{.}(2003)]%
        {hirano2003efficient}
\bibfield{author}{\bibinfo{person}{Keisuke Hirano}, \bibinfo{person}{Guido~W
  Imbens}, {and} \bibinfo{person}{Geert Ridder}.}
  \bibinfo{year}{2003}\natexlab{}.
\newblock \showarticletitle{Efficient estimation of average treatment effects
  using the estimated propensity score}.
\newblock \bibinfo{journal}{\emph{Econometrica}} \bibinfo{volume}{71},
  \bibinfo{number}{4} (\bibinfo{year}{2003}), \bibinfo{pages}{1161--1189}.
\newblock


\bibitem[Hyttinen et~al\mbox{.}(2014)]%
        {hyttinen2014constraint}
\bibfield{author}{\bibinfo{person}{Antti Hyttinen}, \bibinfo{person}{Frederick
  Eberhardt}, {and} \bibinfo{person}{Matti J{\"a}rvisalo}.}
  \bibinfo{year}{2014}\natexlab{}.
\newblock \showarticletitle{Constraint-based Causal Discovery: Conflict
  Resolution with Answer Set Programming.}. In \bibinfo{booktitle}{\emph{The
  Conference on Uncertainty in Artificial Intelligence}}.
  \bibinfo{pages}{340--349}.
\newblock


\bibitem[Hyttinen et~al\mbox{.}(2015)]%
        {hyttinen2015calculus}
\bibfield{author}{\bibinfo{person}{Antti Hyttinen}, \bibinfo{person}{Frederick
  Eberhardt}, {and} \bibinfo{person}{Matti J{\"a}rvisalo}.}
  \bibinfo{year}{2015}\natexlab{}.
\newblock \showarticletitle{Do-calculus when the True Graph Is Unknown.}. In
  \bibinfo{booktitle}{\emph{The Conference on Uncertainty in Artificial
  Intelligence}}. Citeseer, \bibinfo{pages}{395--404}.
\newblock


\bibitem[Imai and Ratkovic(2014)]%
        {imai2014covariate}
\bibfield{author}{\bibinfo{person}{Kosuke Imai} {and} \bibinfo{person}{Marc
  Ratkovic}.} \bibinfo{year}{2014}\natexlab{}.
\newblock \showarticletitle{Covariate balancing propensity score}.
\newblock \bibinfo{journal}{\emph{Journal of the Royal Statistical Society:
  Series B (Statistical Methodology)}} \bibinfo{volume}{76},
  \bibinfo{number}{1} (\bibinfo{year}{2014}), \bibinfo{pages}{243--263}.
\newblock


\bibitem[Imbens(2014)]%
        {imbens2014instrumental}
\bibfield{author}{\bibinfo{person}{Guido~W Imbens}.}
  \bibinfo{year}{2014}\natexlab{}.
\newblock \showarticletitle{Instrumental Variables: An Econometrician’s
  Perspective}.
\newblock \bibinfo{journal}{\emph{Statist. Sci.}} \bibinfo{volume}{29},
  \bibinfo{number}{3} (\bibinfo{year}{2014}), \bibinfo{pages}{323--358}.
\newblock


\bibitem[Imbens(2020)]%
        {imbens2020potential}
\bibfield{author}{\bibinfo{person}{Guido~W Imbens}.}
  \bibinfo{year}{2020}\natexlab{}.
\newblock \showarticletitle{Potential outcome and directed acyclic graph
  approaches to causality: Relevance for empirical practice in economics}.
\newblock \bibinfo{journal}{\emph{Journal of Economic Literature}}
  \bibinfo{volume}{58}, \bibinfo{number}{4} (\bibinfo{year}{2020}),
  \bibinfo{pages}{1129--79}.
\newblock


\bibitem[Imbens and Rubin(2015)]%
        {imbens2015causal}
\bibfield{author}{\bibinfo{person}{Guido~W Imbens} {and}
  \bibinfo{person}{Donald~B Rubin}.} \bibinfo{year}{2015}\natexlab{}.
\newblock \bibinfo{booktitle}{\emph{Causal {I}nference in {S}tatistics,
  {S}ocial, and {B}iomedical {S}ciences}}.
\newblock \bibinfo{publisher}{Cambridge University Press}.
\newblock


\bibitem[Jaber et~al\mbox{.}(2019a)]%
        {jaber2019identification}
\bibfield{author}{\bibinfo{person}{Amin Jaber}, \bibinfo{person}{Jiji Zhang},
  {and} \bibinfo{person}{Elias Bareinboim}.} \bibinfo{year}{2019}\natexlab{a}.
\newblock \showarticletitle{Identification of conditional causal effects under
  markov equivalence}.
\newblock \bibinfo{journal}{\emph{Advances in Neural Information Processing
  Systems}}  \bibinfo{volume}{32} (\bibinfo{year}{2019}).
\newblock


\bibitem[Jaber et~al\mbox{.}(2019b)]%
        {jaber2019causal}
\bibfield{author}{\bibinfo{person}{Amin Jaber}, \bibinfo{person}{Jiji Zhang},
  {and} \bibinfo{person}{Elias Bareinboim}.} \bibinfo{year}{2019}\natexlab{b}.
\newblock \showarticletitle{On causal identification under Markov equivalence}.
  In \bibinfo{booktitle}{\emph{28th International Joint Conference on
  Artificial Intelligence, IJCAI 2019}}. International Joint Conferences on
  Artificial Intelligence, \bibinfo{pages}{6181--6185}.
\newblock


\bibitem[Kalisch et~al\mbox{.}(2012)]%
        {kalisch2012causal}
\bibfield{author}{\bibinfo{person}{Markus Kalisch}, \bibinfo{person}{Martin
  M{\"a}chler}, \bibinfo{person}{Diego Colombo}, {et~al\mbox{.}}}
  \bibinfo{year}{2012}\natexlab{}.
\newblock \showarticletitle{Causal inference using graphical models with the
  {R} package pcalg}.
\newblock \bibinfo{journal}{\emph{Journal of Statistical Software}}
  \bibinfo{volume}{47}, \bibinfo{number}{11} (\bibinfo{year}{2012}),
  \bibinfo{pages}{1--26}.
\newblock


\bibitem[Kang et~al\mbox{.}(2016)]%
        {kang2016instrumental}
\bibfield{author}{\bibinfo{person}{Hyunseung Kang}, \bibinfo{person}{Anru
  Zhang}, \bibinfo{person}{T~Tony Cai}, {and} \bibinfo{person}{Dylan~S Small}.}
  \bibinfo{year}{2016}\natexlab{}.
\newblock \showarticletitle{Instrumental variables estimation with some invalid
  instruments and its application to Mendelian randomization}.
\newblock \bibinfo{journal}{\emph{J. Amer. Statist. Assoc.}}
  \bibinfo{volume}{111}, \bibinfo{number}{513} (\bibinfo{year}{2016}),
  \bibinfo{pages}{132--144}.
\newblock


\bibitem[Koller and Friedman(2009)]%
        {koller2009probabilistic}
\bibfield{author}{\bibinfo{person}{Daphne Koller} {and} \bibinfo{person}{Nir
  Friedman}.} \bibinfo{year}{2009}\natexlab{}.
\newblock \bibinfo{booktitle}{\emph{Probabilistic {G}raphical {M}odels:
  {P}rinciples and {T}echniques}}.
\newblock \bibinfo{publisher}{MIT {P}ress}.
\newblock


\bibitem[Kuroki and Cai(2005)]%
        {kuroki2005instrumental}
\bibfield{author}{\bibinfo{person}{Manabu Kuroki} {and}
  \bibinfo{person}{Zhihong Cai}.} \bibinfo{year}{2005}\natexlab{}.
\newblock \showarticletitle{Instrumental variable tests for Directed Acyclic
  Graph Models.}. In \bibinfo{booktitle}{\emph{International Conference on
  Artificial Intelligence and Statistics}}. \bibinfo{pages}{190--197}.
\newblock


\bibitem[LaLonde(1986)]%
        {lalonde1986evaluating}
\bibfield{author}{\bibinfo{person}{Robert~J LaLonde}.}
  \bibinfo{year}{1986}\natexlab{}.
\newblock \showarticletitle{Evaluating the econometric evaluations of training
  programs with experimental data}.
\newblock \bibinfo{journal}{\emph{The American Economic Review}}
  \bibinfo{volume}{76}, \bibinfo{number}{4} (\bibinfo{year}{1986}),
  \bibinfo{pages}{604--620}.
\newblock


\bibitem[Le et~al\mbox{.}(2013)]%
        {le2013inferring}
\bibfield{author}{\bibinfo{person}{Thuc~Duy Le}, \bibinfo{person}{Lin Liu},
  \bibinfo{person}{Anna Tsykin}, {et~al\mbox{.}}}
  \bibinfo{year}{2013}\natexlab{}.
\newblock \showarticletitle{Inferring micro{RNA}--m{RNA} causal regulatory
  relationships from expression data}.
\newblock \bibinfo{journal}{\emph{Bioinformatics}} \bibinfo{volume}{29},
  \bibinfo{number}{6} (\bibinfo{year}{2013}), \bibinfo{pages}{765--771}.
\newblock


\bibitem[Li et~al\mbox{.}(2020)]%
        {li2020accurate}
\bibfield{author}{\bibinfo{person}{Jiuyong Li}, \bibinfo{person}{Lin Liu},
  {et~al\mbox{.}}} \bibinfo{year}{2020}\natexlab{}.
\newblock \showarticletitle{Accurate data-driven prediction does not mean high
  reproducibility}.
\newblock \bibinfo{journal}{\emph{Nature machine intelligence}}
  \bibinfo{volume}{2}, \bibinfo{number}{1} (\bibinfo{year}{2020}),
  \bibinfo{pages}{13--15}.
\newblock


\bibitem[Louizos et~al\mbox{.}(2017)]%
        {louizos2017causal}
\bibfield{author}{\bibinfo{person}{Christos Louizos}, \bibinfo{person}{Uri
  Shalit}, \bibinfo{person}{Joris Mooij}, {et~al\mbox{.}}}
  \bibinfo{year}{2017}\natexlab{}.
\newblock \showarticletitle{Causal effect inference with deep latent-variable
  models}. In \bibinfo{booktitle}{\emph{The 31st International Conference on
  Neural Information Processing Systems}}. \bibinfo{pages}{6449--6459}.
\newblock


\bibitem[Ma et~al\mbox{.}(2021)]%
        {ma2021multi}
\bibfield{author}{\bibinfo{person}{Jing Ma}, \bibinfo{person}{Ruocheng Guo},
  \bibinfo{person}{Aidong Zhang}, {and} \bibinfo{person}{Jundong Li}.}
  \bibinfo{year}{2021}\natexlab{}.
\newblock \showarticletitle{Multi-cause effect estimation with disentangled
  confounder representation}. In \bibinfo{booktitle}{\emph{Proceedings of the
  Thirtieth International Joint Conference on Artificial Intelligence}}.
  \bibinfo{pages}{2790--2796}.
\newblock


\bibitem[Maathuis et~al\mbox{.}(2015)]%
        {maathuis2015generalized}
\bibfield{author}{\bibinfo{person}{Marloes~H Maathuis}, \bibinfo{person}{Diego
  Colombo}, {et~al\mbox{.}}} \bibinfo{year}{2015}\natexlab{}.
\newblock \showarticletitle{A generalized back-door criterion}.
\newblock \bibinfo{journal}{\emph{The Annals of Statistics}}
  \bibinfo{volume}{43}, \bibinfo{number}{3} (\bibinfo{year}{2015}),
  \bibinfo{pages}{1060--1088}.
\newblock


\bibitem[Maathuis et~al\mbox{.}(2010)]%
        {maathuis2010predicting}
\bibfield{author}{\bibinfo{person}{Marloes~H Maathuis}, \bibinfo{person}{Diego
  Colombo}, \bibinfo{person}{Markus Kalisch}, {and} \bibinfo{person}{Peter
  B{\"u}hlmann}.} \bibinfo{year}{2010}\natexlab{}.
\newblock \showarticletitle{Predicting causal effects in large-scale systems
  from observational data}.
\newblock \bibinfo{journal}{\emph{Nature Methods}} \bibinfo{volume}{7},
  \bibinfo{number}{4} (\bibinfo{year}{2010}), \bibinfo{pages}{247--248}.
\newblock


\bibitem[Maathuis et~al\mbox{.}(2009)]%
        {maathuis2009estimating}
\bibfield{author}{\bibinfo{person}{Marloes~H Maathuis}, \bibinfo{person}{Markus
  Kalisch}, \bibinfo{person}{Peter B{\"u}hlmann}, {et~al\mbox{.}}}
  \bibinfo{year}{2009}\natexlab{}.
\newblock \showarticletitle{Estimating high-dimensional intervention effects
  from observational data}.
\newblock \bibinfo{journal}{\emph{The Annals of Statistics}}
  \bibinfo{volume}{37}, \bibinfo{number}{6A} (\bibinfo{year}{2009}),
  \bibinfo{pages}{3133--3164}.
\newblock


\bibitem[Malinsky and Spirtes(2017)]%
        {malinsky2017estimating}
\bibfield{author}{\bibinfo{person}{Daniel Malinsky} {and}
  \bibinfo{person}{Peter Spirtes}.} \bibinfo{year}{2017}\natexlab{}.
\newblock \showarticletitle{Estimating bounds on causal effects in
  high-dimensional and possibly confounded systems}.
\newblock \bibinfo{journal}{\emph{International Journal of Approximate
  Reasoning}}  \bibinfo{volume}{88} (\bibinfo{year}{2017}),
  \bibinfo{pages}{371--384}.
\newblock


\bibitem[Martens et~al\mbox{.}(2006)]%
        {martens2006instrumental}
\bibfield{author}{\bibinfo{person}{Edwin~P Martens}, \bibinfo{person}{Wiebe~R
  Pestman}, \bibinfo{person}{Anthonius de Boer}, {et~al\mbox{.}}}
  \bibinfo{year}{2006}\natexlab{}.
\newblock \showarticletitle{Instrumental variables: application and
  limitations}.
\newblock \bibinfo{journal}{\emph{Epidemiology}} \bibinfo{volume}{17},
  \bibinfo{number}{3} (\bibinfo{year}{2006}), \bibinfo{pages}{260--267}.
\newblock


\bibitem[Mattei et~al\mbox{.}(2014)]%
        {mattei2014identification}
\bibfield{author}{\bibinfo{person}{Alessandra Mattei},
  \bibinfo{person}{Fabrizia Mealli}, {and} \bibinfo{person}{Barbara Pacini}.}
  \bibinfo{year}{2014}\natexlab{}.
\newblock \showarticletitle{Identification of causal effects in the presence of
  nonignorable missing outcome values}.
\newblock \bibinfo{journal}{\emph{Biometrics}} \bibinfo{volume}{70},
  \bibinfo{number}{2} (\bibinfo{year}{2014}), \bibinfo{pages}{278--288}.
\newblock


\bibitem[Meek(1995)]%
        {meek1995causal}
\bibfield{author}{\bibinfo{person}{Christopher Meek}.}
  \bibinfo{year}{1995}\natexlab{}.
\newblock \showarticletitle{Causal inference and causal explanation with
  background knowledge}. In \bibinfo{booktitle}{\emph{the Eleventh Conference
  on Uncertainty in Artificial Intelligence}}. \bibinfo{pages}{403--411}.
\newblock


\bibitem[Morgan and Harding(2006)]%
        {Morgan2006Matching}
\bibfield{author}{\bibinfo{person}{S.~L. Morgan} {and} \bibinfo{person}{D.~J.
  Harding}.} \bibinfo{year}{2006}\natexlab{}.
\newblock \showarticletitle{Matching Estimators of Causal Effects: Prospects
  and Pitfalls in Theory and Practice}.
\newblock \bibinfo{journal}{\emph{Sociological Methods \& Research}}
  \bibinfo{volume}{35}, \bibinfo{number}{1} (\bibinfo{year}{2006}),
  \bibinfo{pages}{3--60}.
\newblock


\bibitem[Morgan and Winship(2015)]%
        {morgan2015counterfactuals}
\bibfield{author}{\bibinfo{person}{Stephen~L Morgan} {and}
  \bibinfo{person}{Christopher Winship}.} \bibinfo{year}{2015}\natexlab{}.
\newblock \bibinfo{booktitle}{\emph{Counterfactuals and {C}ausal {I}nference}}.
\newblock \bibinfo{publisher}{Cambridge University Press}.
\newblock


\bibitem[Nabi et~al\mbox{.}(2022)]%
        {nabi2022semiparametric}
\bibfield{author}{\bibinfo{person}{Razieh Nabi}, \bibinfo{person}{Todd McNutt},
  {and} \bibinfo{person}{Ilya Shpitser}.} \bibinfo{year}{2022}\natexlab{}.
\newblock \showarticletitle{Semiparametric causal sufficient dimension
  reduction of multidimensional treatments}. In
  \bibinfo{booktitle}{\emph{Uncertainty in Artificial Intelligence}}. PMLR,
  \bibinfo{pages}{1445--1455}.
\newblock


\bibitem[Neapolitan et~al\mbox{.}(2004)]%
        {neapolitan2004learning}
\bibfield{author}{\bibinfo{person}{Richard~E Neapolitan} {et~al\mbox{.}}}
  \bibinfo{year}{2004}\natexlab{}.
\newblock \bibinfo{booktitle}{\emph{Learning {B}ayesian {N}etworks}}.
  Vol.~\bibinfo{volume}{38}.
\newblock \bibinfo{publisher}{Pearson Prentice Hall Upper Saddle River, NJ}.
\newblock


\bibitem[Nogueira et~al\mbox{.}(2022)]%
        {nogueira2022methods}
\bibfield{author}{\bibinfo{person}{Ana~Rita Nogueira}, \bibinfo{person}{Andrea
  Pugnana}, \bibinfo{person}{Salvatore Ruggieri}, \bibinfo{person}{Dino
  Pedreschi}, {and} \bibinfo{person}{Jo{\~a}o Gama}.}
  \bibinfo{year}{2022}\natexlab{}.
\newblock \showarticletitle{Methods and tools for causal discovery and causal
  inference}.
\newblock \bibinfo{journal}{\emph{Wiley Interdisciplinary Reviews: Data Mining
  and Knowledge Discovery}} \bibinfo{volume}{12}, \bibinfo{number}{2}
  (\bibinfo{year}{2022}), \bibinfo{pages}{e1449}.
\newblock


\bibitem[Pearl(1995a)]%
        {pearl1995causal}
\bibfield{author}{\bibinfo{person}{Judea Pearl}.}
  \bibinfo{year}{1995}\natexlab{a}.
\newblock \showarticletitle{Causal diagrams for empirical research}.
\newblock \bibinfo{journal}{\emph{Biometrika}} \bibinfo{volume}{82},
  \bibinfo{number}{4} (\bibinfo{year}{1995}), \bibinfo{pages}{669--688}.
\newblock


\bibitem[Pearl(1995b)]%
        {pearl1995testability}
\bibfield{author}{\bibinfo{person}{Judea Pearl}.}
  \bibinfo{year}{1995}\natexlab{b}.
\newblock \showarticletitle{On the testability of causal models with latent and
  instrumental variables}. In \bibinfo{booktitle}{\emph{The Conference on
  Uncertainty in Artificial Intelligence}}. \bibinfo{pages}{435--443}.
\newblock


\bibitem[Pearl(2009a)]%
        {pearl2009causality}
\bibfield{author}{\bibinfo{person}{Judea Pearl}.}
  \bibinfo{year}{2009}\natexlab{a}.
\newblock \bibinfo{booktitle}{\emph{Causality: {M}odels, {R}easoning, and
  {I}nference}}.
\newblock \bibinfo{publisher}{Cambridge university press}.
\newblock


\bibitem[Pearl(2009b)]%
        {pearl2009myth}
\bibfield{author}{\bibinfo{person}{Judea Pearl}.}
  \bibinfo{year}{2009}\natexlab{b}.
\newblock \showarticletitle{Myth, confusion, and science in causal analysis}.
\newblock \bibinfo{journal}{\emph{Tech. Rep. R-348}} (\bibinfo{year}{2009}).
\newblock
\newblock
\shownote{Los Angeles, CA: University of California}.


\bibitem[Pearl et~al\mbox{.}(2009)]%
        {pearl2009causal}
\bibfield{author}{\bibinfo{person}{Judea Pearl} {et~al\mbox{.}}}
  \bibinfo{year}{2009}\natexlab{}.
\newblock \showarticletitle{Causal inference in statistics: An overview}.
\newblock \bibinfo{journal}{\emph{Statistics Surveys}}  \bibinfo{volume}{3}
  (\bibinfo{year}{2009}), \bibinfo{pages}{96--146}.
\newblock


\bibitem[Pearl and Mackenzie(2018)]%
        {pearl2018book}
\bibfield{author}{\bibinfo{person}{Judea Pearl} {and} \bibinfo{person}{Dana
  Mackenzie}.} \bibinfo{year}{2018}\natexlab{}.
\newblock \bibinfo{booktitle}{\emph{The {B}ook of {W}hy: The {N}ew {S}cience of
  {C}ause and {E}ffect}}.
\newblock \bibinfo{publisher}{Basic Books}.
\newblock


\bibitem[Pe{\~n}a(2018)]%
        {pena2018reasoning}
\bibfield{author}{\bibinfo{person}{Jose~M Pe{\~n}a}.}
  \bibinfo{year}{2018}\natexlab{}.
\newblock \showarticletitle{Reasoning with alternative acyclic directed mixed
  graphs}.
\newblock \bibinfo{journal}{\emph{Behaviormetrika}} \bibinfo{volume}{45},
  \bibinfo{number}{2} (\bibinfo{year}{2018}), \bibinfo{pages}{389--422}.
\newblock


\bibitem[Perkovic et~al\mbox{.}(2017)]%
        {perkovic2017interpreting}
\bibfield{author}{\bibinfo{person}{Emilija Perkovic}, \bibinfo{person}{Markus
  Kalisch}, {and} \bibinfo{person}{Marloes~H Maathuis}.}
  \bibinfo{year}{2017}\natexlab{}.
\newblock \showarticletitle{Interpreting and using {CPDAG}s with background
  knowledge}. In \bibinfo{booktitle}{\emph{The Conference on Uncertainty in
  Artificial Intelligence}}. AUAI Press, \bibinfo{pages}{ID--120}.
\newblock


\bibitem[Perkovi{\'c} et~al\mbox{.}(2018)]%
        {perkovic2018complete}
\bibfield{author}{\bibinfo{person}{Emilija Perkovi{\'c}},
  \bibinfo{person}{Johannes Textor}, \bibinfo{person}{Markus Kalisch}, {and}
  \bibinfo{person}{Marloes~H Maathuis}.} \bibinfo{year}{2018}\natexlab{}.
\newblock \showarticletitle{Complete graphical characterization and
  construction of adjustment sets in {M}arkov equivalence classes of ancestral
  graphs}.
\newblock \bibinfo{journal}{\emph{The Journal of Machine Learning Research}}
  \bibinfo{volume}{18}, \bibinfo{number}{1} (\bibinfo{year}{2018}),
  \bibinfo{pages}{8132--8193}.
\newblock


\bibitem[Peters et~al\mbox{.}(2017)]%
        {peters2017elements}
\bibfield{author}{\bibinfo{person}{Jonas Peters}, \bibinfo{person}{Dominik
  Janzing}, {and} \bibinfo{person}{Bernhard Sch{\"o}lkopf}.}
  \bibinfo{year}{2017}\natexlab{}.
\newblock \bibinfo{booktitle}{\emph{Elements of {C}ausal {I}nference}}.
\newblock \bibinfo{publisher}{The MIT Press}.
\newblock


\bibitem[Ren et~al\mbox{.}(2023)]%
        {rengraphical2023}
\bibfield{author}{\bibinfo{person}{Shangsi Ren}, \bibinfo{person}{Cameron~A
  Beeche}, {et~al\mbox{.}}} \bibinfo{year}{2023}\natexlab{}.
\newblock \showarticletitle{Graphical modeling of causal factors associated
  with the postoperative survival of esophageal cancer subjects}.
\newblock \bibinfo{journal}{\emph{Medical Physics}} (\bibinfo{year}{2023}),
  \bibinfo{pages}{1--10}.
\newblock


\bibitem[Richardson(2003)]%
        {richardson2003markov}
\bibfield{author}{\bibinfo{person}{Thomas Richardson}.}
  \bibinfo{year}{2003}\natexlab{}.
\newblock \showarticletitle{Markov properties for acyclic directed mixed
  graphs}.
\newblock \bibinfo{journal}{\emph{Scandinavian Journal of Statistics}}
  \bibinfo{volume}{30}, \bibinfo{number}{1} (\bibinfo{year}{2003}),
  \bibinfo{pages}{145--157}.
\newblock


\bibitem[Richardson and Spirtes(2002)]%
        {richardson2002ancestral}
\bibfield{author}{\bibinfo{person}{Thomas Richardson} {and}
  \bibinfo{person}{Peter Spirtes}.} \bibinfo{year}{2002}\natexlab{}.
\newblock \showarticletitle{Ancestral graph {M}arkov models}.
\newblock \bibinfo{journal}{\emph{The Annals of Statistics}}
  \bibinfo{volume}{30}, \bibinfo{number}{4} (\bibinfo{year}{2002}),
  \bibinfo{pages}{962--1030}.
\newblock


\bibitem[Richardson and Spirtes(2003)]%
        {spirtes2003causal}
\bibfield{author}{\bibinfo{person}{Thomas~S Richardson} {and}
  \bibinfo{person}{Peter Spirtes}.} \bibinfo{year}{2003}\natexlab{}.
\newblock \showarticletitle{Causal inference via ancestral graph models}.
\newblock \bibinfo{journal}{\emph{Oxford Statistical Science Series}}
  \bibinfo{volume}{27} (\bibinfo{year}{2003}), \bibinfo{pages}{83--105}.
\newblock


\bibitem[Robins(1986)]%
        {robins1986new}
\bibfield{author}{\bibinfo{person}{James Robins}.}
  \bibinfo{year}{1986}\natexlab{}.
\newblock \showarticletitle{A new approach to causal inference in mortality
  studies with a sustained exposure period—application to control of the
  healthy worker survivor effect}.
\newblock \bibinfo{journal}{\emph{Mathematical Modelling}} \bibinfo{volume}{7},
  \bibinfo{number}{9-12} (\bibinfo{year}{1986}), \bibinfo{pages}{1393--1512}.
\newblock


\bibitem[Robins(1997)]%
        {robins1997causal}
\bibfield{author}{\bibinfo{person}{James~M Robins}.}
  \bibinfo{year}{1997}\natexlab{}.
\newblock \showarticletitle{Causal inference from complex longitudinal data}.
\newblock In \bibinfo{booktitle}{\emph{Latent variable modeling and
  applications to causality}}. \bibinfo{publisher}{Springer},
  \bibinfo{pages}{69--117}.
\newblock


\bibitem[Robins and Greenland(1992)]%
        {robins1992identifiability}
\bibfield{author}{\bibinfo{person}{James~M Robins} {and}
  \bibinfo{person}{Sander Greenland}.} \bibinfo{year}{1992}\natexlab{}.
\newblock \showarticletitle{Identifiability and exchangeability for direct and
  indirect effects}.
\newblock \bibinfo{journal}{\emph{Epidemiology}} \bibinfo{volume}{3},
  \bibinfo{number}{2} (\bibinfo{year}{1992}), \bibinfo{pages}{143--155}.
\newblock


\bibitem[Robins et~al\mbox{.}(2000a)]%
        {robins2000marginal}
\bibfield{author}{\bibinfo{person}{James~M Robins},
  \bibinfo{person}{Miguel~Angel Hern{\'a}n}, {and} \bibinfo{person}{Babette
  Brumback}.} \bibinfo{year}{2000}\natexlab{a}.
\newblock \showarticletitle{Marginal Structural Models and Causal Inference in
  Epidemiology}.
\newblock \bibinfo{journal}{\emph{Epidemiology}} \bibinfo{volume}{11},
  \bibinfo{number}{5} (\bibinfo{year}{2000}), \bibinfo{pages}{551}.
\newblock


\bibitem[Robins et~al\mbox{.}(2000b)]%
        {robins2000sensitivity}
\bibfield{author}{\bibinfo{person}{James~M Robins}, \bibinfo{person}{Andrea
  Rotnitzky}, {and} \bibinfo{person}{Daniel~O Scharfstein}.}
  \bibinfo{year}{2000}\natexlab{b}.
\newblock \showarticletitle{Sensitivity analysis for selection bias and
  unmeasured confounding in missing data and causal inference models}.
\newblock   \bibinfo{volume}{116} (\bibinfo{year}{2000}),
  \bibinfo{pages}{1--94}.
\newblock


\bibitem[Rosenbaum and Rubin(1985)]%
        {rosenbaum1985bias}
\bibfield{author}{\bibinfo{person}{Paul~R Rosenbaum} {and}
  \bibinfo{person}{Donald~B Rubin}.} \bibinfo{year}{1985}\natexlab{}.
\newblock \showarticletitle{The bias due to incomplete matching}.
\newblock \bibinfo{journal}{\emph{Biometrics}} \bibinfo{volume}{41},
  \bibinfo{number}{1} (\bibinfo{year}{1985}), \bibinfo{pages}{103--116}.
\newblock


\bibitem[Rotnitzky and Smucler(2020)]%
        {rotnitzky2020efficient}
\bibfield{author}{\bibinfo{person}{Andrea Rotnitzky} {and}
  \bibinfo{person}{Ezequiel Smucler}.} \bibinfo{year}{2020}\natexlab{}.
\newblock \showarticletitle{Efficient adjustment sets for population average
  causal treatment effect estimation in graphical models}.
\newblock \bibinfo{journal}{\emph{The Journal of Machine Learning Research}}
  \bibinfo{volume}{21}, \bibinfo{number}{1} (\bibinfo{year}{2020}),
  \bibinfo{pages}{7642--7727}.
\newblock


\bibitem[Rubin(1973)]%
        {rubin1973matching}
\bibfield{author}{\bibinfo{person}{Donald~B Rubin}.}
  \bibinfo{year}{1973}\natexlab{}.
\newblock \showarticletitle{Matching to remove bias in observational studies}.
\newblock \bibinfo{journal}{\emph{Biometrics}}  \bibinfo{volume}{29}
  (\bibinfo{year}{1973}), \bibinfo{pages}{159--183}.
\newblock


\bibitem[Rubin(1974)]%
        {rubin1974estimating}
\bibfield{author}{\bibinfo{person}{Donald~B Rubin}.}
  \bibinfo{year}{1974}\natexlab{}.
\newblock \showarticletitle{Estimating causal effects of treatments in
  randomized and nonrandomized studies.}
\newblock \bibinfo{journal}{\emph{Journal of Educational Psychology}}
  \bibinfo{volume}{66}, \bibinfo{number}{5} (\bibinfo{year}{1974}),
  \bibinfo{pages}{688}.
\newblock


\bibitem[Rubin(2007)]%
        {rubin2007design}
\bibfield{author}{\bibinfo{person}{Donald~B Rubin}.}
  \bibinfo{year}{2007}\natexlab{}.
\newblock \showarticletitle{The design versus the analysis of observational
  studies for causal effects: parallels with the design of randomized trials}.
\newblock \bibinfo{journal}{\emph{Statistics in Medicine}}
  \bibinfo{volume}{26}, \bibinfo{number}{1} (\bibinfo{year}{2007}),
  \bibinfo{pages}{20--36}.
\newblock


\bibitem[Runge(2021)]%
        {runge2021necessary}
\bibfield{author}{\bibinfo{person}{Jakob Runge}.}
  \bibinfo{year}{2021}\natexlab{}.
\newblock \showarticletitle{Necessary and sufficient graphical conditions for
  optimal adjustment sets in causal graphical models with hidden variables}.
\newblock \bibinfo{journal}{\emph{Advances in Neural Information Processing
  Systems}}  \bibinfo{volume}{34} (\bibinfo{year}{2021}),
  \bibinfo{pages}{15762--15773}.
\newblock


\bibitem[Runge et~al\mbox{.}(2019)]%
        {runge2019inferring}
\bibfield{author}{\bibinfo{person}{Jakob Runge}, \bibinfo{person}{Sebastian
  Bathiany}, \bibinfo{person}{Erik Bollt}, \bibinfo{person}{Gustau
  Camps-Valls}, \bibinfo{person}{Dim Coumou}, \bibinfo{person}{Ethan Deyle},
  \bibinfo{person}{Clark Glymour}, \bibinfo{person}{Marlene Kretschmer},
  \bibinfo{person}{Miguel~D Mahecha}, \bibinfo{person}{Jordi
  Mu{\~n}oz-Mar{\'\i}}, {et~al\mbox{.}}} \bibinfo{year}{2019}\natexlab{}.
\newblock \showarticletitle{Inferring causation from time series in Earth
  system sciences}.
\newblock \bibinfo{journal}{\emph{Nature communications}} \bibinfo{volume}{10},
  \bibinfo{number}{1} (\bibinfo{year}{2019}), \bibinfo{pages}{2553}.
\newblock


\bibitem[Sachs et~al\mbox{.}(2005)]%
        {sachs2005causal}
\bibfield{author}{\bibinfo{person}{Karen Sachs}, \bibinfo{person}{Omar Perez},
  {et~al\mbox{.}}} \bibinfo{year}{2005}\natexlab{}.
\newblock \showarticletitle{Causal protein-signaling networks derived from
  multiparameter single-cell data}.
\newblock \bibinfo{journal}{\emph{Science}} \bibinfo{volume}{308},
  \bibinfo{number}{5721} (\bibinfo{year}{2005}), \bibinfo{pages}{523--529}.
\newblock


\bibitem[Sauer et~al\mbox{.}(2013)]%
        {sauer2013review}
\bibfield{author}{\bibinfo{person}{Brian~C Sauer}, \bibinfo{person}{M~Alan
  Brookhart}, \bibinfo{person}{Jason Roy}, {and} \bibinfo{person}{Tyler
  VanderWeele}.} \bibinfo{year}{2013}\natexlab{}.
\newblock \showarticletitle{A review of covariate selection for
  non-experimental comparative effectiveness research}.
\newblock \bibinfo{journal}{\emph{Pharmacoepidemiology and drug safety}}
  \bibinfo{volume}{22}, \bibinfo{number}{11} (\bibinfo{year}{2013}),
  \bibinfo{pages}{1139--1145}.
\newblock


\bibitem[Scharfstein et~al\mbox{.}(2021)]%
        {scharfstein2021semiparametric}
\bibfield{author}{\bibinfo{person}{Daniel~O Scharfstein},
  \bibinfo{person}{Razieh Nabi}, \bibinfo{person}{Edward~H Kennedy},
  \bibinfo{person}{Ming-Yueh Huang}, \bibinfo{person}{Matteo Bonvini}, {and}
  \bibinfo{person}{Marcela Smid}.} \bibinfo{year}{2021}\natexlab{}.
\newblock \showarticletitle{Semiparametric sensitivity analysis: Unmeasured
  confounding in observational studies}.
\newblock \bibinfo{journal}{\emph{arXiv preprint arXiv:2104.08300}}
  (\bibinfo{year}{2021}).
\newblock


\bibitem[Sch{\"o}lkopf(2022)]%
        {scholkopf2022causality}
\bibfield{author}{\bibinfo{person}{Bernhard Sch{\"o}lkopf}.}
  \bibinfo{year}{2022}\natexlab{}.
\newblock \showarticletitle{Causality for machine learning}.
\newblock In \bibinfo{booktitle}{\emph{Probabilistic and Causal Inference: The
  Works of Judea Pearl}}. \bibinfo{pages}{765--804}.
\newblock


\bibitem[Scutari(2010)]%
        {scutari2010learning}
\bibfield{author}{\bibinfo{person}{M Scutari}.}
  \bibinfo{year}{2010}\natexlab{}.
\newblock \showarticletitle{Learning Bayesian networks with the bnlearn R
  Package}.
\newblock \bibinfo{journal}{\emph{Journal of Statistical Software}}
  \bibinfo{volume}{35}, \bibinfo{number}{3} (\bibinfo{year}{2010}),
  \bibinfo{pages}{1--22}.
\newblock


\bibitem[Sekhon(2011)]%
        {sekhon2008multivariate}
\bibfield{author}{\bibinfo{person}{Jasjeet~S Sekhon}.}
  \bibinfo{year}{2011}\natexlab{}.
\newblock \showarticletitle{Multivariate and Propensity Score Matching Software
  with Automated Balance Optimization: The Matching package for {R}}.
\newblock \bibinfo{journal}{\emph{Journal of Statistical Software}}
  \bibinfo{volume}{42} (\bibinfo{year}{2011}), \bibinfo{pages}{1--52}.
\newblock


\bibitem[Shalit et~al\mbox{.}(2017)]%
        {shalit2017estimating}
\bibfield{author}{\bibinfo{person}{Uri Shalit}, \bibinfo{person}{Fredrik~D
  Johansson}, {and} \bibinfo{person}{David Sontag}.}
  \bibinfo{year}{2017}\natexlab{}.
\newblock \showarticletitle{Estimating individual treatment effect:
  generalization bounds and algorithms}. In
  \bibinfo{booktitle}{\emph{International Conference on Machine Learning}}.
  PMLR, \bibinfo{pages}{3076--3085}.
\newblock


\bibitem[Shpitser and Pearl(2006)]%
        {shpitser2006identification}
\bibfield{author}{\bibinfo{person}{Ilya Shpitser} {and} \bibinfo{person}{Judea
  Pearl}.} \bibinfo{year}{2006}\natexlab{}.
\newblock \showarticletitle{Identification of joint interventional
  distributions in recursive semi-{M}arkovian causal models}. In
  \bibinfo{booktitle}{\emph{Proceedings of the National Conference on
  Artificial Intelligence}}, Vol.~\bibinfo{volume}{21}.
  \bibinfo{pages}{1219--1226}.
\newblock


\bibitem[Shpitser and Pearl(2008)]%
        {shpitser2008complete}
\bibfield{author}{\bibinfo{person}{Ilya Shpitser} {and} \bibinfo{person}{Judea
  Pearl}.} \bibinfo{year}{2008}\natexlab{}.
\newblock \showarticletitle{Complete identification methods for the causal
  hierarchy}.
\newblock \bibinfo{journal}{\emph{Journal of Machine Learning Research}}
  \bibinfo{volume}{9}, \bibinfo{number}{Sep} (\bibinfo{year}{2008}),
  \bibinfo{pages}{1941--1979}.
\newblock


\bibitem[Shpitser et~al\mbox{.}(2010)]%
        {shpitser2012validity}
\bibfield{author}{\bibinfo{person}{Ilya Shpitser}, \bibinfo{person}{Tyler
  VanderWeele}, {and} \bibinfo{person}{James~M Robins}.}
  \bibinfo{year}{2010}\natexlab{}.
\newblock \showarticletitle{On the validity of covariate adjustment for
  estimating causal effects}. In \bibinfo{booktitle}{\emph{The Twenty-Sixth
  Conference on Uncertainty in Artificial Intelligence}}.
  \bibinfo{pages}{527--536}.
\newblock


\bibitem[Sieswerda et~al\mbox{.}(2023)]%
        {sieswerda2023identifying}
\bibfield{author}{\bibinfo{person}{Melle Sieswerda}, \bibinfo{person}{Shixuan
  Xie}, {et~al\mbox{.}}} \bibinfo{year}{2023}\natexlab{}.
\newblock \showarticletitle{Identifying confounders using bayesian networks and
  estimating treatment effect in prostate cancer with observational data}.
\newblock \bibinfo{journal}{\emph{JCO Clinical Cancer Informatics}}
  \bibinfo{volume}{7} (\bibinfo{year}{2023}), \bibinfo{pages}{e2200080}.
\newblock


\bibitem[Silva et~al\mbox{.}(2011)]%
        {silva2011mixed}
\bibfield{author}{\bibinfo{person}{Ricardo Silva}, \bibinfo{person}{Charles
  Blundell}, {and} \bibinfo{person}{Yee~Whye Teh}.}
  \bibinfo{year}{2011}\natexlab{}.
\newblock \showarticletitle{Mixed cumulative distribution networks}. In
  \bibinfo{booktitle}{\emph{Proceedings of the Fourteenth International
  Conference on Artificial Intelligence and Statistics}}. JMLR Workshop and
  Conference Proceedings, \bibinfo{pages}{670--678}.
\newblock


\bibitem[Silva and Shimizu(2017)]%
        {silva2017learning}
\bibfield{author}{\bibinfo{person}{Ricardo Silva} {and} \bibinfo{person}{Shohei
  Shimizu}.} \bibinfo{year}{2017}\natexlab{}.
\newblock \showarticletitle{Learning instrumental variables with structural and
  non-gaussianity assumptions}.
\newblock \bibinfo{journal}{\emph{Journal of Machine Learning Research}}
  \bibinfo{volume}{18}, \bibinfo{number}{120} (\bibinfo{year}{2017}),
  \bibinfo{pages}{1--49}.
\newblock


\bibitem[Singh et~al\mbox{.}(2019)]%
        {singh2019kernel}
\bibfield{author}{\bibinfo{person}{Rahul Singh}, \bibinfo{person}{Maneesh
  Sahani}, {and} \bibinfo{person}{Arthur Gretton}.}
  \bibinfo{year}{2019}\natexlab{}.
\newblock \showarticletitle{Kernel instrumental variable regression}. In
  \bibinfo{booktitle}{\emph{International Conference on Neural Information
  Processing Systems}}. \bibinfo{pages}{4593--4605}.
\newblock


\bibitem[Sjolander and Martinussen(2019)]%
        {sjolander2019instrumental}
\bibfield{author}{\bibinfo{person}{Arvid Sjolander} {and}
  \bibinfo{person}{Torben Martinussen}.} \bibinfo{year}{2019}\natexlab{}.
\newblock \showarticletitle{Instrumental variable estimation with the {R}
  package ivtools}.
\newblock \bibinfo{journal}{\emph{Epidemiologic Methods}} \bibinfo{volume}{8},
  \bibinfo{number}{1} (\bibinfo{year}{2019}), \bibinfo{pages}{1--20}.
\newblock


\bibitem[Sobel and Lindquist(2020)]%
        {sobel2020estimating}
\bibfield{author}{\bibinfo{person}{Michael~E Sobel} {and}
  \bibinfo{person}{Martin~A Lindquist}.} \bibinfo{year}{2020}\natexlab{}.
\newblock \showarticletitle{Estimating causal effects in studies of human brain
  function: New models, methods and estimands}.
\newblock \bibinfo{journal}{\emph{The annals of applied statistics}}
  \bibinfo{volume}{14}, \bibinfo{number}{1} (\bibinfo{year}{2020}),
  \bibinfo{pages}{452}.
\newblock


\bibitem[Spirtes(2010)]%
        {spirtes2010introduction}
\bibfield{author}{\bibinfo{person}{Peter Spirtes}.}
  \bibinfo{year}{2010}\natexlab{}.
\newblock \showarticletitle{Introduction to causal inference.}
\newblock \bibinfo{journal}{\emph{Journal of Machine Learning Research}}
  \bibinfo{volume}{11}, \bibinfo{number}{5} (\bibinfo{year}{2010}),
  \bibinfo{pages}{1643--1662}.
\newblock


\bibitem[Spirtes et~al\mbox{.}(2000)]%
        {spirtes2000causation}
\bibfield{author}{\bibinfo{person}{Peter Spirtes}, \bibinfo{person}{Clark~N
  Glymour}, \bibinfo{person}{Richard Scheines}, {et~al\mbox{.}}}
  \bibinfo{year}{2000}\natexlab{}.
\newblock \bibinfo{booktitle}{\emph{Causation, {P}rediction, and {S}earch}}.
\newblock \bibinfo{publisher}{MIT Press}.
\newblock


\bibitem[Stuart(2010)]%
        {stuart2010matching}
\bibfield{author}{\bibinfo{person}{Elizabeth~A Stuart}.}
  \bibinfo{year}{2010}\natexlab{}.
\newblock \showarticletitle{Matching methods for causal inference: A review and
  a look forward}.
\newblock \bibinfo{journal}{\emph{Statistical science: a review journal of the
  Institute of Mathematical Statistics}} \bibinfo{volume}{25},
  \bibinfo{number}{1} (\bibinfo{year}{2010}), \bibinfo{pages}{1--21}.
\newblock


\bibitem[Sun et~al\mbox{.}(2021)]%
        {sun2021identification}
\bibfield{author}{\bibinfo{person}{Xiaoru Sun}, \bibinfo{person}{Lu Wang},
  {et~al\mbox{.}}} \bibinfo{year}{2021}\natexlab{}.
\newblock \showarticletitle{Identification of microenvironment related
  potential biomarkers of biochemical recurrence at 3 years after prostatectomy
  in prostate adenocarcinoma}.
\newblock \bibinfo{journal}{\emph{Aging (Albany NY)}} \bibinfo{volume}{13},
  \bibinfo{number}{12} (\bibinfo{year}{2021}), \bibinfo{pages}{16024}.
\newblock


\bibitem[Textor et~al\mbox{.}(2016)]%
        {textor2016robust}
\bibfield{author}{\bibinfo{person}{Johannes Textor}, \bibinfo{person}{Benito
  van~der Zander}, \bibinfo{person}{Mark~S Gilthorpe}, {et~al\mbox{.}}}
  \bibinfo{year}{2016}\natexlab{}.
\newblock \showarticletitle{Robust causal inference using directed acyclic
  graphs: the {R} package ‘dagitty’}.
\newblock \bibinfo{journal}{\emph{International journal of epidemiology}}
  \bibinfo{volume}{45}, \bibinfo{number}{6} (\bibinfo{year}{2016}),
  \bibinfo{pages}{1887--1894}.
\newblock


\bibitem[Tian and Pearl(2002)]%
        {tian2002general}
\bibfield{author}{\bibinfo{person}{Jin Tian} {and} \bibinfo{person}{Judea
  Pearl}.} \bibinfo{year}{2002}\natexlab{}.
\newblock \showarticletitle{A general identification condition for causal
  effects}. In \bibinfo{booktitle}{\emph{Aaai/iaai}}.
  \bibinfo{pages}{567--573}.
\newblock


\bibitem[Tibshirani(1996)]%
        {tibshirani1996regression}
\bibfield{author}{\bibinfo{person}{Robert Tibshirani}.}
  \bibinfo{year}{1996}\natexlab{}.
\newblock \showarticletitle{Regression shrinkage and selection via the lasso}.
\newblock \bibinfo{journal}{\emph{Journal of the Royal Statistical Society:
  Series B (Methodological)}} \bibinfo{volume}{58}, \bibinfo{number}{1}
  (\bibinfo{year}{1996}), \bibinfo{pages}{267--288}.
\newblock


\bibitem[Tortorelli and Michaleris(1994)]%
        {tortorelli1994design}
\bibfield{author}{\bibinfo{person}{Daniel~A Tortorelli} {and}
  \bibinfo{person}{Panagiotis Michaleris}.} \bibinfo{year}{1994}\natexlab{}.
\newblock \showarticletitle{Design sensitivity analysis: overview and review}.
\newblock \bibinfo{journal}{\emph{Inverse problems in Engineering}}
  \bibinfo{volume}{1}, \bibinfo{number}{1} (\bibinfo{year}{1994}),
  \bibinfo{pages}{71--105}.
\newblock


\bibitem[Tsamardinos et~al\mbox{.}(2006)]%
        {tsamardinos2006max}
\bibfield{author}{\bibinfo{person}{Ioannis Tsamardinos},
  \bibinfo{person}{Laura~E Brown}, {and} \bibinfo{person}{Constantin~F
  Aliferis}.} \bibinfo{year}{2006}\natexlab{}.
\newblock \showarticletitle{The max-min hill-climbing {B}ayesian network
  structure learning algorithm}.
\newblock \bibinfo{journal}{\emph{Machine learning}} \bibinfo{volume}{65},
  \bibinfo{number}{1} (\bibinfo{year}{2006}), \bibinfo{pages}{31--78}.
\newblock


\bibitem[van~der Zander et~al\mbox{.}(2014)]%
        {van2014constructing}
\bibfield{author}{\bibinfo{person}{Benito van~der Zander},
  \bibinfo{person}{Maciej Li{\'s}kiewicz}, {and} \bibinfo{person}{Johannes
  Textor}.} \bibinfo{year}{2014}\natexlab{}.
\newblock \showarticletitle{Constructing separators and adjustment sets in
  ancestral graphs}. In \bibinfo{booktitle}{\emph{The Thirtieth Conference on
  Uncertainty in Artificial Intelligence}}. \bibinfo{pages}{907--916}.
\newblock


\bibitem[Van~der Zander et~al\mbox{.}(2015)]%
        {van2015efficiently}
\bibfield{author}{\bibinfo{person}{Benito Van~der Zander},
  \bibinfo{person}{Maciej Li{\'s}kiewicz}, {and} \bibinfo{person}{Johannes
  Textor}.} \bibinfo{year}{2015}\natexlab{}.
\newblock \showarticletitle{Efficiently finding conditional instruments for
  causal inference}.
\newblock  (\bibinfo{year}{2015}), \bibinfo{pages}{3243--3249}.
\newblock


\bibitem[van~der Zander et~al\mbox{.}(2019)]%
        {van2019separators}
\bibfield{author}{\bibinfo{person}{Benito van~der Zander},
  \bibinfo{person}{Maciej Li{\'s}kiewicz}, {and} \bibinfo{person}{Johannes
  Textor}.} \bibinfo{year}{2019}\natexlab{}.
\newblock \showarticletitle{Separators and adjustment sets in causal graphs:
  Complete criteria and an algorithmic framework}.
\newblock \bibinfo{journal}{\emph{Artificial Intelligence}}
  \bibinfo{volume}{270} (\bibinfo{year}{2019}), \bibinfo{pages}{1--40}.
\newblock


\bibitem[VanderWeele and Ding(2017)]%
        {vanderweele2017sensitivity}
\bibfield{author}{\bibinfo{person}{Tyler~J VanderWeele} {and}
  \bibinfo{person}{Peng Ding}.} \bibinfo{year}{2017}\natexlab{}.
\newblock \showarticletitle{Sensitivity analysis in observational research:
  introducing the E-value}.
\newblock \bibinfo{journal}{\emph{Annals of internal medicine}}
  \bibinfo{volume}{167}, \bibinfo{number}{4} (\bibinfo{year}{2017}),
  \bibinfo{pages}{268--274}.
\newblock


\bibitem[VanderWeele and Shpitser(2011)]%
        {vanderweele2011new}
\bibfield{author}{\bibinfo{person}{Tyler~J VanderWeele} {and}
  \bibinfo{person}{Ilya Shpitser}.} \bibinfo{year}{2011}\natexlab{}.
\newblock \showarticletitle{A new criterion for confounder selection}.
\newblock \bibinfo{journal}{\emph{Biometrics}} \bibinfo{volume}{67},
  \bibinfo{number}{4} (\bibinfo{year}{2011}), \bibinfo{pages}{1406--1413}.
\newblock


\bibitem[Verbeek(2008)]%
        {verbeek2008guide}
\bibfield{author}{\bibinfo{person}{Marno Verbeek}.}
  \bibinfo{year}{2008}\natexlab{}.
\newblock \bibinfo{booktitle}{\emph{A {G}uide to {M}odern {E}conometrics}}.
\newblock \bibinfo{publisher}{John Wiley \& Sons}.
\newblock


\bibitem[Vowels et~al\mbox{.}(2022)]%
        {vowels2021d}
\bibfield{author}{\bibinfo{person}{Matthew~J Vowels},
  \bibinfo{person}{Necati~Cihan Camgoz}, {and} \bibinfo{person}{Richard
  Bowden}.} \bibinfo{year}{2022}\natexlab{}.
\newblock \showarticletitle{D’ya like dags? a survey on structure learning
  and causal discovery}.
\newblock \bibinfo{journal}{\emph{Comput. Surveys}} \bibinfo{volume}{55},
  \bibinfo{number}{4} (\bibinfo{year}{2022}), \bibinfo{pages}{1--36}.
\newblock


\bibitem[Wang and Blei(2019)]%
        {wang2019blessings}
\bibfield{author}{\bibinfo{person}{Yixin Wang} {and} \bibinfo{person}{David~M
  Blei}.} \bibinfo{year}{2019}\natexlab{}.
\newblock \showarticletitle{The blessings of multiple causes}.
\newblock \bibinfo{journal}{\emph{J. Amer. Statist. Assoc.}}
  \bibinfo{volume}{114}, \bibinfo{number}{528} (\bibinfo{year}{2019}),
  \bibinfo{pages}{1574--1596}.
\newblock


\bibitem[Witte and Didelez(2019)]%
        {witte2019covariate}
\bibfield{author}{\bibinfo{person}{Janine Witte} {and} \bibinfo{person}{Vanessa
  Didelez}.} \bibinfo{year}{2019}\natexlab{}.
\newblock \showarticletitle{Covariate selection strategies for causal
  inference: Classification and comparison}.
\newblock \bibinfo{journal}{\emph{Biometrical Journal}} \bibinfo{volume}{61},
  \bibinfo{number}{5} (\bibinfo{year}{2019}), \bibinfo{pages}{1270--1289}.
\newblock


\bibitem[Witte et~al\mbox{.}(2020)]%
        {witte2020efficient}
\bibfield{author}{\bibinfo{person}{Janine Witte}, \bibinfo{person}{Leonard
  Henckel}, \bibinfo{person}{Marloes~H Maathuis}, {and}
  \bibinfo{person}{Vanessa Didelez}.} \bibinfo{year}{2020}\natexlab{}.
\newblock \showarticletitle{On efficient adjustment in causal graphs}.
\newblock \bibinfo{journal}{\emph{The Journal of Machine Learning Research}}
  \bibinfo{volume}{21}, \bibinfo{number}{1} (\bibinfo{year}{2020}),
  \bibinfo{pages}{9956--10000}.
\newblock


\bibitem[Wu et~al\mbox{.}(2022)]%
        {wu2022instrumental}
\bibfield{author}{\bibinfo{person}{Anpeng Wu}, \bibinfo{person}{Kun Kuang},
  \bibinfo{person}{Ruoxuan Xiong}, {and} \bibinfo{person}{Fei Wu}.}
  \bibinfo{year}{2022}\natexlab{}.
\newblock \showarticletitle{Instrumental Variables in Causal Inference and
  Machine Learning: A Survey}.
\newblock \bibinfo{journal}{\emph{arXiv preprint arXiv:2212.05778}}
  (\bibinfo{year}{2022}).
\newblock


\bibitem[Yao et~al\mbox{.}(2021)]%
        {yao2021survey}
\bibfield{author}{\bibinfo{person}{Liuyi Yao}, \bibinfo{person}{Zhixuan Chu},
  \bibinfo{person}{Sheng Li}, \bibinfo{person}{Yaliang Li},
  \bibinfo{person}{Jing Gao}, {and} \bibinfo{person}{Aidong Zhang}.}
  \bibinfo{year}{2021}\natexlab{}.
\newblock \showarticletitle{A survey on causal inference}.
\newblock \bibinfo{journal}{\emph{ACM Transactions on Knowledge Discovery from
  Data (TKDD)}} \bibinfo{volume}{15}, \bibinfo{number}{5}
  (\bibinfo{year}{2021}), \bibinfo{pages}{1--46}.
\newblock


\bibitem[Yao et~al\mbox{.}(2018)]%
        {yao2018representation}
\bibfield{author}{\bibinfo{person}{Liuyi Yao}, \bibinfo{person}{Sheng Li},
  \bibinfo{person}{Yaliang Li}, {et~al\mbox{.}}}
  \bibinfo{year}{2018}\natexlab{}.
\newblock \showarticletitle{Representation Learning for Treatment Effect
  Estimation from Observational Data}. In \bibinfo{booktitle}{\emph{Advances in
  Neural Information Processing Systems}}. \bibinfo{pages}{2638--2648}.
\newblock


\bibitem[Yu et~al\mbox{.}(2021)]%
        {yu2021unified}
\bibfield{author}{\bibinfo{person}{Kui Yu}, \bibinfo{person}{Lin Liu}, {and}
  \bibinfo{person}{Jiuyong Li}.} \bibinfo{year}{2021}\natexlab{}.
\newblock \showarticletitle{A unified view of causal and non-causal feature
  selection}.
\newblock \bibinfo{journal}{\emph{ACM Transactions on Knowledge Discovery from
  Data}} \bibinfo{volume}{15}, \bibinfo{number}{4} (\bibinfo{year}{2021}),
  \bibinfo{pages}{1--46}.
\newblock


\bibitem[Yuan et~al\mbox{.}(2022)]%
        {yuan2022auto}
\bibfield{author}{\bibinfo{person}{Junkun Yuan}, \bibinfo{person}{Anpeng Wu},
  \bibinfo{person}{Kun Kuang}, {et~al\mbox{.}}}
  \bibinfo{year}{2022}\natexlab{}.
\newblock \showarticletitle{Auto {IV}: Counterfactual Prediction via Automatic
  Instrumental Variable Decomposition}.
\newblock \bibinfo{journal}{\emph{ACM Transactions on Knowledge Discovery from
  Data}} \bibinfo{volume}{16}, \bibinfo{number}{4} (\bibinfo{year}{2022}),
  \bibinfo{pages}{1--20}.
\newblock


\bibitem[Zander and Li{\'s}kiewicz(2016)]%
        {zander2016separators}
\bibfield{author}{\bibinfo{person}{Benito van~der Zander} {and}
  \bibinfo{person}{Maciej Li{\'s}kiewicz}.} \bibinfo{year}{2016}\natexlab{}.
\newblock \showarticletitle{Separators and adjustment sets in Markov equivalent
  DAGs}. In \bibinfo{booktitle}{\emph{Proceedings of the Thirtieth AAAI
  Conference on Artificial Intelligence}}. \bibinfo{pages}{3315--3321}.
\newblock


\bibitem[Zhang(2008a)]%
        {zhang2008causal}
\bibfield{author}{\bibinfo{person}{Jiji Zhang}.}
  \bibinfo{year}{2008}\natexlab{a}.
\newblock \showarticletitle{Causal reasoning with ancestral graphs}.
\newblock \bibinfo{journal}{\emph{Journal of Machine Learning Research}}
  \bibinfo{volume}{9}, \bibinfo{number}{Jul} (\bibinfo{year}{2008}),
  \bibinfo{pages}{1437--1474}.
\newblock


\bibitem[Zhang(2008b)]%
        {zhang2008completeness}
\bibfield{author}{\bibinfo{person}{Jiji Zhang}.}
  \bibinfo{year}{2008}\natexlab{b}.
\newblock \showarticletitle{On the completeness of orientation rules for causal
  discovery in the presence of latent confounders and selection bias}.
\newblock \bibinfo{journal}{\emph{Artificial Intelligence}}
  \bibinfo{volume}{172}, \bibinfo{number}{16-17} (\bibinfo{year}{2008}),
  \bibinfo{pages}{1873--1896}.
\newblock


\bibitem[Zhang et~al\mbox{.}(2016)]%
        {zhang2016predicting}
\bibfield{author}{\bibinfo{person}{Weijia Zhang}, \bibinfo{person}{Thuc~Duy
  Le}, \bibinfo{person}{Lin Liu}, \bibinfo{person}{Zhi-Hua Zhou}, {and}
  \bibinfo{person}{Jiuyong Li}.} \bibinfo{year}{2016}\natexlab{}.
\newblock \showarticletitle{Predicting mi{RNA} targets by integrating gene
  regulatory knowledge with expression profiles}.
\newblock \bibinfo{journal}{\emph{PloS one}} \bibinfo{volume}{11},
  \bibinfo{number}{4} (\bibinfo{year}{2016}), \bibinfo{pages}{e0152860}.
\newblock


\bibitem[Zisoulis et~al\mbox{.}(2012)]%
        {zisoulis2012autoregulation}
\bibfield{author}{\bibinfo{person}{Dimitrios~G Zisoulis},
  \bibinfo{person}{Zoya~S Kai}, \bibinfo{person}{Roger~K Chang}, {and}
  \bibinfo{person}{Amy~E Pasquinelli}.} \bibinfo{year}{2012}\natexlab{}.
\newblock \showarticletitle{Autoregulation of micro{RNA} biogenesis by let-7
  and {A}rgonaute}.
\newblock \bibinfo{journal}{\emph{Nature}} \bibinfo{volume}{486},
  \bibinfo{number}{7404} (\bibinfo{year}{2012}), \bibinfo{pages}{541}.
\newblock


\end{thebibliography}

\end{document}